\documentclass[10pt,twocolumn,letterpaper]{article}

\usepackage{iccv}              
\usepackage{amsmath}
\usepackage{graphicx}  
\usepackage{subcaption} 
\usepackage{adjustbox}  
\definecolor{mydarkgreen}{RGB}{0, 128, 0}
\definecolor{mydarkred}{RGB}{139, 0, 0}

\DeclareMathOperator*{\argmax}{arg\,max}
\newcommand{\R}{\mathbb{R}}
\newcommand{\Z}{\mathbb{Z}}

\newcommand{\meshc}{\mathbf{M}^c}

\newcommand{\vertc}{\mathbf{V}^c}
\newcommand{\verti}{\mathbf{V}}

\newcommand{\facec}{\mathbf{A}^c}
\newcommand{\facei}{\mathbf{A}}
\newcommand{\texc}{\mathbf{F}^c_{3D}}
\newcommand{\texi}{\mathbf{F}_{3D}}

\newcommand{\scalemc}{\Bar{\sigma}^c}
\newcommand{\size}{\mathbf{s}_i}
\newcommand{\scalei}{\sigma_i}
\newcommand{\sizemc}{\overline{\mathbf{s}}^c}

\newcommand{\meshi}{\mathbf{M}_i}

\newcommand{\vk}{v_k}
\newcommand{\tk}{\theta_k}
\newcommand{\pk}{p_k}
\newcommand{\ok}{o_k}
\newcommand{\fk}{f_k}

\newcommand{\net}{\Phi}
\newcommand{\fmap}{\mathbf{F}_{2D}}
\newcommand{\fmapmean}{\overline{\mathbf{F}}_{2D}}
\newcommand{\bgpred}{\mathbf{H}}
\newcommand{\bggt}{\mathbf{H}_{GT}}

\newcommand{\pcorr}{P(f|\theta_k)}
\newcommand{\pcorrk}{P(f|\theta_k, \kappa)}

\newcommand{\corr}{\mathcal{N}_{3D}^{2D}}
\newcommand{\mcorr}{\bar{\mathcal{N}}_{3D}^{2D}}

\usepackage{bm}
\usepackage{multirow}
\usepackage[accsupp]{axessibility} 

\usepackage{booktabs}
\usepackage{float}
\usepackage{pgfplots}
\pgfplotsset{compat=1.18}

\definecolor{iccvblue}{rgb}{0.21,0.49,0.74}
\usepackage[pagebackref,breaklinks,colorlinks,allcolors=iccvblue]{hyperref}

\def\paperID{12237} 
\def\confName{ICCV}
\def\confYear{2025}

\title{
Unified Category-Level Object Detection\\and Pose Estimation from RGB Images using 3D Prototypes
}
 
\author{Tom Fischer\thanks{Equal contribution. Corresponding email: {\tt\{tom.fischer, xiaojie.zhang\}@utn.de.}}\textsuperscript{\rm ~~1,2}\quad Xiaojie Zhang\footnotemark[1]\textsuperscript{\rm ~~1,2}\quad Eddy Ilg\textsuperscript{\rm 2}\\
\textsuperscript{\rm 1} Saarland University
\quad \textsuperscript{\rm 2} University of Technology Nuremberg 
}

\begin{document}
\maketitle
\begin{abstract}
Recognizing objects in images is a fundamental problem in computer vision. Although detecting objects in 2D images is common, many applications require determining their pose in 3D space. Traditional category-level methods rely on RGB-D inputs, which may not always be available, or employ two-stage approaches that use separate models and representations for detection and pose estimation.   
For the first time, we introduce a unified model that integrates detection and pose estimation into a single framework for RGB images by leveraging neural mesh models with learned features and multi-model RANSAC. Our approach achieves state-of-the-art results for RGB category-level pose estimation on REAL275, improving on the current state-of-the-art by 22.9\% averaged across all scale-agnostic metrics.
Finally, we demonstrate that our unified method exhibits greater robustness compared to single-stage baselines.
Our code and models are available at \href{https://github.com/Fischer-Tom/unified-detection-and-pose-estimation}{github.com/Fischer-Tom/unified-detection-and-pose-estimation}.

\end{abstract}    
\section{Introduction}

\begin{figure}
    \centering
    \includegraphics[width=1\linewidth,trim={6cm 0 4cm 0},clip]{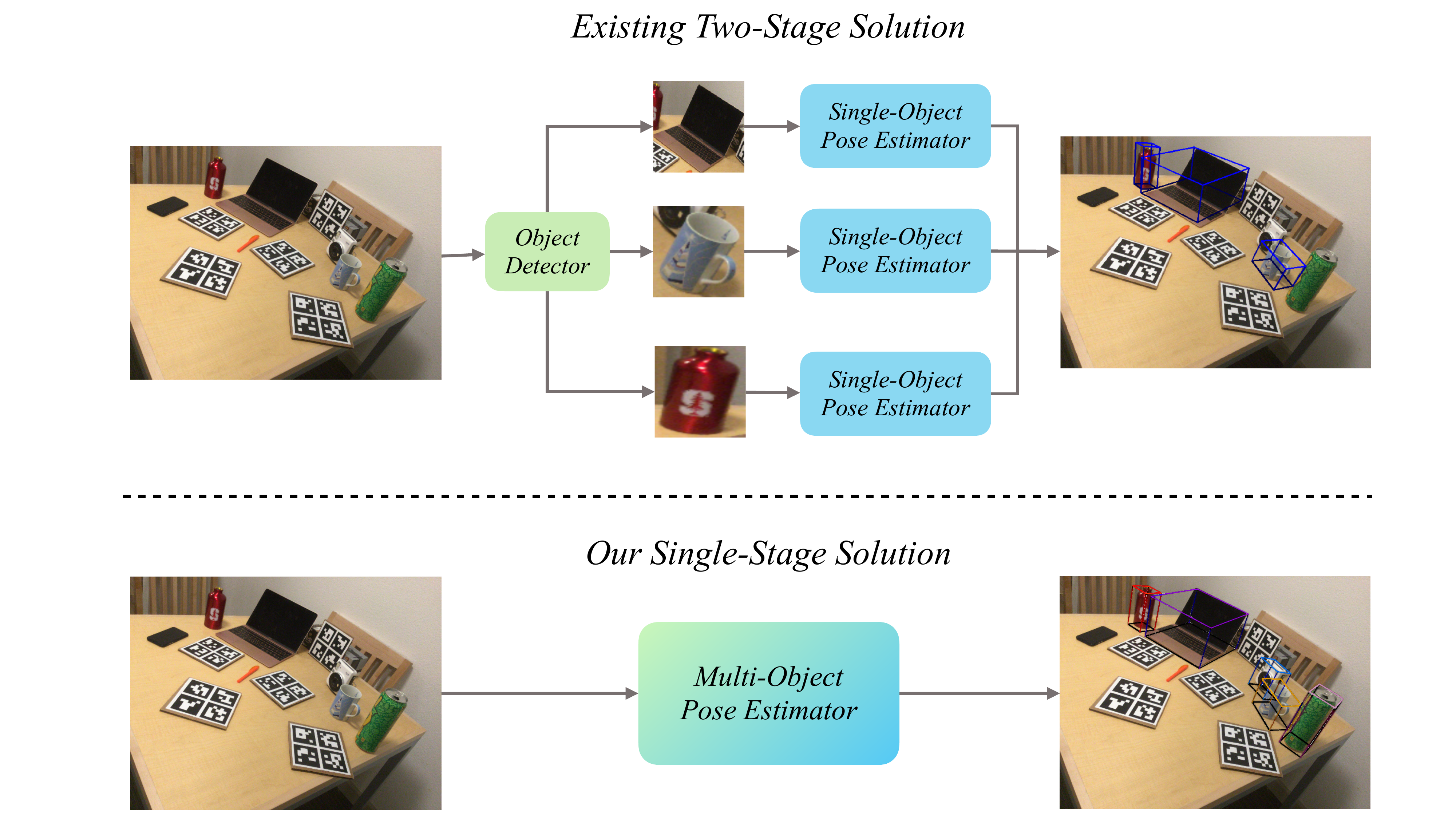}
    \caption{
    The top shows existing pipelines that are made up of separate stages for object detection and and single-object pose estimation. In the bottom, we show our pipeline that uses a single representation and model for detection and pose estimation.  
    }
    \label{fig:teaser}
\end{figure}

Finding out whether an object is present in an image, where it is located, and how it is oriented in space, are fundamental problems of scene understanding. They are commonly referred to as object detection and pose estimation, and are critically relevant for applications in robotics~\cite{kappler2018real,wen2022catgrasp,wen2022you}, autonomous driving~\cite{kothari2017pose,su2019deep} and augmented reality~\cite{marchand2015pose}. In object pose estimation, one distinguishes between methods that work only on known instances~\cite{li2018deepim,he2020pvn3d,cosypose,su2022zebrapose,zhou2023deep}, and methods that are able to generalize within object categories~\cite{nocs, secondpose, agpose, genpose, dualpose,oldnet,dmsr,lapose}. 

Category-level pose estimation has made significant progress in recent years~\cite{nocs, secondpose, agpose, genpose, dualpose,oldnet,dmsr,lapose}. However, most current work still relies on accurate depth information to recover precise poses~\cite{secondpose, genpose}, restricting the use to devices and working conditions where depth can be captured. 
Pose estimation from RGB images alone~\cite{msos, oldnet, dmsr, centerpose, lapose} is less explored and a much harder task. 

Methods can be further classified according to the used detection approach.
Although there are some RGB-D methods that utilize a shared representation with individual detection and pose heads~\cite{nocs, centersnap}, most methods completely decouple the problem and use separate models and representations.
All current RGB methods fall into this category of two-stage methods and use separate models per task, where as shown in Figure~\ref{fig:teaser}, objects are first detected, cropped, and then passed into a single-object pose estimator. This naturally places strong emphasis on the accuracy of the detection model, as failures cannot be recovered later, and comes with the downside of having to maintain two separate models.

For the first time, we show that a single unified model can be established for both tasks in the case of RGB images, and that this approach leads to improved accuracy and robustness. 
Our model leverages neural mesh models~\cite{nemo3d, nemo6d} as a 3D representation of object categories, where a 2D feature extractor for images and the neural features on the meshes are learned jointly during training. During inference, the image features are matched with features of the 3D meshes and multi-model RANSAC PnP~\cite{prog} is used to detect different object instances and estimate their poses. 

Our approach achieves a new state of the art and outperforms the current RGB-only state-of-the-art~\citep{lapose} in pose estimation by $22.9\%$ when averaged across all scale-agnostic metrics.
The results show that a common error source in two-stage pipelines is the detector and that our single-stage model is significantly more robust than its two-stage counterparts.  
In summary, our contributions are:

\begin{itemize}
\item We propose the first single-stage framework for category-level multi-object pose estimation from single RGB images. 
\item For the first time, we leverage neural mesh models jointly for object detection and pose estimation by combining them with multi-model RANSAC PnP and subsequent pose refinement. 
\item We achieve state-of-the-art performance on pose estimation benchmark REAL275 w.r.t. to accuracy and robustness, verifying the effectiveness of our object representation and single-stage multi-object inference pipeline.
\end{itemize}

\label{sec:intro}

\section{Related Work}
\label{sec:related work}

Both object detection~\cite{faster-rcnn,detr} and object pose estimation~\cite{survey} are long-standing tasks in computer vision. 
Although the first has converged to broadly accepted models, a variety of approaches are still being explored for the latter. 
Early pose estimation methods~\cite{choi20123d,vfh,birdal2015point} operate by matching hand-crafted features to instance-level 3D models. 
Today, deep learning has replaced feature descriptors, but many works are still restricted to estimate poses of object instances that were present during training~\cite{li2018deepim,he2020pvn3d,cosypose,su2022zebrapose,zhou2023deep}.

\textbf{Category-level pose estimation}~\cite{nocs, secondpose, agpose, genpose, lapose} was introduced by~\cite{nocs} to allow generalization to unseen instances within a set of established categories without requiring additional training or accurate CAD models during inference.
The strong variation in geometry and scale within categories makes this a highly challenging problem. 
Shape-prior-free methods~\cite{nocs,dualpose,agpose,secondpose} try to learn a generalizable representation without any guidance from CAD models during training.

Shape-prior-based methods, on the other hand, use CAD models and/or simple derived properties such as sizes or bounding boxes to integrate geometric priors.
The pose is then either regressed directly from a trained network, or solved for using predicted correspondences in the form of a Normalized Object Coordinate Space (NOCS)~\cite{nocs} with the (often deformed) shape prior using a non-differentiable alignment algorithm such as PnP~\cite{pnp, epnp} or the Umeyama algorithm~\cite{umeyama}. 
Our method falls into the above category, but estimates the correspondences from feature matching with 3D neural meshes.

\textbf{Neural Mesh Models} ~\cite{nemo3d} are shape-prior-based methods and were initially proposed for robust category-level single-object rotation estimation.
Follow-up work extended it to 6D pose estimation~\cite{nemo6d}, however, they still relied on single-object training datasets and both methods~\cite{nemo3d, nemo6d} are constrained to per-category networks that have to be calibrated against each other or require a (multi-label) classification model to identify present object categories.
Later works removed the need for separate feature representations to allow robust classification, but they cannot estimate translations or handle multiple objects~\cite{novum, inemo}.
Our work removes all these restrictions and can both detect objects and estimate their poses with a single, shared representation.

\begin{figure*}
    \centering
    \includegraphics[width=\linewidth]{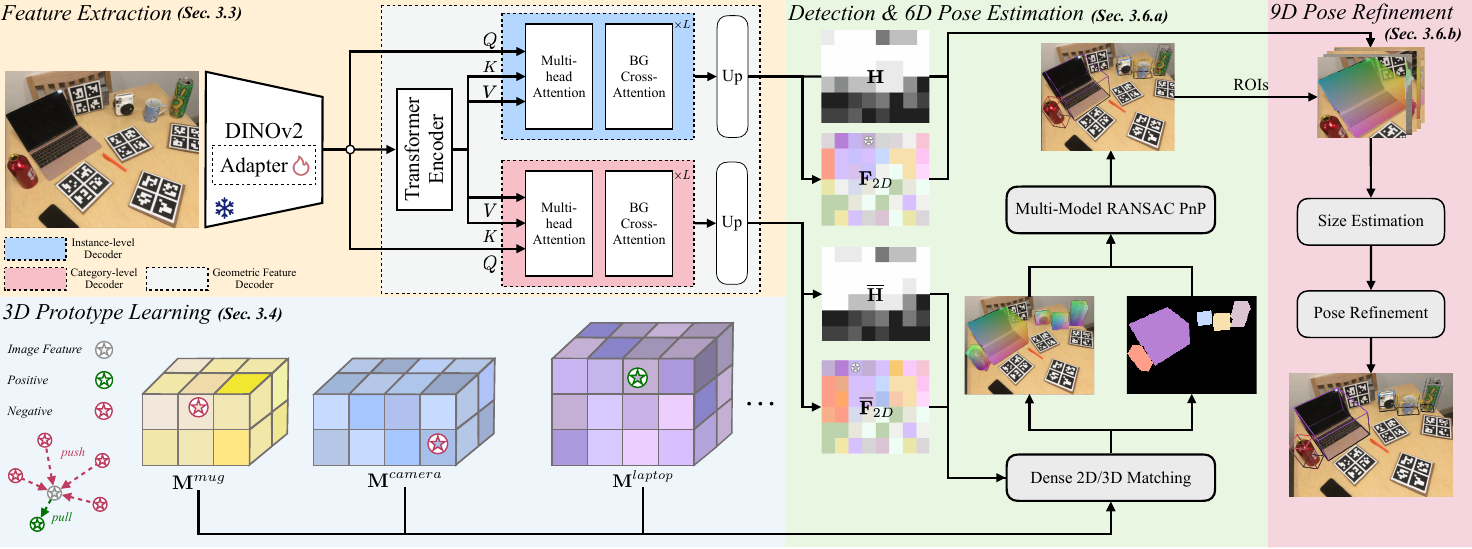}
    \caption{We present a unified framework for solving detection and pose estimation with 3D object prototypes.
    \textit{Feature Extraction} (\cref{sec: FE}): Given an RGB image, a frozen DINOv2 model with trainable adapters is combined with a Geometric Feature Decoder to produce two feature maps $\fmapmean, \fmap$ and foreground segmentations $\Bar{\bgpred}, \bgpred$.
    \textit{3D Prototype Learning} (\cref{sec: protolearn}): During training, the Adapters and the Geometric Feature Decoder are jointly trained with a set of Neural Mesh Models~\cite{nemo3d} using a contrastive loss.
    \textit{Detection \& 6D Pose Estimation} (\cref{sec: det} a): After training, 6D object poses are obtained with a multi-model fitting algorithm from dense 2D/3D correspondences between category-level features $\fmapmean$ and all vertex features.
    \textit{9D Pose Refinement} (\cref{sec: 9d} b): For each identified instance, a deformation parameter is optimized for by aligning correspondences from instance-level features $\fmap$ with its 6D pose.
    }
    \label{fig:framework}
\end{figure*}

\subsection{Two-Stage Methods}

The vast majority of object pose estimation methods first locates the objects using off-the-shelf detection models~\cite{maskrcnn, faster-rcnn} to detect Regions of Interest (RoI) as 2D segmentation masks or bounding boxes.
The objects are then cropped from the image and processed by a second model to predict the final poses, splitting the approaches into two stages.

\textbf{RGB-D Approaches} have access to the incomplete point clouds of the observed instances derived from the depth map. 
These can be used for regression in combination with~\cite{dpdn} or without~\cite{agpose, socs, secondpose} shape priors, or to align them with a shape prior~\cite{spd,sgpa} using the Umeyama algorithm~\cite{umeyama}.
In contrast to the above, our method does not require depth as input.

\textbf{RGB-Only Approaches} face a more challenging problem due to the inherent ambiguity in this setting, which is why these methods are less prominent. 
In particular, the lack of depth information introduces a fundamental ambiguity between distance and scale, making it impossible to recover these properties from observations.
To this end, Zhang \textit{et al.}~\cite{lapose} introduced a set of scale-agnostic metrics that we also utilize in our work.
OLD-Net~\cite{oldnet} and DMSR~\cite{dmsr} leverage a shape prior-based alignment technique and LaPose~\cite{lapose} uses a shape prior-based regression approach.
MSOS~\cite{msos}, on the other hand, does not utilize any shape prior and directly estimates NOCS shape and the metric-scale shape of the object for downstream alignment. Although these two-stage approaches are successful in practice, they learn separate representations for both stages and are prone to errors of the first stage. In contrast to the above, we present a single-stage method with higher accuracy and robustness. 

\subsection{Single-Stage Methods}

Single-stage methods couple detection and pose estimation into an end-to-end architecture.
Existing work on end-to-end pose estimation focuses mainly on the instance-level setting~\cite{tekin2018real,hu2020single,efficientpose,cope,dpod,posecnn,segdriven,pointposenet}. 
However, there are category-level methods that utilize a shared representation and task-specific heads.
NOCS~\cite{nocs} uses a shared feature extraction backbone that provides ROIs paired with individual heads for NOCS-map and instance segmentation.
The category-level approaches~\cite{centersnap, shapo} require RGB-D as input and are shape-prior-based regression methods that perform joint center-point detection, reconstruction, and pose estimation.
CenterPose~\cite{centerpose} comes close to an RGB-only single-stage method, but requires the class labels of the objects present in an image and therefore is not a stand-alone method. 
In contrast, our work proposes the first RGB-only single-stage approach to unify the complete detection and pose estimation in a shared representation.

\section{Method}
\label{sec:method}

We present a single-stage framework for RGB-only object detection and category-level 9D pose estimation.
Our method aligns image features with a set of 3D object prototypes to infer object properties from 2D/3D correspondences obtained from feature matching. 
An overview is shown in \cref{fig:framework}.

\subsection{Task Definition}

Given an input image $\mathbf{I} \in \mathbb{R}^{H \times W \times 3}$ and camera intrinsics $\mathbf{K} \in \mathbb{R}^{3 \times 3}$, our goal is to estimate the set of object properties:
\begin{align}
 \mathcal{O}(\mathbf{I}, \mathbf{K}) = \{(c_i, \size, \mathbf{R}_i, \mathbf{t}_i)\}^N_{i=1} \, ,
\end{align}
where each object $i$ is described by category $c_i$, 3D size $\mathbf{s}_i \in \mathbb{R}^3$, rotation $\mathbf{R}_i \in SO(3)$, and translation $\mathbf{t}_i \in \mathbb{R}^3$. 
We define its scalar instance scale as $\scalei = \lVert\mathbf{s}_i\rVert_2$.
Throughout the remaining parts of this section, we refer to sets in script font (i.e. $\mathcal{O}$);  vectors, tuples, and tensors in bold font (i.e. $\mathbf{R}$); and scalars in standard math font (i.e. $c$).

\subsection{Representation}

We represent 3D object prototypes with Neural Mesh Models~\cite{nemo3d, inemo, novum, nemo6d}, which have shown high generalization and robustness for single-object pose estimation.
Each category $c$ has a prototypical mesh $\meshc = (\vertc, \facec, \texc)$, where $\vertc \in \R^{V \times 3}$ are vertex coordinates in object space, $\facec\in\Z^{A\times 3}$ are triangle indices, and $\texc \in \R^{V \times D}$ are learned vertex features. 
The geometry of each mesh is obtained by regularly sampling the surface of the bounding box obtained from the average size $\sizemc$ of the training set, which we normalize to have $\lVert\sizemc\rVert_2=1$.
We also keep a scaling factor $\scalemc$ that transforms the mesh into one that has the mean scale of the category.
If camera poses are available, scenes can be composed from the object prototypes and rendered into an image.

\subsection{Feature Extraction}
\label{sec: FE}

Accurate object localization from 2D/3D correspondences requires that the object geometry aligns well with the visual observations.
In category-level pose estimation, one does not have access to instance-level geometries.
To address this, we train a 2D feature extractor $\net$ that learns to map the input image $\mathbf{I}$ to a rendering of a scene composed of cuboid prototypes parameterized by $\mathcal{O}(\mathbf{I}, \mathbf{K})$.

\textbf{Modeling.}
As solving for 9D poses from correspondences alone is challenging and tends to fail in practice, we first perform an initial 6D pose estimation, where we rely on a feature map $\fmapmean$ that is aligned with the prototypes that follow category-level sizes $\sizemc$.
Next, we perform a size refinement with a feature map $\fmap$, which is aligned with deformed prototypes that match the instance-level sizes $\size$.
Since both feature maps are conditioned on the same neural textures, $\fmapmean$ can be imagined as a warped variant of $\fmap$. 

\textbf{Backbone.}
The feature extractor $\net$ is trained jointly with the vertex features.
This introduces ambiguities that make a good initialization crucial to guide each $\texc$ towards a meaningful representation.
Therefore, we build $\net$ on top of the self-supervised pre-trained DINOv2 ViT~\cite{dinov2} and embed task-specific information using Parameter-Efficient Fine-Tuning (PEFT)~\cite{peft}.
Following~\cite{adaptformer}, we use a low-rank adaptation (Adapter) in the MLP in each transformer block:
\begin{equation}
    x'_l = \text{MLP}(\text{LN}(x_l)) + x_l + \lambda(\text{GeLU}(\text{LN}(x_l) \cdot \bm{D})) \cdot \bm{U},
\end{equation}
where $x_l$ is the output of the previous multi-head attention, $\lambda$ is a scaling factor~\cite{adaptformer} and $\bm{D},\bm{U}$ are the down- and up-projection matrices.

\textbf{Geometric Feature Decoder.} 
To obtain the features corresponding to the object prototype geometry, we first apply a shared transformer encoder to the backbone features, and subsequently two transformer decoders to obtain $\fmapmean, \fmap$. 
Finally, both feature maps are upsampled to a certain output resolution using a combination of bilinear upsampling and shared convolution layers.

\subsection{Training}
\label{sec: protolearn}

We consider a training sample with $N$ objects and outline the loss calculation for an object $i$ with prototypical mesh $\meshi = (\verti, \facei, \texi)$, intrinsic matrix $\mathbf{K}$, and pose $\hat{\mathbf{R}}_i, \hat{\mathbf{t}}_i$.

\textbf{Setup.}
We first obtain pixel-level training annotations from $\verb|rasterize|(\meshi, \mathbf{K}, \hat{\mathbf{R}}_i, \hat{\mathbf{t}}_i)$ to obtain projected image location $\pk$ and binary visibility $\ok$ for each vertex $\vk\in\verti$.
Given a 2D feature map $\mathbf{F}$, we define the annotation set as $\mathcal{Y}(\meshi, \mathbf{F})=\{(\tk, \fk, \ok) \mid \forall \vk \in\verti\}$, where $\fk$ is the pixel feature at projection $\pk$ in $\mathbf{F}$, and $\tk$ is the corresponding vertex feature from $\mathbf{F}_{3D}$.

\textbf{Optimizing $\net$.}
Following~\cite{nemo3d}, we model the probability distribution of any feature $f$ being generated from vertex $\vk$ by defining $\pcorr$ using a von Mises-Fisher (vMF) distribution to express the likelihood:
\begin{equation}
    \label{eq:vmf}
    \pcorrk=C(\kappa)e^{\kappa(f^\top\cdot \tk)},
\end{equation}
with mean $\tk$, concentration parameter $\kappa$, and normalization constant $C(\kappa)$.

Accurate object poses are obtained when $\fk$ can be matched to $\tk$, leading to precise 2D/3D correspondences.
We optimize for this task using contrastive learning by maximizing \cref{eq:vmf} with the feature $\fk$, and minimize the likelihood of all other vertices:
\begin{align}
    &\max \hspace{0.85cm} P(\fk|\tk, \kappa), \label{eq:contrastiveproblem1}\\
    &\min \sum_{\theta_m\in\overline{\theta}_k}P(\fk|\theta_m, \kappa), \label{eq:contrastiveproblem2}
\end{align}
where the negatives $\overline{\theta}_k$ are the features of all non-corresponding vertices including those from different classes.
Equations \ref{eq:contrastiveproblem1} and \ref{eq:contrastiveproblem2} can be combined into a single loss by taking the negative log-likelihood:
\begin{equation}
\label{eq: permeshloss}
    \mathcal{L}(\mathcal{Y}(\meshi, \mathbf{F}))=-\sum_k \ok\cdot \log \left( \frac{e^{\kappa(\fk\cdot \tk)}}{\sum_{\theta_m\in\overline{\theta}_k}e^{\kappa(\fk^\top\cdot \theta_m)}}\right),
\end{equation}
where considering $\kappa$ as a global hyperparameters allows canceling out the normalization constants $C(\kappa)$.
We compute the full training loss as:
\begin{equation}
    \label{eq: loss}
    \mathcal{L}_{train} = \sum_i \frac{\mathcal{L}(\mathcal{Y}(\mathbf{M}^{c_i}, \fmapmean))+\mathcal{L}(\mathcal{Y}(\phi(\mathbf{M}^{c_i}), \fmap))}{2N},
\end{equation}
where $\phi(\mathbf{M})=(\phi(\verti), \facei, \texi)$ is the deformation operation we discussed in \cref{sec: FE}.
For mean and deformed meshes we scale the vertices with ground-truth scale $\scalei$ before rasterizing.

\textbf{Optimizing Vertex Features.}
Previous work~\cite{coke, nemo3d, nemo6d, inemo} advocates for an exponential moving average strategy when updating vertex features.
We found that this approach is sensitive to initialization and hyperparameters that lead to highly varying performance between training runs.
For a more stable and consistent training, we optimize them through backpropagation with $\net$ using the same loss in \cref{eq: loss}.

\subsection{Foreground Detection}
Although the contrastive loss in \cref{eq: loss} helps localize objects implicitly, background regions still produce false-positive matches. 
Inspired by the emergent behavior of foreground segmentation in ViT attention maps~\cite{deepVIT}, we extract foreground masks $\overline{\mathbf{H}},\mathbf{H}$ using cross-attention between foreground tokens and decoder features.

We initialize two foreground tokens  $\overline{\bm{b}},\bm{b}$ with the $\textbf{CLS}-$token of the backbone ViT.
After each decoder layer or upsampling step, the per-pixel affinity to the foreground token is measured via cross attention:
\begin{equation}
    \bm{b}^{i+1}, \bm{W}^{i+1} = \text{CrossAttn}(\bm{b}^i, F^i),
\end{equation}
where $F^i$ are the current image features, $\bm{b}^{i+1}$ is the updated foreground token, and $\bm{W}^{i+1}$ are the attention weights.
At the final level, we use min-max normalization to obtain the segmentation $\mathbf{H}=\verb|minmax|(\bm{W}^N)$. 
The mean-shape segmentation $\overline{\mathbf{H}}$ is predicted in the same way.  

We supervise the final segmentations with masks obtained from rendering the prototypes using ground-truth poses.
Since it is a popular choice for segmentation~\cite{mask2former,maskrcnn} and we have found practical success with it, we adopted the dice loss~\cite{dice} to optimize the attention weights:
\begin{equation}
    \mathcal{L}_{mask}=\frac{\mathcal{L}_{dice}(\bgpred, \bggt)+\mathcal{L}_{dice}(\overline{\bgpred}, \overline{\mathbf{H}}_{GT})}{2}.
\end{equation}

\subsection{Inference}
\label{sec:inference}

During inference time, we determine the objects contained in an image $\mathbf{I}$ using the trained feature extractor $\net$ and the prototypes through correspondences from feature matching. 
First, $\net$ is queried for the feature maps $\fmapmean, \fmap$ and the foreground heatmaps $\overline{\bgpred},\bgpred$, which are thresholded using a confidence parameter $t_1$ to obtain binary foreground segmentations. 
Then, two dense correspondence mappings $\mcorr, \corr$ are established between the prototypes and the feature maps.
Considering $\fmapmean$, the set of correspondences is established as: 

\begin{equation}
\begin{aligned}
\mcorr = \{ (\bm{p}_i, v_k, \rho(v_k)) \;\mid\;
\bm{p}_i \in \fmapmean,\\
v_k = \argmax_{\tk \in \bigcup_{c} \texc} f_i^\top \tk \},
\end{aligned}
\end{equation}

where the membership function $\rho(\vk)=c$ returns the category label $c$ of the mesh to which the matched vertex $\vk$ belongs.
We remove noisy correspondences which fall outside of $ \overline{\bgpred}$ or have a similarity score below a second confidence threshold $t_2$.

\textbf{a) Detection and 6D Pose Estimation.} 
\label{sec: det}
We use the correspondence set $\mcorr$ to simultaneously detect objects and estimate their initial 6D pose.
If it is known a-priori that there is only a single instance per category in the scene, one could group correspondences by category label and solve this problem via sequential application of PnP~\cite{epnp} paired with RANSAC~\cite{ransac}. 
However, this assumption is rarely valid in practice.
To handle scenes with multiple instances from the same category, we use Progressive-X (ProgX)~\cite{prog}, a multi-model fitting algorithm that enables PnP to detect multiple object poses.
ProgX extends Graph-Cut RANSAC~\cite{GCRansac} to the multi-model setting via a sequential hypothesis generation algorithm that can separate correspondences into individual instances.
For each class $c$, we run ProgX on the subset $\bar{\mathcal{N}}^c\subset\mcorr$ to obtain a set of 6D poses:
\begingroup\makeatletter\def\f@size{9}\check@mathfonts
\def\maketag@@@#1{\hbox{\m@th\normalsize\normalfont#1}}%
\begin{equation}
\mathcal{P}_{6D} = \bigcup_{c} \left\{ (\overline{\mathbf{R}}, \bar{\mathbf{t}}, c) \mid \forall (\overline{\mathbf{R}}, \bar{\mathbf{t}}) \in \text{MM-PnP}(\overline{\mathcal{N}}^c) \right\},
\end{equation}
\endgroup

\textbf{b) 9D Pose Refinement.}
\label{sec: 9d}
For each detected object $i$, we refine the initial pose $(\overline{\mathbf{R}}_i, \bar{\mathbf{t}}_i, c)\in\mathcal{P}_{6D}$ to account for instance-specific size and improve their accuracy using $\corr$.
We select a subset $\mathcal{N}_i\subset\corr$, where 1) the 2D point lies within $\bgpred$, 2) its similarity score is above the threshold $t_2$, and 3) its 2D coordinate was considered an inlier according to the 6D pose and category-level geometry.  

We deform the category-level mesh by scaling vertices along the principal axes with a deformation multiplier $\mathbf{d}\in\R^3$.
Refinement is done by minimizing the reprojection error:
\begingroup\makeatletter\def\f@size{9}\check@mathfonts
\def\maketag@@@#1{\hbox{\m@th\normalsize\normalfont#1}}%
\begin{equation}
\bm{E}(\mathbf{K}, \mathbf{R}, \mathbf{t}, \mathbf{d}, \mathcal{N}_i) = \sum_{v_k,\bm{p}_k\in\mathcal{N}_i} \lVert (\mathbf{K}\mathbf{R} (\mathbf{d} \odot \vk) + \mathbf{t}) - \bm{p}_k \rVert_2^2.
\end{equation}
\endgroup
The optimization proceeds in two steps, where the deformation is obtained using the 6D rotation and translation, which are subsequently refined using the deformation:
\begin{align}
\mathbf{d}_i &= \min_{\mathbf{d}} \bm{E}(\mathbf{K}, \overline{\mathbf{R}}_i, \bar{\mathbf{t}}_i, \mathbf{d}, \mathcal{N}_i) \\
\mathbf{R}_i, \mathbf{t}_i &= \min_{\mathbf{R}, \mathbf{t}} \bm{E}(\mathbf{K}, \mathbf{R}, \mathbf{t}, \mathbf{d}_i, \mathcal{N}_i).
\end{align}
The final pose of object $i$ is then given as $(\mathbf{R}_i, \mathbf{t}_i, \mathbf{s}_i, c_i)$, where $\mathbf{s}_i$ is obtained as the size of the bounding box of $\mathbf{V}^{c_i}$ multiplied with $\mathbf{d}_i$.

\section{Experiments}
\label{sec:experiments}

\begin{table*}[t]
    \centering
    \begin{tabular}{@{}c|c c c|c c c c c c c c@{}}
    \toprule
        Method & $NIoU_{25}$ & $NIoU_{50}$ & $NIoU_{75}$ & $5^\circ 0.2d$  & $5^\circ 0.5d$  & $10^\circ 0.2d$ & $10^\circ 0.5d$ & $0.2d$ & $0.5d$ & $5^\circ$ & $10^\circ$\\ \hline
        MSOS ~\cite{msos} & 36.9 & \hphantom{0}9.7 & \hphantom{0}0.7 & - & - &\hphantom{0}3.3 & 15.3 & 10.6 & 50.8 & - & 17.0 \\ 
        OLD-Net ~\cite{oldnet} & 35.4 & 11.4 & \hphantom{0}0.4 & \hphantom{0}0.9 & \hphantom{0}3.0 & \hphantom{0}5.0 & 16.0 & 12.4 & 46.2 & \hphantom{0}4.2 & 20.9  \\
        DMSR ~\cite{dmsr} & 57.2 & 38.4 & \hphantom{0}9.5 & 15.1 & 23.7  & 25.6 & 45.2 & 35.0 & 67.3 &  27.4 & 52.0\\ 
        {LaPose} ~\cite{lapose}  & 70.7 & 47.9 & 15.8 & 15.7 &  21.3& 37.4 & 57.4 & 46.9 & 78.8 & 23.4 & 60.7\\ 
        \hline
        {Ours} & \textbf{75.2} & \textbf{53.7} & \textbf{19.2} & \textbf{25.1}& \textbf{31.8}  & \textbf{43.7} &\textbf{66.1} & \textbf{53.5} & \textbf{83.7} &  \textbf{32.1} & \textbf{68.8}\\
        \bottomrule
    \end{tabular}
    \caption{Comparison with state-of-the-art methods on REAL275 using scale-agnostic evaluation metrics. As visible, we set a new state of the art by outperforming the baseline methods in all metrics. }
    \label{tab:main_real}
\end{table*}

\begin{table*}[t]
    \centering
    \begin{tabular}{@{}c|c c c|c c c c c c c c@{}}
    \toprule
        Method & $NIoU_{25}$ & $NIoU_{50}$ & $NIoU_{75}$ & $5^\circ 0.2d$  & $5^\circ 0.5d$  & $10^\circ 0.2d$ & $10^\circ 0.5d$ & $0.2d$ & $0.5d$ & $5^\circ$ & $10^\circ$\\ \hline
        MSOS \cite{msos} & 35.1 & \hphantom{0}9.9 & \hphantom{0}0.8 & - & - &\hphantom{0}5.9 & 31.6 & \hphantom{0}8.9 & 47.2 & - & 48.6 \\ 
        OLD-Net \cite{oldnet} & 48.7 & 19.5 & \hphantom{0}1.7 & \hphantom{0}4.7 & 15.9 & 12.9 & 39.6 & 19.1 & 60.0 & 20.5 & 50.5\\
        DMSR \cite{dmsr} &73.6 & 45.1 & 11.1 & 26.1 & 43.6 & 38.2 & 67.5 & 42.5 & 79.6 & 47.3 & 74.2\\ 
        {LaPose\cite{lapose}} &\textbf{74.1} &45.2&12.5&29.4&53.9&39.2 &74.4 & 41.2 &80.2 & 58.5 &\textbf{82.7} \\ 
        \hline
        {Ours} & 73.9 & \textbf{47.6} & \textbf{18.1} & \textbf{37.2} & \textbf{57.6} & \textbf{45.2} & \textbf{75.5} & \textbf{47.6} & \textbf{83.4} & \textbf{60.5}& 80.8 \\
        \bottomrule
    \end{tabular}
    \caption{Comparison with state-of-the-art methods on CAMERA25 using scale-agnostic evaluation metrics.
    We assume that the lower $10^\circ$ performance is due to small objects in CAMERA25, yielding less correspondences.
    Our method achieves the highest accuracy on 9 out of the 11 metrics.}
    \label{tab:main_camera}
\end{table*}

\begin{table}[t]
    \centering
    \resizebox{\columnwidth}{!}{
    \begin{tabular}{@{}c|c c|c c c c@{}}
    \toprule
        Method & $IoU_{50}$ & $IoU_{75}$ & $5^\circ 5cm$  & $5^\circ 10cm$  & $10^\circ 5cm$ & $10^\circ10cm$\\ 
        \hline
        MSOS ~\cite{msos} & 13.6 &  \hphantom{0}1.0 & - & - & - & 11.8 \\ 
        OLD-Net ~\cite{oldnet} & \hphantom{0}8.1 & \hphantom{0}0.3 & \hphantom{0}0.5 & \hphantom{0}2.0 & \hphantom{0}2.8 & 10.4 \\
        DMSR ~\cite{dmsr} & 15.9 & \hphantom{0}\textbf{3.3} & \hphantom{0}6.1 & 13.4 & 10.5 & 25.0\\ 
        {LaPose} ~\cite{lapose} & \textbf{17.5} & \hphantom{0}2.6 & \hphantom{0}6.3 & 13.6 & \textbf{12.5}& \textbf{30.5} \\ 
        \hline
        {Ours} & \textbf{17.5} &  \hphantom{0}2.6 &  \hphantom{0}\textbf{7.2} & \textbf{17.7} & 10.4 & 29.4 \\
        \bottomrule
    \end{tabular}}
    \caption{Results on REAL275 using absolute metrics. Predicting object scales from RGB images alone is challenging and not having 2D bounding boxes during inference exacerbates this problem, leading to worse performance of our trained scale prediction network, and consequently less precise metric scale translations. 
    Nevertheless, our method is competitive in performance, either matching or outperforming the baselines in 3 out of the 6 considered metrics.
    \label{tab:abs_real}
}
\end{table}

In the following, we explain our experimental setup and then discuss the results of our single-stage pose estimation approach on real, synthetic, and corrupted data.
For further results and more detailed comparisons, we refer the reader to our supplemental material.

\subsection{Experiment Setup}

\textbf{Datasets.}
We evaluate our method on the popular category-level 9D pose estimation benchmarks CAMERA25~\cite{nocs} and REAL275~\cite{nocs}.
REAL275 is a real-world dataset that consists of 4.3K training images of 7 scenes and 2.75K testing images of 6 scenes.
CAMERA25 supplements the real data with 275K synthetic training images and 25K evaluation images of rendered objects with real-world backgrounds.
Both datasets contain objects from six categories: bottle, bowl, camera, can, laptop, and mug.

\textbf{Evaluation Metrics.} 
Following previous work ~\cite{lapose}, we use scale-agnostic evaluation metrics to remove the scale-depth ambiguity. 
To evaluate 3D detection and object size estimation, we report the mean average of the Normalized 3D Intersection over Union (NIoU) metric at thresholds 25\%, 50\%, and 75\%. 
For pose estimation, we individually report the rotation error and scale-normalized translation error at the different thresholds $0.2d$, $0.5d$, $5^\circ$, $10^\circ$, and on joint thresholds $5^\circ 0.2d$, $5^\circ 0.5d$, $10^\circ 0.2d$, $10^\circ 0.5d$. 
In addition, we also report results on absolute metrics for comparable thresholds.
Our work focuses on joint detection and pose estimation; therefore, we do not remove objects with low detection thresholds as in previous work~\cite{nocs}, but evaluate performance on all detected objects as proposed by~\cite{lapose}.

\textbf{Implementation Details.} To train our model on the REAL275 dataset, we use a mixture of 25\%
real-world images and 75\% synthetic images from the CAMERA25 training set, similar to ~\cite{nocs}.
For the results on CAMERA25, we follow prior work~\cite{lapose, dmsr, msos, oldnet} and train a separate model without the REAL275 images.

For each baseline, we use the detection results from~\cite{spd} for a fair comparison and reproduce the numbers with the public checkpoints of their methods where possible.
Notably, MSOS~\cite{msos} does not have code publicly available at the time of writing, which is why we use the numbers provided by~\cite{lapose}.
For more implementation details about our method, we refer the reader to our supplementary material.

\subsection{Comparison with State-of-the-Art Methods}

\textbf{Results with Scale-Agnostic Metrics.}
In \cref{tab:main_real}, we compare our method with state-of-the-art RGB-only category-level pose estimation methods on the REAL275 benchmark under scale-agnostic evaluation metrics. 
As visible, our method consistently outperforms all baselines under scale-agnostic metrics.

As can be seen in \cref{tab:main_camera}, our method also shows strong performance on synthetic data. While it falls behind on coarse rotation accuracy, it still outperforms baselines in 9 out of 11 metrics.
A likely reason is that CAMERA25 contains many smaller objects (high depth or small scale), and performance could likely be increased by using higher-resolution features at the cost of slower inference speed.

\begin{table*}[ht]
    \centering
    \resizebox{\textwidth}{!}{%
     \begin{tabular}{@{}c|c|c c c|c c c c c c c c@{}}
         \toprule
        Source&Method & $\Delta NIoU_{25}$ & $\Delta NIoU_{50}$ & $\Delta NIoU_{75}$ & $\Delta5^\circ 0.2d$  & $\Delta5^\circ 0.5d$  & $\Delta10^\circ 0.2d$ & $\Delta10^\circ 0.5d$ &$\Delta0.2d$ & $\Delta0.5d$ &  $\Delta5^\circ$ & $\Delta10^\circ$ \\ \hline

        \multirow{3}{*}{\rotatebox[origin=c]{90}{ROI}}
        &OLD-Net\cite{oldnet} &-14.4\% &-24.6\% &-50.0\% &-44.4\% &-36.7\% &-28.0\% &-23.8\% &-16.1\% &-10.8\% &-23.8\% &-18.2\%  \\
        &DMSR\cite{dmsr} & -\hphantom{0}\underline{3.3}\% &-\hphantom{0}\underline{7.3}\% &-\underline{13.7}\% &-\underline{17.9}\% &-\underline{18.1}\% &-\hphantom{0}\underline{8.2}\% &-\underline{10.4}\% &-\hphantom{0}\underline{4.6}\% &-\hphantom{0}\underline{2.7}\% &-\underline{17.5}\% &-\underline{11.5}\% \\
        &LaPose\cite{lapose}  & -\hphantom{0}9.3\% &-16.9\% &-18.4\% &-28.7\% &-23.9\% &-25.4\% &-16.6\% &-18.3\% &-\hphantom{0}6.2\% &-21.4\% &-14.3\%  \\
        \hline
        
        \multirow{4}{*}{\rotatebox[origin=c]{90}{Image}}
        &OLD-Net\cite{oldnet} & -28.0\% &-38.6\% &-50.0\% &-55.6\% &-43.3\% &-46.0\% &-41.3\% &-34.7\% &-26.4\% &-23.8\% &-24.9\%  \\
        &DMSR\cite{dmsr} & -13.8\% &-14.3\% &-22.1\% &-24.5\% &-27.8\% &-16.4\% &-22.1\% &-13.1\% &-15.2\% &-26.6\% &-22.5\%  \\
        &LaPose\cite{lapose} & -22.9\% &-24.6\% &-22.8\% &-29.3\% &-28.6\% &-29.4\% &-27.7\% &-25.6\% &-21.2\% &-26.1\% &-25.9\%   \\
        &{Ours} & -\textbf{11.0}\% &-\textbf{14.0}\% &-\textbf{21.7}\% &-\textbf{17.2}\% &-\textbf{13.5}\% &-\textbf{16.0}\% &-\textbf{12.9}\% &-\textbf{12.7}\% &-\hphantom{0}\textbf{9.5}\% &-\textbf{13.5}\% &-\textbf{12.6}\% \\
        \bottomrule
    \end{tabular}
    }
    \caption{Robustness experiments on the REAL275~\cite{nocs} benchmark. 
    We report loss in accuracy compared to the model evaluated on uncorrupted data averaged over $8$ types of corruptions.
    The top shows results for corruption-free detection with only corrupted crops during pose estimation and the bottom shows corruptions during both stages.
    Corruptions during pose estimation alone already reduce the performance significantly. 
    However, the on the image level the two-stage approaches suffer even more, indicating that failures in the detection are severe. 
    Our single-stage approach is much less affected by noise and shows the greatest robustness.
    \label{tab:ablation_noise_relative}
    }
\end{table*}

\begin{table*}[ht]
    \centering
    \begin{tabular}{@{}c|c c c|c c c c c c c c@{}}
    \toprule
        Detection Source & $NIoU_{25}$ & $NIoU_{50}$ & $NIoU_{75}$ & $5^\circ 0.2d$  & $5^\circ 0.5d$  & $10^\circ 0.2d$ & $10^\circ 0.5d$ & $0.2d$ & $0.5d$ & $5^\circ$ & $10^\circ$\\ \hline 
                GT Object Mask & 72.1 &50.3 & 17.8 & 23.0 & 29.6 & 40.4 & 61.4 & 50.4 & 81.0 & 29.9 & 63.7 \\
        GT Proto. Mask  & \textbf{79.5} & \textbf{55.3} & \textbf{19.5} & \textbf{25.6} & \textbf{32.2} & \textbf{44.1} & \textbf{66.7} & \underline{53.4} & \textbf{84.2} & \textbf{32.5} & \textbf{69.2} \\ 
        \hline
        
        Mask-RCNN~\cite{maskrcnn}  & 70.4 & 48.8 & 17.7 & 22.4 & 28.7 & 38.9 & 58.4 & 48.9 & 79.1 & 28.9 & 60.4  \\ 
        None (ours) & \underline{75.2} & \underline{53.7} & \underline{19.2} & \underline{25.1} & \underline{31.8} & \underline{43.7} & \underline{66.1} & \textbf{53.5} & \underline{83.7} &  \underline{32.1} & \underline{68.8}\\   

        \bottomrule
    \end{tabular}
    \caption{To isolate the pose estimation accuracy of our method, we frame the problem as single-object pose estimation by providing different ROIs. 
    As visible, our method comes close to the upper bound (GT Proto. Mask), showing our model can detect objects well and simultaneously reason about their poses with high accuracy.
    Our method also works well on the tight "GT Object Masks" but the performance drops significantly when using the fine-tuned Mask-RCNN, indicating that our approach outperforms it. 
    \label{tab:ablation_detection}}
\end{table*}

\begin{figure*}
    \centering
    \renewcommand{\arraystretch}{1.2}  
    \setlength{\tabcolsep}{5pt}  

    \begin{tabular}{*{5}{c}}  
        \includegraphics[width=0.18\textwidth]{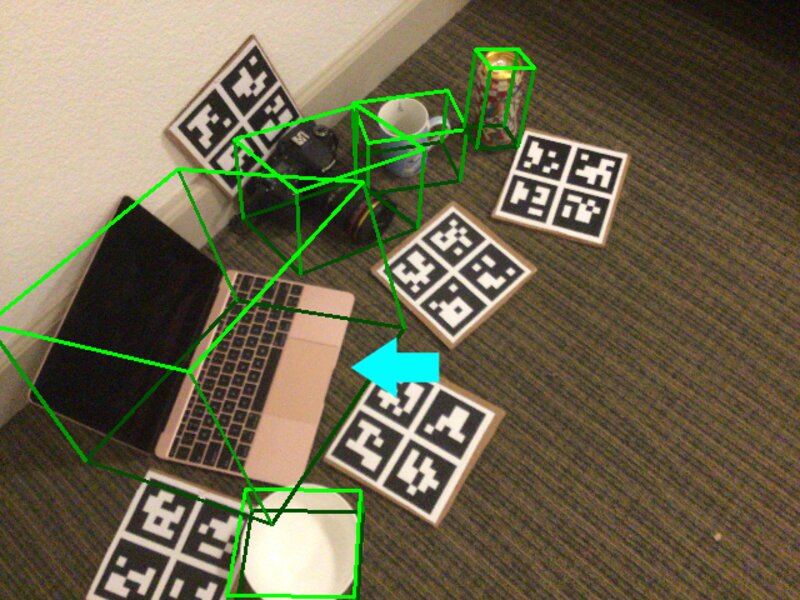} &
        \includegraphics[width=0.18\textwidth]{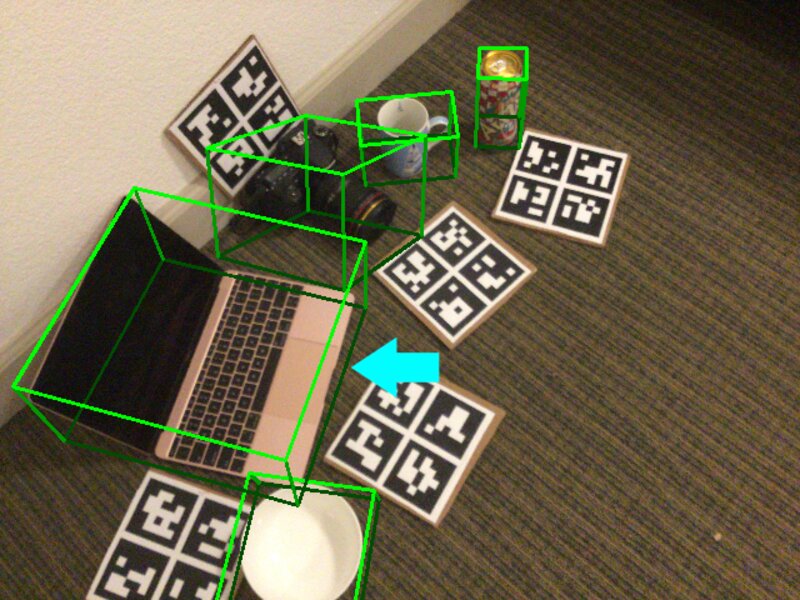} &
        \includegraphics[width=0.18\textwidth]{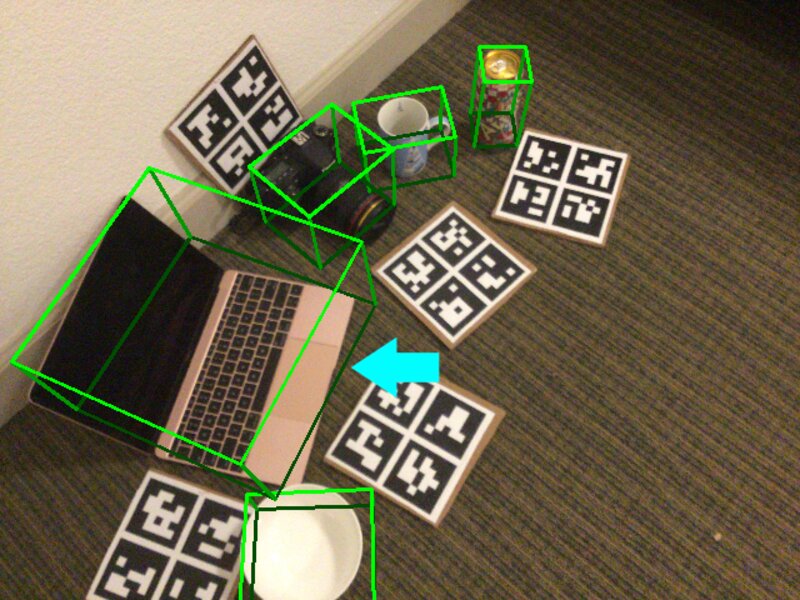} &
        \includegraphics[width=0.18\textwidth]{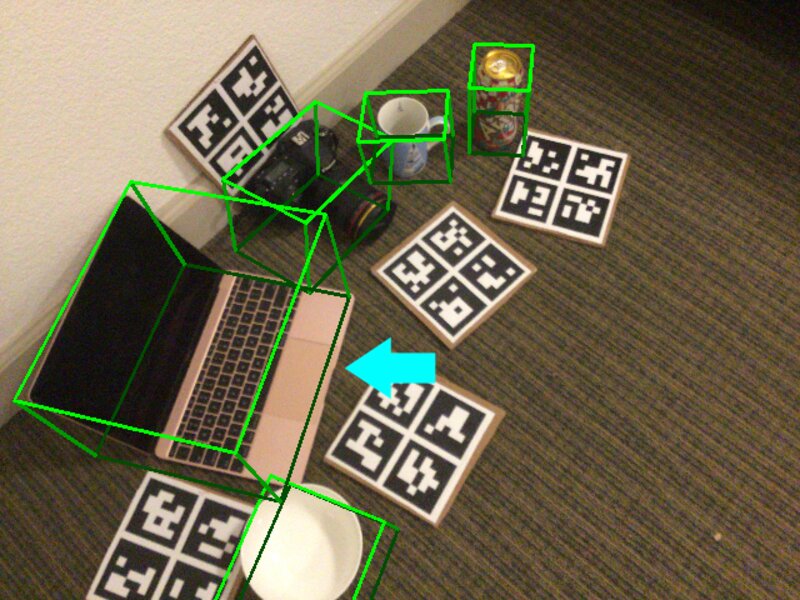} &
        \includegraphics[width=0.18\textwidth]{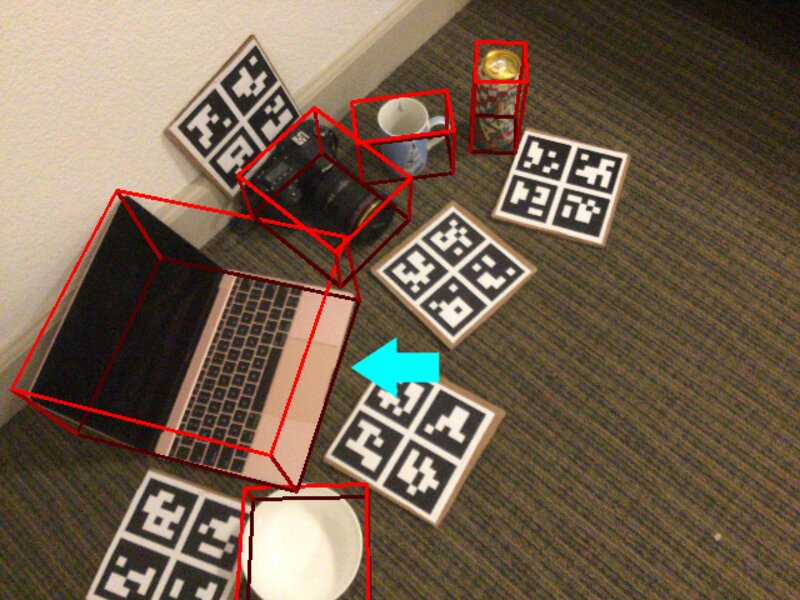} \\
        
        \includegraphics[width=0.18\textwidth]{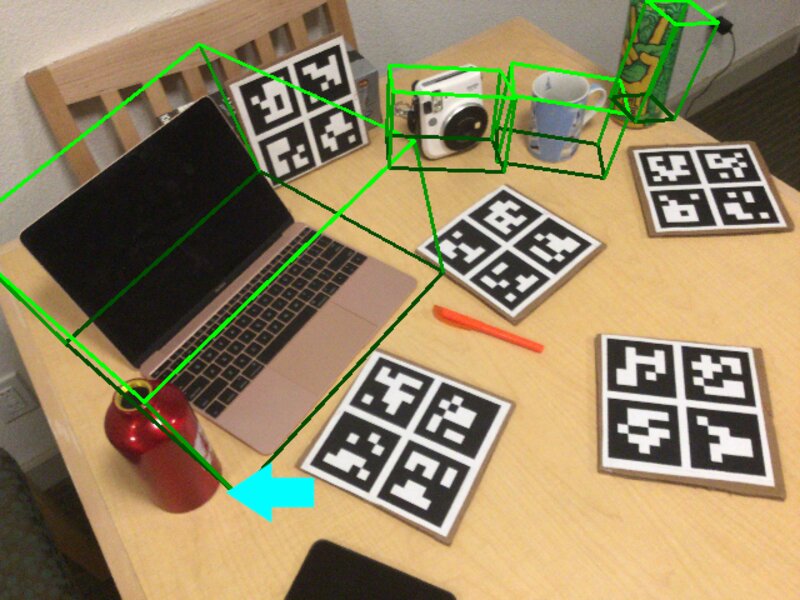} &
        \includegraphics[width=0.18\textwidth]{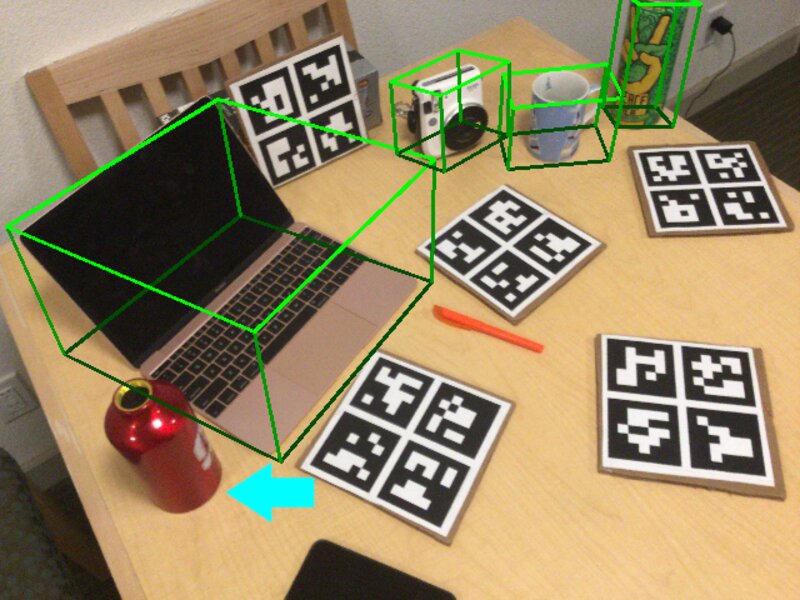} &
        \includegraphics[width=0.18\textwidth]{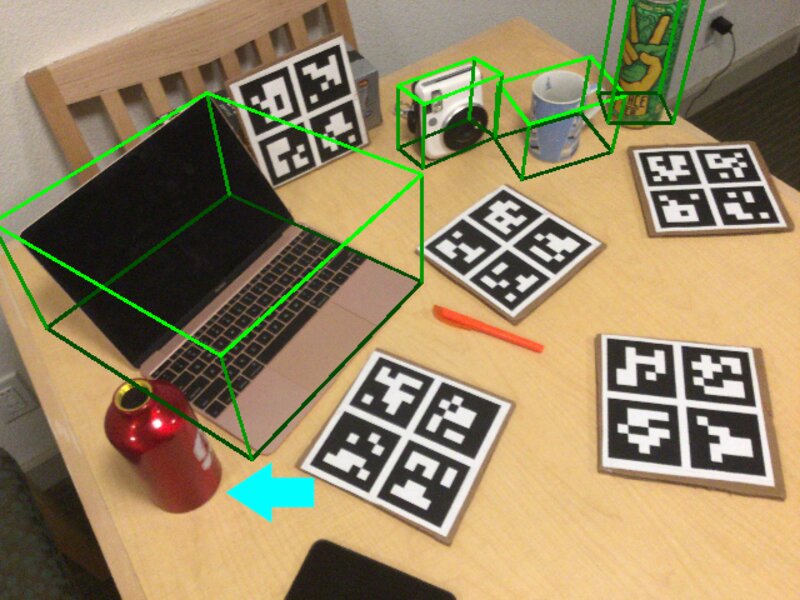} &
        \includegraphics[width=0.18\textwidth]{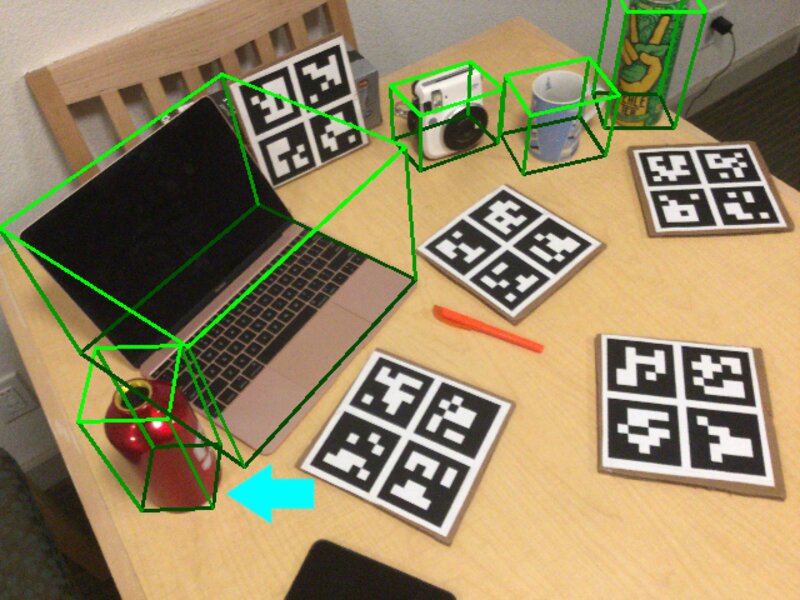} &
        \includegraphics[width=0.18\textwidth]{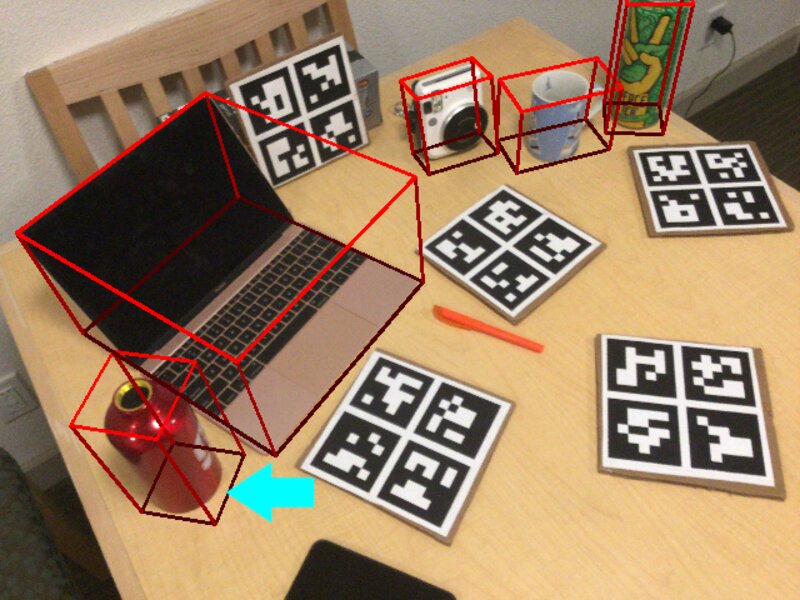} \\
        
        \includegraphics[width=0.18\textwidth]{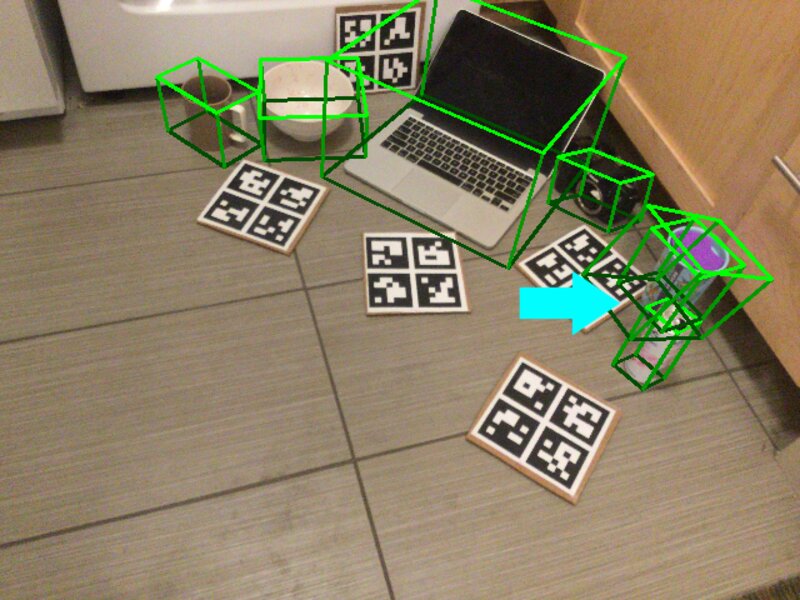} &
        \includegraphics[width=0.18\textwidth]{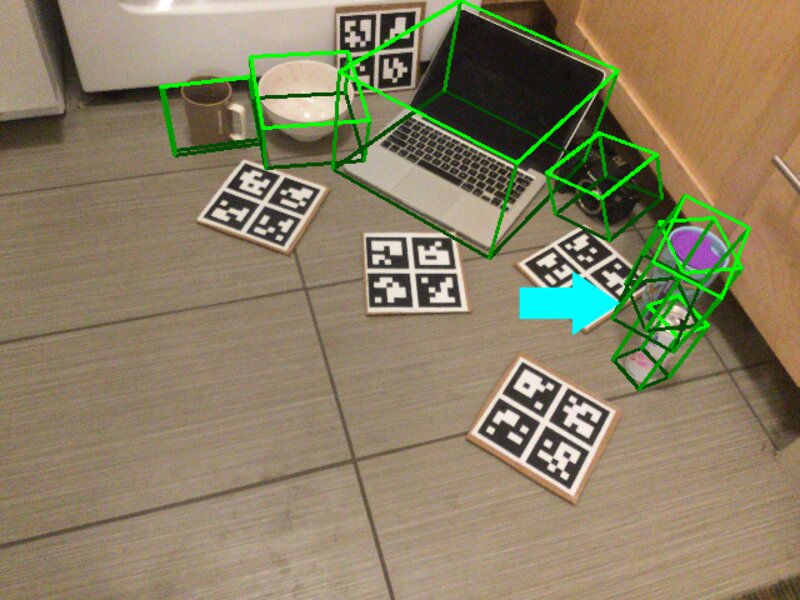} &
        \includegraphics[width=0.18\textwidth]{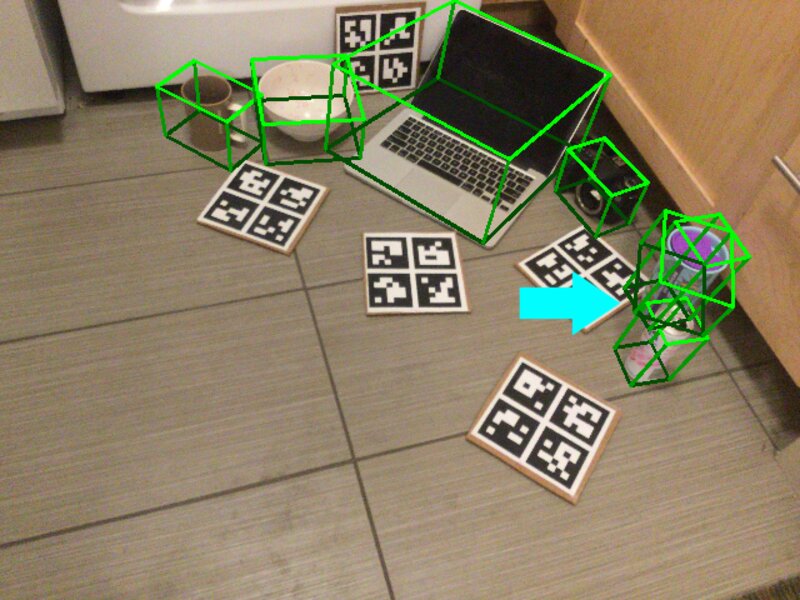} &
        \includegraphics[width=0.18\textwidth]{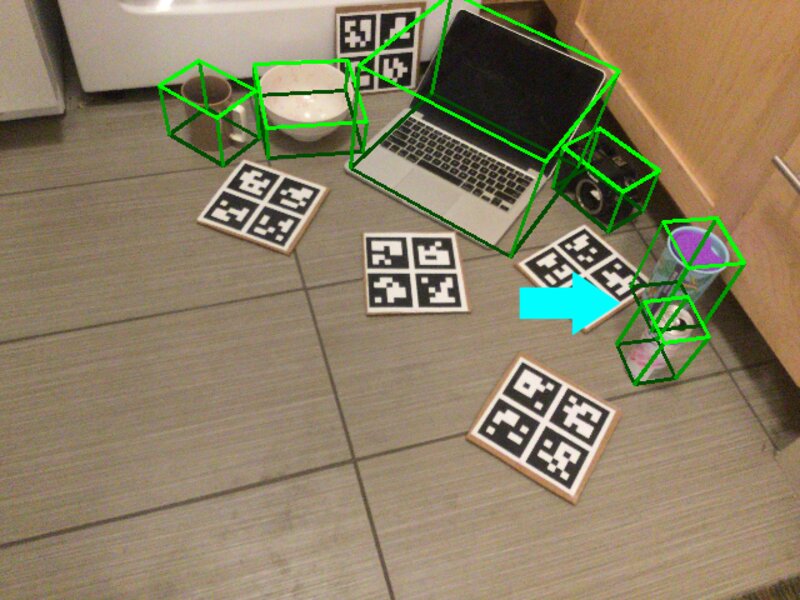} &
        \includegraphics[width=0.18\textwidth]{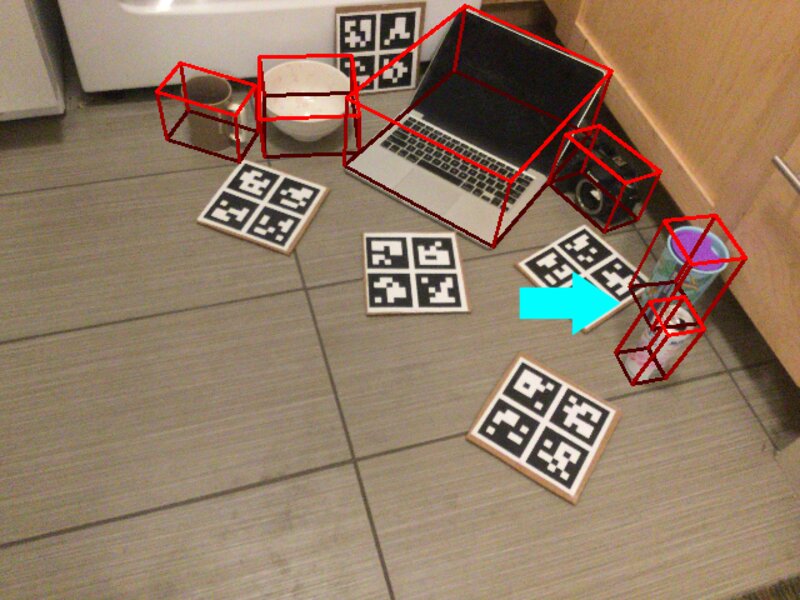} \\
        
        \includegraphics[width=0.18\textwidth]{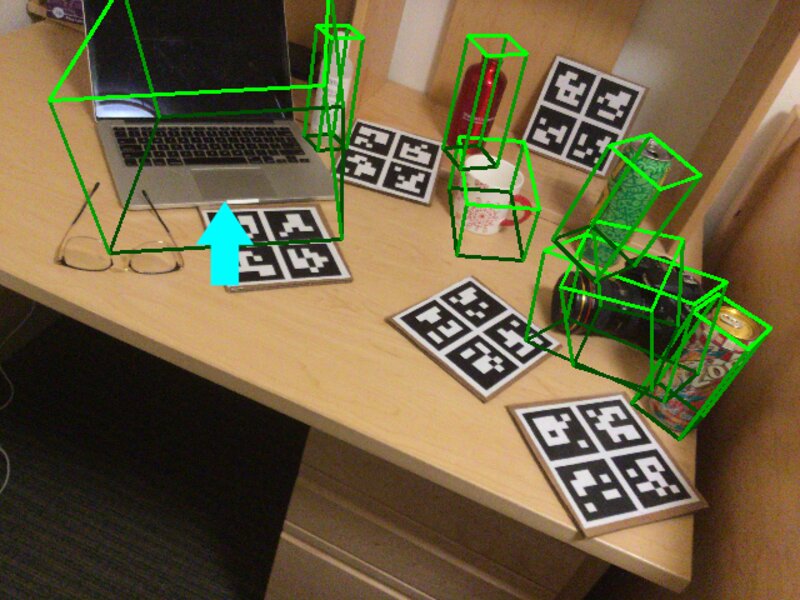} &
        \includegraphics[width=0.18\textwidth]{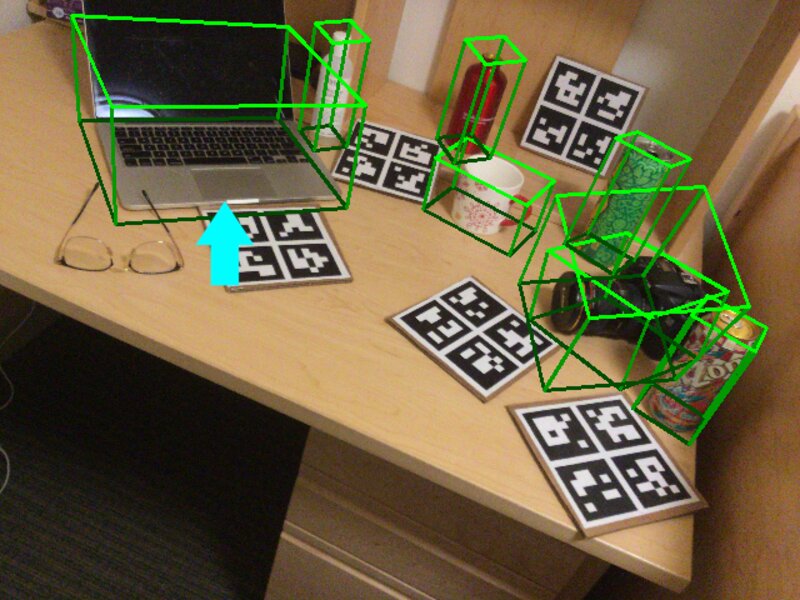} &
        \includegraphics[width=0.18\textwidth]{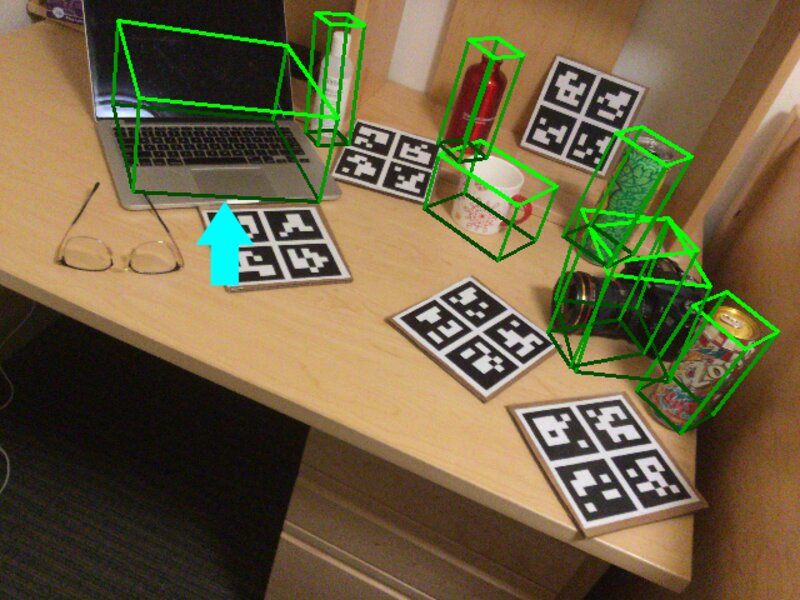} &
        \includegraphics[width=0.18\textwidth]{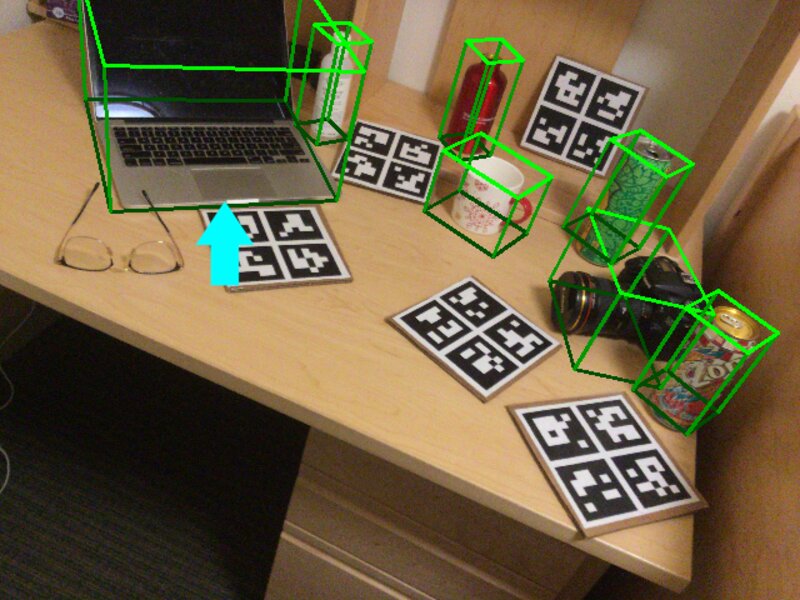} &
        \includegraphics[width=0.18\textwidth]{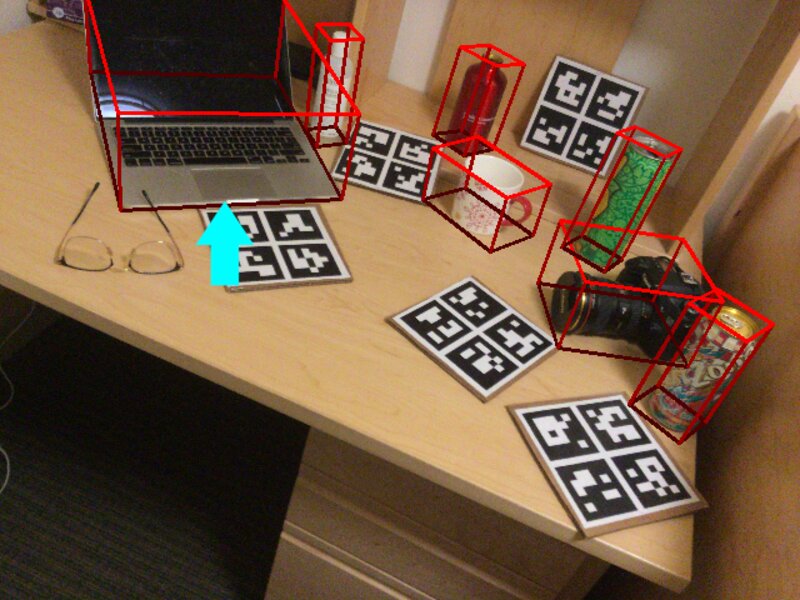} \\
        
        \includegraphics[width=0.18\textwidth]{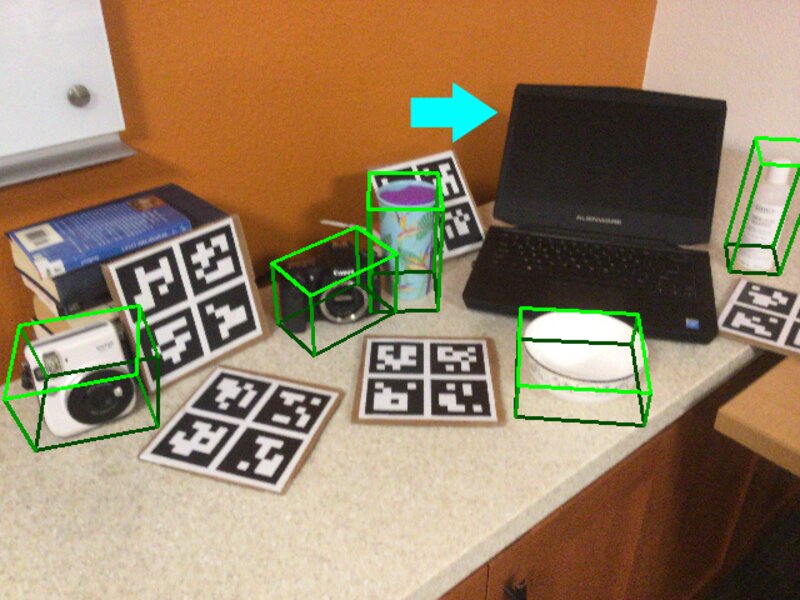} &
        \includegraphics[width=0.18\textwidth]{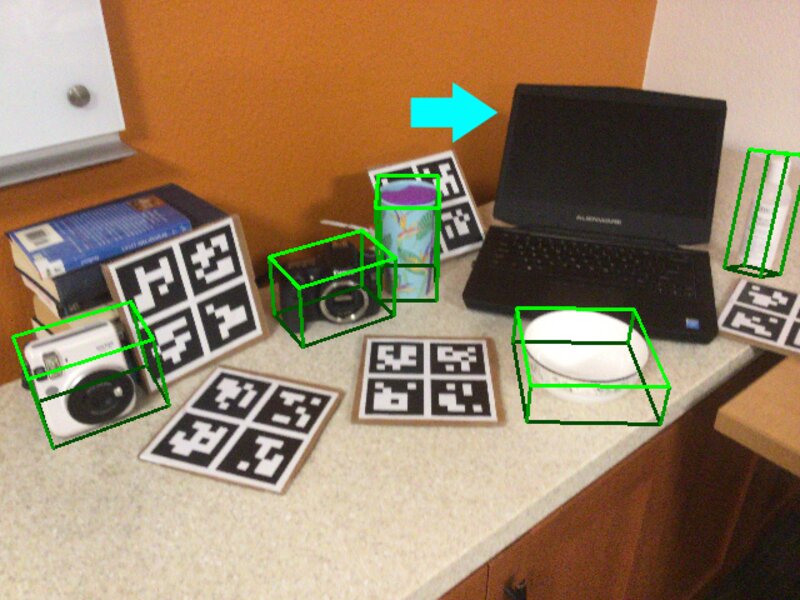} &
        \includegraphics[width=0.18\textwidth]{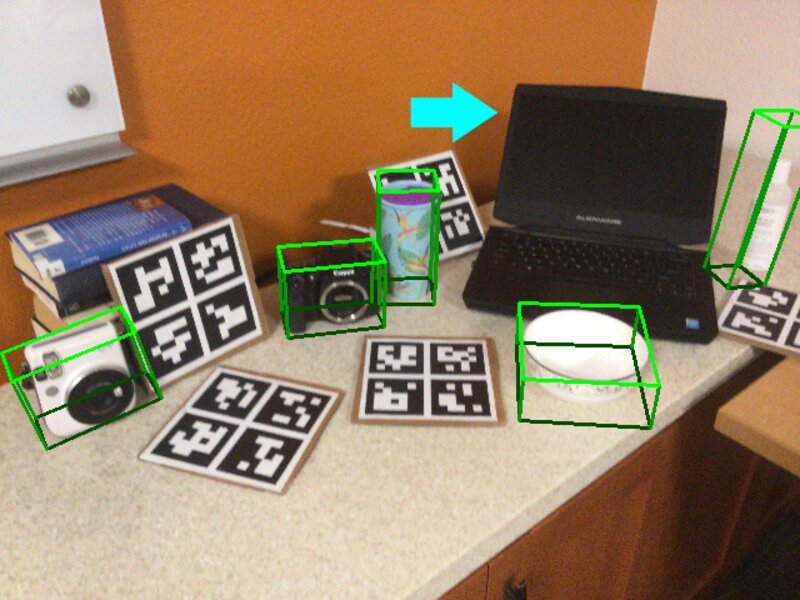} &
        \includegraphics[width=0.18\textwidth]{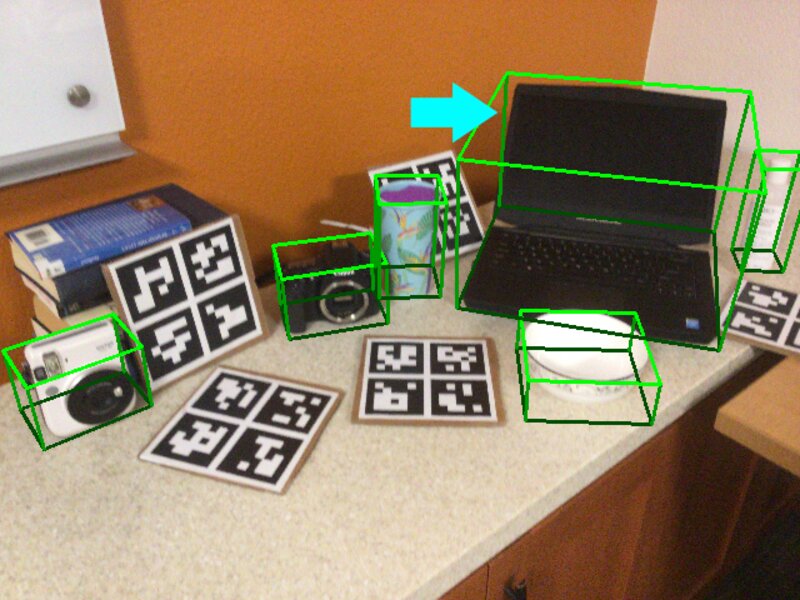} &
        \includegraphics[width=0.18\textwidth]{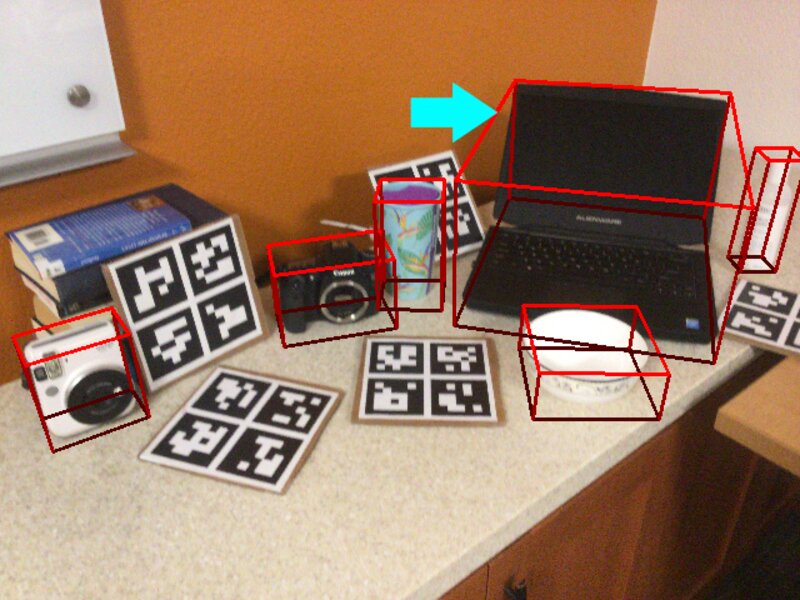} \\

        \multicolumn{1}{c}{{OLD-Net}} &
        \multicolumn{1}{c}{{DMSR}} &
        \multicolumn{1}{c}{{LaPose}} &
        \multicolumn{1}{c}{{Ours}} &
        \multicolumn{1}{c}{{Ground Truth}} \\
    \end{tabular}

    \caption{Qualitative comparison on REAL275\cite{nocs}. We compare our model (second to last column) with all baselines and with ground truth (last column).
    Some  failures due to poor detection results  of the baselines are highlighted (row 2, 3, and 5).
    In rows 1 and 4 our method outperforms baselines in terms of rotation accuracy. 
    \label{fig:qualitative}
    }
\end{figure*}

\textbf{Results with Absolute Metrics.}
To measure pose estimation quality under metric scale, we follow~\cite{lapose} and decouple scale prediction from pose estimation.
Since we do not use a dedicated detection model, we cannot use their pre-trained model.
Instead, we train the same scale network using their code to predict a per-instance scale parameter, given the 2D projection of our object prototype with the estimated pose.

We show the performance of our method in combination with the scale prediction network \cref{tab:abs_real}.
Our results are competitive due to the strong performance in the normalized space.
However, we could not reproduce the same level of performance from\cite{lapose} for the scale estimation network.
We believe that this is due to the network that was trained on 2D bounding boxes being more robust during inference than our version, which was trained on the prototype projections.

\textbf{Qualitative Results.}
Finally, \cref{fig:qualitative} shows estimated poses on REAL275 of our method and the baselines.
As can be seen, two-stage approaches suffer from the error propagation of detection failures, whereas our method identifies objects much more confidently.
Additionally, our method significantly improves the rotation accuracy which is most notable in the laptop category.
For more detailed per-category results, we refer the reader to our supplemental material.

\subsection{Robustness to Image Degradations}

\textbf{Setting.} To verify the robustness of two-stage vs. single-stage pose estimation, we conduct experiments on the REAL275 dataset with added corruptions from the imagecorruptions~\cite{imagecorruptions} library.
We choose $8$ corruptions of different types and report the average performance in the main part of the paper.
For the choice of corruptions and per-corruption performance, we refer the reader to the supplemental material.
For consistency, each method is evaluated using the same set of images and has been trained with the same data augmentations.

We design two sets of experiments to study the robustness of the two-stage baselines.
To measure the robustness of the pose estimator in isolation, we run the detector on clean images and apply the corruption only on its provided RoI.
To evaluate the robustness of the whole framework, we apply corruptions on an image level.
As it is the most commonly used model, we use the pre-trained Mask-RCNN~\cite{maskrcnn} from~\cite{spd} to get detection results for all baselines.
Note that we omit MSOS~\cite{msos} from the robustness experiments, since to the best of our knowledge their code is not (or no longer) public.

\textbf{Results.}
We report the decrease in performance under image corruptions for the full task at the bottom of~\cref{tab:ablation_noise_relative}.
Our performance drops by only $14\%$ when considering the mean and median on all metrics.
In comparison, baselines suffer much more, losing anywhere from $19\%-37\%$ mean or $22\%-38\%$ median performance. 

\subsection{Pose Estimation Quality}

By framing detection and pose estimation as a single-stage problem, it is hard to quantify the success of either component individually.
To measure the quality of our pose estimation in isolation, we try to decouple our framework into two stages.
We do so by obtaining masks from three different sources.
First, we use ground-truth object segmentation masks.
Second, we use projected object prototype masks using ground-truth 9D poses.
Third, we use bounding boxes provided by the same fine-tuned Mask-RCNN~\cite{maskrcnn} as our baselines.
The masks are then used to identify per-object correspondences for single-object pose estimation.
To be consistent with our main method, we also use Progressive-X~\cite{prog} as the solver, but we limit the maximum of predicted models to one per RoI.

\cref{tab:ablation_detection} shows the results of providing RoIs to our method.
It is important to note that using instance-level detections from either Mask-RCNN or the ground-truth segmentation mask is not ideal for our method as it is optimized to work with the projected prototype masks.  
However, we find that our method also works well with the tightly fitting ground-truth segmentation masks (first row) but the performance drops steeply when using the predicted Mask-RCNN masks. 
The upper bound is shown in the second row, which provides ground-truth detections and 2D correspondences to our method.
As can be seen, our method (fourth row) only falls noticeably behind in the $NIoU_{25}$ and performs close to the upper bound on all other metrics. 
This shows that our joint approach outperforms Mask-RCNN in detection performance.


\section{Conclusion}
\label{sec:conclusion}

In this work, we introduced the first unified RGB-only category-level detection and pose estimation framework that operates with a single representation.
Using neural mesh models as 3D object prototypes leads to robust feature representations that can be used for 2D/3D correspondence estimation and downstream multi-model pose estimation to jointly solve the detection and pose estimation tasks. 
We achieve state-of-the-art results for pose estimation on the popular REAL275 benchmark, and our evaluation has shown that our single-stage model is significantly more robust to image degradations than its two-stage counterparts.



\section*{Acknowledgements}
We gratefully acknowledge the stimulating research environment of the GRK 2853/1 “Neuroexplicit Models of Language, Vision, and Action”, funded by the Deutsche Forschungsgemeinschaft (DFG, German Research Foundation) under project number 471607914.
The authors gratefully acknowledge the scientific support and HPC resources provided by the Erlangen National High Performance Computing Center (NHR@FAU) of the Friedrich-Alexander-Universität Erlangen-Nürnberg (FAU) under the BayernKI project \verb|b266be|. BayernKI funding is provided by Bavarian state authorities.

{
    \small
    \bibliographystyle{ieeenat_fullname}
    \bibliography{main}
}

\newpage

\maketitlesupplementary
\appendix

\section{Further Implementation Details}

In this section, we provide additional details on the training regimen of our method.

\textbf{Parameters.}
We train the model on a single NVIDIA H100 GPU for 150k iterations using the \emph{AdamW}~\cite{AdamW} optimizer with weight decay of $0.05$ and a Cosine Annealing schedule~\cite{annealing} that decays the learning rate from $1e-4$ to $1e-7$.
As training and inference-specific hyperparameters, we choose $\kappa=\frac{1}{0.07}$, $t_1=0.5$, and $t_2=0.7$.
The adapters are included in each transformer block of the DINOv2~\cite{dinov2} model with a low-rank dimensionality of $128$.

\textbf{Training Data.}
To improve generalization, we use a similar data augmentation strategy during training as~\cite{secondpose}. 
In each training iteration, we use a batch size of 10 images which are stacked into dynamic batches according to the present objects. 

\textbf{Annotation Generation.}
We use PyTorch3D~\cite{pytorch3d} to rasterize the object prototypes into the annotation set.
For each object, we render the given pose individually and then extract the per-pixel annotations.
To account for objects occluding each other (which happens frequently for the category-level feature map) we mask out pixels where more than one object is visible by setting the visibility to $0$.
Alternatively, one could also opt for supervising with the closest object.
However, we found that this results in the model to miss small, or partially visible objects more frequently.

\begin{table*}[hbt!]
    \centering
    \resizebox{\textwidth}{!}{%
     \begin{tabular}{@{}c|c c c|c c c c c c c c@{}}
        \toprule
        Method & $ NIoU_{25}$ & $ NIoU_{50}$ & $ NIoU_{75}$ & $5^\circ 0.2d$  & $5^\circ 0.5d$  & $10^\circ 0.2d$ & $10^\circ 0.5d$ & $0.2d$ & $0.5d$ & $5^\circ$ & $10^\circ$\\ \hline

        Ours & \hphantom{-}75.2 & \hphantom{-}53.7 & \hphantom{-}19.2 & \hphantom{-}25.1 & \hphantom{-}31.8 & \hphantom{-}43.7 & \hphantom{-}66.1 & \hphantom{-}53.5 & \hphantom{-}83.7 & \hphantom{-}32.1 & \hphantom{-}68.8 \\ \hline
        w/o adapter  & -16.4 & - 21.0 & -15.9 & -19.7 & -22.1 & -25.8 & -31.1 & -21.6 & -8.7 & - 21.8 & -30.8 \\
        \hline
        w/o CA - $t_1=0,t_2=0.7$  & -\hphantom{0}3.8 & -\hphantom{0}9.4\ & -\hphantom{0}4.8 & -\hphantom{0}7.7 & -\hphantom{0}4.0&-\hphantom{0}7.3 & -\hphantom{0}4.5 & -\hphantom{0}8.7 & -\hphantom{0}0.1 &  -\hphantom{0}3.4 & -\hphantom{0}3.8 \\
        w/o CA - $t_1=0,t_2=$opt. & +\hphantom{0}0.8 & -\hphantom{0}4.9 & -\hphantom{0}2.6 & -\hphantom{0}5.7 & -\hphantom{0}1.9&-\hphantom{0}5.6 & -\hphantom{0}2.9 & -\hphantom{0}6.3 & -\hphantom{0}0.3 &  -\hphantom{0}1.7 & -\hphantom{0}2.9 \\
        w CA - $t_1=0, t_2=0.7$ & -\hphantom{0}1.4 & -\hphantom{0}4.9 & -\hphantom{0}3.0 & -\hphantom{0}4.2 & -\hphantom{0}1.9 & -\hphantom{0}4.5&-\hphantom{0}2.6 & -\hphantom{0}5.4 & +\hphantom{0}0.2 & -\hphantom{0}1.8 &  -\hphantom{0}2.1  \\
        w CA - $t_1=0, t_2=$opt. & +\hphantom{0}0.6 & -\hphantom{0}0.9 & -\hphantom{0}0.7 & -\hphantom{0}0.9 & -\hphantom{0}0.3 & -\hphantom{0}1.9&-\hphantom{0}1.2 & -\hphantom{0}1.7 & -\hphantom{0}0.2 & -\hphantom{0}0.3 &  -\hphantom{0}1.2  \\
        
        \hline

        w/o refinement + w/o size  & -\hphantom{0}0.7 & -\hphantom{0}2.7 & -\hphantom{0}3.9 & -\hphantom{0}2.1 & -\hphantom{0}2.1&-\hphantom{0}2.4 & -\hphantom{0}3.9 & -\hphantom{0}2.1\ & -\hphantom{0}1.0 &  -\hphantom{0}1.8 & -\hphantom{0}3.5 \\

        w refinement & -\hphantom{0}0.3 & -\hphantom{0}2.4 & -\hphantom{0}2.8 & -\hphantom{0}1.3 & -\hphantom{0}1.1&-\hphantom{0}1.9 & -\hphantom{0}2.7 & -\hphantom{0}1.7\ & -\hphantom{0}0.8 &  -\hphantom{0}0.8 & -\hphantom{0}2.2 \\
        
        w size estimation & -\hphantom{0}1.0 & -\hphantom{0}2.2 & +\hphantom{0}0.1 & -\hphantom{0}2.4 & -\hphantom{0}2.4&-\hphantom{0}2.5 & -\hphantom{0}3.9 & -\hphantom{0}2.2\ & -\hphantom{0}0.9 &  -\hphantom{0}2.2& -\hphantom{0}3.6 \\

        \bottomrule
      \end{tabular} 
    }
   
    \caption{Ablation over the key design choices in $\net$. The adapters are crucial for precise pose estimation and their removal leads to massive performance drops.
    In the second block, we evaluate without using the foreground mask obtained from the CrossAttention layers and solely from confidence values.
    "w/o CA" was trained without any CrossAttention layers, with the rest of the architecture being identical.
    Confidence measures alone lead to reasonable performance, especially with the optimal threshold parameter ($t_2=0.8$).
    However, it is still consistently worse than the network trained with CrossAttention ("w CA"), even without using the mask during inference when setting $t_1=0$.
    In the third block, we ablate over the choices in the instance-level 9D pose refinement part of our pipeline.
    "w/o refinement + w/o size" refers to directly outputting the poses after ProgX, yielding consistently worse results. 
    "w refinement" refers to the instance level pose refinement given the category-level correspondences, which gives a marginal improvement across all metrics.
    "w size estimation" includes the size optimization from the instance-level correspondences and shows that this is crucial for good performance on the tight $NIoU_{75}$ metric.
        \label{tab:ablation}
        }
\end{table*}

\section{Architectural choices}

We show an ablation on the core design choices of our method in \cref{tab:ablation}.

\textbf{Adapter.}
The PEFT strategy to introduce dataset- and task-specific information into the pre-trained feature extractor massively benefits our method.
Only by modifying the feedforward part of the transformer blocks shows significant performance improvements which is especially noticeable for rotation accuracy.

\textbf{Foreground Modeling.}
Next, we evaluated the effect of our foreground modeling via CrossAttention.
Specifically, we compare two variants.
First, we evaluate the effect of focusing the model onto the foreground region during training via a baseline that has all CrossAttention layers removed and filters outliers during inference with confidence scores.
We found that the same threshold $t_2=0.7$ we used for the full model is not ideal in this case and a more robust segmentation is obtained with $t_2=0.8$.
This naive baseline achieves good results, indicating that our method can identify vertices with high likelihood.
Next, we use our full model but ignore the provided mask during inference by setting $t_1=0$.
This variant consistently outperforms the previous, indicating that focusing the model on the foreground region explicitly during training leads to better representation learning.
However, utilizing the foreground mask still provides consistent improvements across all metrics, indicating its importance for correspondence estimation.

\textbf{Pose refinement.}
Finally, we ablate over the components in or 9D pose refinement stage that utilizes the features that follow the instance-level prototype geometries.
We show that returning the poses obtained from ProgX (i.e. 6D poses) leads to consistently worse performance.
Even an instance level refinement using the category-level correspondences $\mcorr$ leads to performance improvements (see row "w refinement").
However, to obtain strong results for the tightest bounding box threshold $NIoU_{75}$ the size optimization using the instance-level correspondences $\corr$ is required (see row "w size estimation").
Best performance, however, is obtained when refining the deformed 2D/3D correspondences, leading to our full pipeline.

\section{Inference Speed}

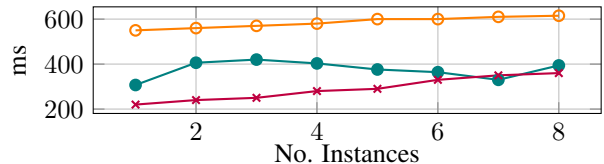
\begin{figure}[h]
\centering
\begin{tikzpicture}
  \begin{axis}[
    width=\columnwidth,
    height=3cm,
    xlabel={No. Instances},
    xlabel style={yshift=5pt},
    ylabel={ms},
    grid=major,
    mark options={fill=blue},
    every axis plot/.append style={thick},
  ]
    \addplot[
      color=teal,
      mark=*,
      mark options={fill=teal},
    ] coordinates {
      (1, 307)
      (2, 406)
      (3, 420)
      (4, 403)
      (5, 376)
      (6, 364)
      (7, 330)
      (8, 394)
    };
        \addplot[
      color=purple,
      mark=x,
    ] coordinates {
      (1, 220)
      (2, 240)
      (3, 250)
      (4, 280)
      (5, 290)
      (6, 330)
      (7, 350)
      (8, 360)
    };
    \addplot[
      color=orange,
      mark=o,
    ] coordinates {
      (1, 550)
      (2, 560)
      (3, 570)
      (4, 580)
      (5, 600)
      (6, 600)
      (7, 610)
      (8, 615)
    };
  \end{axis}
\end{tikzpicture}
\caption{Average inference runtime \textcolor{teal}{our method} and the baselines \textcolor{purple}{LaPose} and \textcolor{orange}{DMSR}. 
    Runtime of our method depends on the found correspondences instead of present instances.
    }
    \label{fig: runtime}
\end{figure}

We compare the inference speed of our method with the two-stage baselines in \cref{fig: runtime}.
In contrast to baselines, our method requires only a single forward pass.
Two-stage methods require one forward pass of the detection model and one call for each detected object.
Our method, on the other hand, is more reliant on the choice of output resolution and found correspondences.
With a more aggressive outlier rejection or subsampling of correspondences inference speed can be greatly improved with only minor loss in accuracy.
In this work, however, we did not optimize for inference speed and consider speed-up strategies as future work.

\section{Pose Estimation for Overlapping Objects}

REAL275~\cite{nocs} does not contain many overlapping objects of the same category.
To showcase that our method can deal with intra-category overlaps we captured in-the-wild images with a smartphone and approximated its intrinsic matrix.
We show the predictions of our method in \cref{fig: overlap}.
\begin{minipage}{\columnwidth}
    \includegraphics[trim={0 2cm 0 1cm},clip,width=0.495\columnwidth]{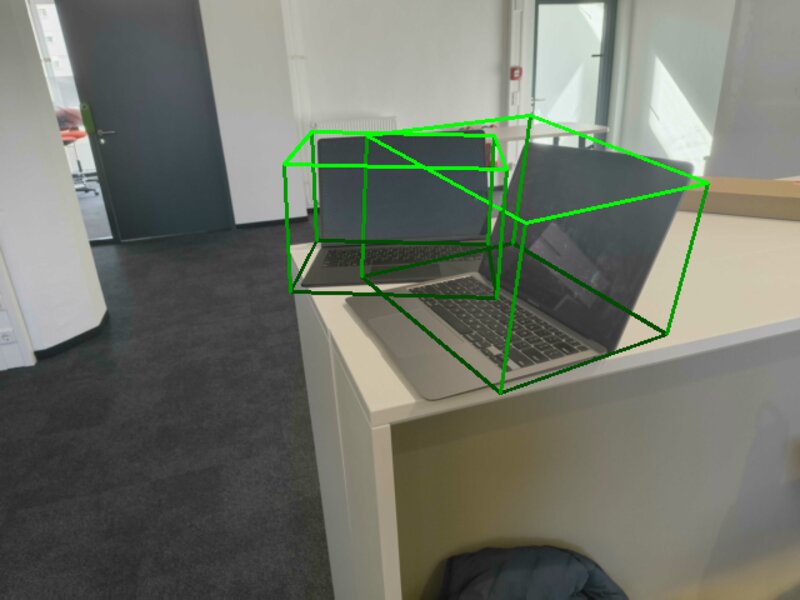}
    \includegraphics[trim={0 2cm 0 1cm}, clip, width=0.495\columnwidth]{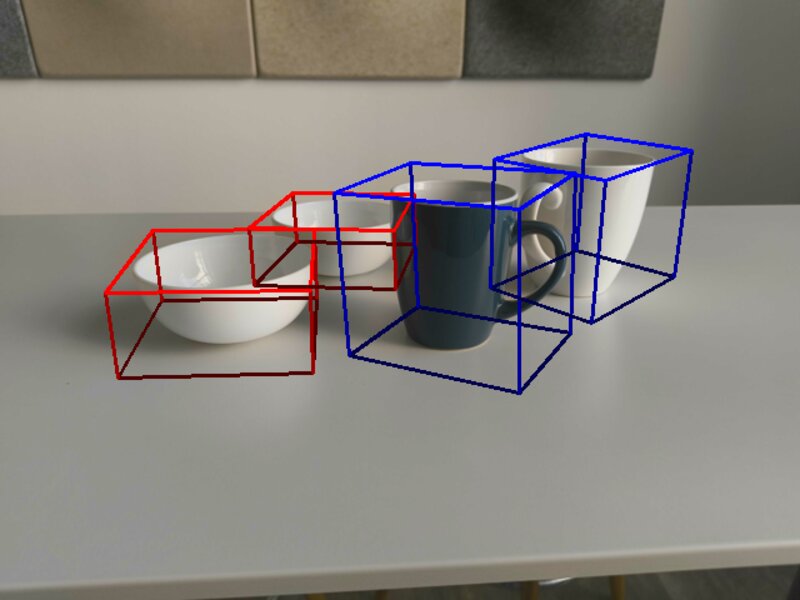}
    \captionof{figure}{Detections and poses from our model with same-category occlusions on self-captured images.}
    \label{fig: overlap}
\end{minipage}

\section{Additional Quantitative Results on Robustness Study}

In \cref{tab:noise_mean}, we show the pose estimation accuracy under scale-agnostic metrics averaged over all corruption types.
The corrupted images are generated using the method proposed in ~\cite{imagecorruptions} and using their public code. 
We consider four types of image degradations, encompassing a total of eight corruption types: noise (speckle noise, Gaussian noise), blur (Gaussian blur, defocus blur), digital artifacts (JPEG compression, elastic transformation), and weather effects (frost, fog). 
The corruption strength follows the default setting, varying in severity per image basis. 
For a fair comparison, we run each method on the same set of images.

We report the mean scale-agnostic 3D Intersection over Union (NIoU), rotation, and translation metrics for all methods. 
Additionally, we show results for each corruption type at ROI level in \cref{tab:noise_rois} and at image level in \cref{tab:noise_full}. 

\begin{table*}[hbt!]
    \centering
    \resizebox{\textwidth}{!}{%
    \begin{tabular}{@{}c|c|c c c|c c c c c c c c@{}}
    \toprule
        Source &Method & $NIoU_{25}$ & $NIoU_{50}$ & $NIoU_{75}$ & $5^\circ 0.2d$  & $5^\circ 0.5d$  & $10^\circ 0.2d$ & $10^\circ 0.5d$ &$0.2d$ & $0.5d$ &  $5^\circ$ & $10^\circ$ \\ \hline

        \multirow{4}{*}{\rotatebox[origin=c]{90}{None}}
        &MSOS~\cite{msos} & 36.9 & \hphantom{0}9.7 & \hphantom{0}0.7 & - & - &\hphantom{0}3.3 & 15.3 & 10.6 & 50.8 & - & 17.0 \\ 
        &OLD-Net~\cite{oldnet} & 35.4 & 11.4 & \hphantom{0}0.4 & \hphantom{0}0.9 & \hphantom{0}3.0 & \hphantom{0}5.0 & 16.0 & 12.4 & 46.2 & \hphantom{0}4.2 & 20.9  \\
        &DMSR~\cite{dmsr} & 57.2 & 38.4 & \hphantom{0}9.5 & 15.1 & 23.7  & 25.6 & 45.2 & 35.0 & 67.3 &  27.4 & 52.0\\ 
        &LaPose~\cite{lapose}  & 70.7 & 47.9 & 15.8 & 15.7 &  21.3& 37.4 & 57.4 & 46.9 & 78.8 & 23.4 & 60.7\\ 
        &Ours  & \textbf{71.6} & \textbf{50.5} & \textbf{18.3} & \textbf{21.6} & \textbf{28.6} & \textbf{41.9} & \textbf{61.6} & \textbf{51.0} & \textbf{80.9} & \textbf{29.1} & \textbf{64.1}\\
        \hline

        \multirow{3}{*}{\rotatebox[origin=c]{90}{ROI}}
        &OLD-Net\cite{oldnet} & 30.3  & \hphantom{0}8.6 & \hphantom{0}0.2 & \hphantom{0}0.5 & \hphantom{0}1.9 &\hphantom{0}3.6 & 12.2 &10.4 & 41.2 & \hphantom{0}3.2 &17.1   \\
        &DMSR\cite{dmsr}  & 55.3 & 35.6 & \hphantom{0}8.2 & 12.4 & 19.4 & 23.5 & 40.5 & 33.4 & 65.5 & 22.6  & 46.0 \\
        &LaPose\cite{lapose}   & 64.1 & 39.8 & 12.9 & 11.2 & 16.2 & 27.9 & 47.9 & 38.3 & 73.9 & 18.4 & 52.0  \\
        \hline
        
        \multirow{4}{*}{\rotatebox[origin=c]{90}{Image}}
        &OLD-Net\cite{oldnet} & 25.5 & \hphantom{0}7.0 & \hphantom{0}0.2 & \hphantom{0}0.4 & \hphantom{0}1.7 & \hphantom{0}2.7 & \hphantom{0}9.4 & \hphantom{0}8.1 & 34.0 & \hphantom{0}3.2 & 15.7  \\
        &DMSR\cite{dmsr} & 49.3	& 32.9	& \hphantom{0}7.4	&11.4	&17.1	&21.4	&35.2	&30.4 &57.1	& 20.1 &	40.3 \\
        &LaPose\cite{lapose} & 54.5 & 36.1 & 12.2 & 11.1 & 15.2 & 26.4 & 41.5 & 34.9 & 62.1 & 17.3 & 45.0  \\
        
        &Ours  & \textbf{63.8} & \textbf{43.4} & \textbf{14.3} & \textbf{17.9} & \textbf{24.7} & \textbf{35.2} & \textbf{53.7} & \textbf{44.5} & \textbf{73.2} & \textbf{25.2} & \textbf{56.0}\\
        \bottomrule

    \end{tabular}
    
    }
    \caption{Ablation study of robustness under image noises. We report the performance of all methods on clean data (top), as well as ROI corruptions for baselines (middle), and image-level corruptions (bottom).
    Note, that performance of our method on clean data is different due to the changed training regiment to make this comparison fair.
    \label{tab:noise_mean}
    }
\end{table*}

\begin{table*}[t]
    \centering
    \resizebox{\textwidth}{!}{%
     \begin{tabular}{@{}c|c|c c c|c c c c c c c c@{}}
     \toprule
Method  & Noise & $NIoU_{25}$ & $NIoU_{50}$ & $NIoU_{75}$ & $5^\circ 0.2d$  & $5^\circ 0.5d$  & $10^\circ 0.2d$ & $10^\circ 0.5d$ & $0.2d$ & $0.5d$ & $5^\circ$ & $10^\circ$\\  \hline
LaPose  & \multirow{4}{*}{Speckle Noise}                                            & 56.7   & 35.8   & \textbf{13.3}   & \hphantom{0}8.6     & 11.7    & 25.1    & 39.8     & 34.9 & 64.9 & 13.9 & 44.2 \\
DMSR  & & 48.8 &	31.1	& \hphantom{0}6.4&	 \hphantom{0}8.9	&15.2	&17.9	&33.4&	27.9&	55.7&	19.4&	40.3 \\
Old-Net &                                                                           & 23.2   & \hphantom{0}6.7    & \hphantom{0}0.2    & \hphantom{0}0.2     & \hphantom{0}1.3     & \hphantom{0}2.6      & \hphantom{0}0.3      & \hphantom{0}7.7  & 32.1 & \hphantom{0}3.2  & 17.0 \\
Ours    &                                                                           & \textbf{61.8}  & \textbf{40.3}   & 12.6    & \textbf{15.0}   & \textbf{22.6}   & \textbf{31.1}     &\textbf{50.6}    & \textbf{40.8} & \textbf{71.6} & \textbf{23.2} & \textbf{53.5} \\ \hline

LaPose  & \multirow{4}{*}{\begin{tabular}[c]{@{}l@{}}Gaussian Blur\end{tabular}}  & 58.2   & 41.9   & 17.3  & 13.4    & 18.0    & 31.1     & 44.1     & 40.1 & 64.0 & 20.2 & 47.6 \\
DMSR    &                                                                           & 54.3 &	40.4&	10.3&	14.1&	18.5&	27.7&	39.4&	36.6&	60.3&	21.2&	43.5 \\
Old-Net &                                                                           & 30.7   & \hphantom{0}7.4    & \hphantom{0}0.3    & \hphantom{0}0.4     & \hphantom{0}2.6     & \hphantom{0}3.3      & 13.2     & \hphantom{0}9.7  & 39.1 & \hphantom{0}4.1  & 18.5 \\
Ours    &                                                                           & \textbf{69.9}  & \textbf{48.9}   & \textbf{17.4}    & \textbf{19.6}   & \textbf{25.9}    & \textbf{39.7}    & \textbf{59.3}     & \textbf{49.5} & \textbf{78.9} & \textbf{26.2} & \textbf{61.3} \\ \hline

LaPose  & \multirow{4}{*}{\begin{tabular}[c]{@{}l@{}}Gaussian Noise\end{tabular}} & 52.2   & 32.5   & 11.9   & \hphantom{0}8.3     & 11.3    & 23.9     & 36.4     & 31.9 & 60.8 & 13.0 & 40.0 \\
DMSR    &                                                                           & 45.2&	30.1&	\hphantom{0}7.2&	10.6&	17.1&	19.3&	34.0&	28.0&	52.9&	19.3&	39.2 \\
Old-Net &                                                                           & 24.8   & \hphantom{0}8.7    & \hphantom{0}0.2    & \hphantom{0}0.6     & \hphantom{0}2.2     & \hphantom{0}3.3      & 11.4     & \hphantom{0}9.0  & 32.3 & \hphantom{0}4.1  & 17.2 \\
Ours    &                                                                           & \textbf{56.9}   & \textbf{37.8}   &\textbf{12.2}    & \textbf{14.9}   & \textbf{21.0}   & \textbf{30.4}     & \textbf{47.1}     & \textbf{38.7} & \textbf{65.4} & \textbf{21.7} & \textbf{49.7} \\ \hline

LaPose  & \multirow{4}{*}{Defocus Blur}                                             & 54.1   & 39.3   & 11.4   & 12.2    & 16.1    & 29.0    & 42.1     & 36.9& 60.2 & 17.4 & 44.3 \\
DMSR    &                                                                           & 53.2&	37.9&	\hphantom{0}8.8&	10.8&	15.4&	24.5&	37.2&	36&	61.8&	17.4&	41.1 \\
Old-Net &                                                                           & 22.6   & \hphantom{0}5.9    & \hphantom{0}0.1    & \hphantom{0}0.2     & \hphantom{0}1.3     & \hphantom{0}2.1      & \hphantom{0}8.8      & \hphantom{0}7.6  & 30.0 & \hphantom{0}3.1  & 15.1 \\
Ours    &                                                                           & \textbf{63.2}  & \textbf{42.3}   & \textbf{14.7}    & \textbf{17.8}   & \textbf{24.6}    & \textbf{34.6}     & \textbf{53.5}    & \textbf{43.1} & \textbf{73.0} & \textbf{24.9} & \textbf{55.9} \\ \hline

LaPose  & \multirow{4}{*}{JPEG Compression}                                                     & 57.3   & 37.6   & \hphantom{0}9.9    & 11.1    & 16.6    & 27.1     & 45.1     & 36.1 & 66.9 & 19.3 & 49.2 \\
DMSR    &                                                                           & 51.3&	32.4&	\hphantom{0}8.1&	11.3&	14.8&	21.5&	32.3&	31.7&	60.3&	16.5&	34.9\\
Old-Net &                                                                           & 35.9   & 12.4   & \hphantom{0}0.3    & \hphantom{0}0.8     & \hphantom{0}2.7     & \hphantom{0}5.0      & 15.8     & 13.5 & 44.8 & \hphantom{0}4.1  & 20.1 \\
Ours    &                                                                           & \textbf{70.4 }  & \textbf{50.7}   & \textbf{18.5}    & \textbf{22.1}   & \textbf{29.2 }   & \textbf{41.4}   & \textbf{59.5}    & \textbf{51.6}& \textbf{80.2} & \textbf{29.7} & \textbf{62.4}
\\ \hline

LaPose  & \multirow{4}{*}{Elastic Transform}                                       & 58.6   & 36.9   & 10.5  & 12.0    & 17.3    & 27.7    & 47.1     & 35.2 & 66.4 & 19.5 & 50.8 \\
DMSR    &                                                                           & 49.0 &	30.9&	\hphantom{0}6.4&	11.2&	17.4&	19.5&	35.0&	27.9&	57.5&	20.5&	40.8 \\
Old-Net &                                                                           & 25.3   & \hphantom{0}5.6    & \hphantom{0}0.2    & \hphantom{0}0.4     & \hphantom{0}2.6     & \hphantom{0}2.2      & 11.9     & \hphantom{0}6.2  & 34.8 & \hphantom{0}4.3  & 17.6 \\
Ours    &                                                                           & \textbf{68.0}  & \textbf{45.4}   & \textbf{15.3}   & \textbf{17.8}    & \textbf{24.7}   & \textbf{36.9}     & \textbf{57.9}     & \textbf{45.9} & \textbf{78.5} & \textbf{25.1} & \textbf{60.5}
\\ \hline

LaPose  & \multirow{4}{*}{Frost}                                                    & 49.4   & 32.3   & 11.5   & 11.6  & 15.1    & 23.7     & 38.5     & 31.9 & 56.8 & 17.7 & 42.0 \\
DMSR    &                                                                           & 39.5&	25.1&	\hphantom{0}4.8&	10.1&	16.9&	16.4&	29.0&	23.4&	46.6&	20.0&	34.7 \\
Old-Net &                                                                           & 23.8   & \hphantom{0}6.6    & \hphantom{0}0.2    & \hphantom{0}0.2     & \hphantom{0}0.9     & \hphantom{0}2.2      & \hphantom{0}8.9      & \hphantom{0}7.4  & 31.6 & \hphantom{0}1.6  & 12.8 \\
Ours    &                                                                           & \textbf{60.9}   & \textbf{41.1}  & \textbf{12.8}   & \textbf{18.4}    & \textbf{25.4}    & \textbf{33.9 }    & \textbf{51.2}   & \textbf{43.8 }& \textbf{70.7} & \textbf{25.7} & \textbf{53.0} \\ \hline

LaPose  & \multirow{4}{*}{Fog}                                                      & \textbf{66.9}  & \textbf{46.5}  & \textbf{17.2}   & 15.2    & 19.9    & \textbf{35.0}    & \textbf{53.9}     & \textbf{46.0} & \textbf{75.3} & 22.0 & \textbf{57.0} \\
DMSR    &                                                                           & 53.1&	35.2&	\hphantom{0}7.5&	14.0&	21.7&	24.1&	41.1&	32.1&	61.7&	26.3&	47.9 \\
Old-Net &                                                                           & 17.8   & \hphantom{0}2.4    & \hphantom{0}0.1    & \hphantom{0}0.1     & \hphantom{0}0.3     & \hphantom{0}1.1      & \hphantom{0}4.9      & \hphantom{0}4.0  & 27.6 & \hphantom{0}0.7  & \hphantom{0}7.3  \\
Ours    &                                                                           & 58.9   & 40.6   & 11.1    & \textbf{17.4}    & \textbf{24.4}    & 33.4    & 50.2     & 42.6 & 67.5 & \textbf{24.9} & 51.8 \\
\bottomrule
\end{tabular}}
\caption{We show per corruption accuracy of all methods. Corruptions are applied to the full image, affecting both detection and pose estimation. Our method outperforms the baselines on a majority of the corruption types.
}
\label{tab:noise_full}

\end{table*}
\begin{table*}[t]
    \centering
    \resizebox{\textwidth}{!}{%
     \begin{tabular}{@{}c|c|c c c|c c c c c c c c@{}}
    \toprule
Method  & Noise & $NIoU_{25}$ & $NIoU_{50}$ & $NIoU_{75}$ & $5^\circ 0.2d$  & $5^\circ 0.5d$  & $10^\circ 0.2d$ & $10^\circ 0.5d$ & $0.2d$ & $0.5d$ & $5^\circ$ & $10^\circ$\\ \hline
LaPose  & \multirow{4}{*}{Speckle Noise}                                            & \textbf{62.2}   & \textbf{36.5}   & \textbf{12.6} & \hphantom{0}7.0     & 10.1    & \textbf{23.7}     & 40.2     & \textbf{34.5} & \textbf{71.6} & 11.9 & 45.3 \\
DMSR    &                                                                           & 54.9   & 35.4   & \hphantom{0}8.9    & \textbf{10.9}  & \textbf{18.0}    & 22.9     & \textbf{40.5}     & 32.8 & 65.4 & \textbf{21.1} & \textbf{46.0} \\
Old-Net &                                                                           & 30.9   & \hphantom{0}9.1    & \hphantom{0}0.2    & \hphantom{0}0.5     & \hphantom{0}2.0     & \hphantom{0}4.1      & 13.4     & 10.6 & 41.6 & \hphantom{0}3.6  & 19.1 \\

\hline
LaPose  & \multirow{4}{*}{\begin{tabular}[c]{@{}l@{}}Gaussian Noise\end{tabular}} & \textbf{62.6}   & \textbf{34.3}   & \textbf{12.0}  & \hphantom{0}7.6     & 10.8    & \textbf{22.5}     & 37.2     & \textbf{33.8} & \textbf{72.4} & 12.3 & 41.5 \\
DMSR    &                                                                           & 55.3   & \textbf{34.3}   & \hphantom{0}7.6    & \textbf{11.8}   & \textbf{19.0}    & 22.3     & \textbf{39.5}     & 32.0 & 65.3 & \textbf{21.4} & \textbf{44.9} \\
Old-Net &                                                                           & 33.3   & 10.4   & \hphantom{0}0.2    & \hphantom{0}0.7     & \hphantom{0}2.2     & \hphantom{0}4.6      & 13.6     & 11.6 & 43.5 & \hphantom{0}3.5  & 19.0 \\

\hline
LaPose  & \multirow{4}{*}{\begin{tabular}[c]{@{}l@{}}Gaussian Blur\end{tabular}}  & \textbf{61.8}  & \textbf{41.5}   & \textbf{15.3}   & 13.3    & 19.0    & \textbf{31.4}     & \textbf{49.9}     & \textbf{40.5} & \textbf{71.9} & 22.1 & \textbf{54.5} \\
DMSR    &                                                                           & 57.9   & 39.8   & 10.1   & \textbf{14.4}   & \textbf{22.2}    & 26.7     & 45.1     & 37.5 & 67.7 & \textbf{26.1} & 51.4 \\
Old-Net &                                                                           & 30.6   & \hphantom{0}7.6    & \hphantom{0}0.3    & \hphantom{0}0.5     & \hphantom{0}2.9     & \hphantom{0}3.5      & 13.7     & \hphantom{0}9.9  & 41.5 & \hphantom{0}4.6  & 18.9 \\

\hline
LaPose  & \multirow{4}{*}{Defocus Blur}                                            & \textbf{60.8}   & \textbf{38.8}  & \textbf{11.0}   & \textbf{11.3}    & \textbf{16.7}    & \textbf{27.3}    & \textbf{48.0}     & \textbf{36.1} & \textbf{71.3} & \textbf{18.8} & \textbf{51.8} \\
DMSR    &                                                                           & 56.3   & 37.9   & \hphantom{0}8.7    & 10.6    & 15.7    & 24.0     & 38.9     & 35.8 & 66.2 & 18.5 & 44.0 \\
Old-Net &                                                                           & 25.9   & \hphantom{0}6.2    & \hphantom{0}0.1    & \hphantom{0}0.2     & \hphantom{0}1.3     & \hphantom{0}1.9      & \hphantom{0}8.6      & \hphantom{0}9.0  & 36.6 & \hphantom{0}3.4  & 15.7 \\
 \hline
LaPose  & \multirow{4}{*}{JPEG Compression}                                        & \textbf{58.3}   & \textbf{34.2}   & \textbf{\hphantom{0}8.9}    & \hphantom{0}9.9     & \textbf{16.3}    & \textbf{24.6}     & \textbf{46.3}     & \textbf{33.3} & \textbf{70.5} & \textbf{19.1} & \textbf{51.3} \\
DMSR    &                                                                           & 50.6   & 29.9   & \hphantom{0}6.4    & \textbf{10.6}    & 14.9    & 20.4     & 33.5     & 29.1 & 61.9 & 16.7 & 36.9 \\
Old-Net &                                                                           & 35.7   & 11.5   & \hphantom{0}0.4    & \hphantom{0}0.9     & \hphantom{0}2.9     & \hphantom{0}5.0      & 15.8     & 12.9 & 46.5 & 4.3  & 20.4 \\
 \hline
LaPose  & \multirow{4}{*}{Elastic Transform}                                       & \textbf{69.1}   & \textbf{44.5}  & \textbf{13.4}   & \textbf{13.6}    & 18.9    & \textbf{33.8}    & \textbf{54.8}    & \textbf{42.7} & \textbf{77.8} & 21.2 & \textbf{58.2} \\
DMSR    &                                                                           & 56.8   & 37.6   & \hphantom{0}8.7    & 13.4    & \textbf{20.2}    & 25.0     & 42.1     & 35.0 & 66.5 & \textbf{23.3} & 48.2 \\
Old-Net &                                                                           & 33.2   & \hphantom{0}9.5    & \hphantom{0}0.3    & \hphantom{0}0.7     & \hphantom{0}2.6     & \hphantom{0}4.1      & 14.4     & 11.1 & 44.2 & \hphantom{0}3.8  & 19.5 \\
 \hline
LaPose  & \multirow{4}{*}{Frost}                                                    & \textbf{68.0}  & \textbf{41.6}   & \textbf{13.4}   & 11.8    & 16.8    & 23.3     & \textbf{49.7}     & \textbf{40.0} & \textbf{76.9} & 18.7 & \textbf{52.9} \\
DMSR    &                                                                           & 55.6   & 35.3   & \hphantom{0}8.0    & \textbf{13.3}  & \textbf{21.8}   & \textbf{23.5}     & 40.6     & 33.5 & 66.2 & \textbf{24.9} & 45.6 \\
Old-Net &                                                                           & 35.6   & 12.0   & \hphantom{0}0.3    & \hphantom{0}0.4     & \hphantom{0}1.2     & \hphantom{0}4.8      & 13.1     & 14.0 & 47.6 & \hphantom{0}1.9  & 17.0 \\
 \hline
LaPose  & \multirow{4}{*}{Fog}                                                      & \textbf{70.3}   & \textbf{47.1}   & \textbf{16.2}   & \textbf{15.3}    & 20.9    & \textbf{36.2}     & \textbf{57.4}    & \textbf{45.8} & \textbf{78.8} & 23.1 & \textbf{60.4} \\
DMSR    &                                                                           & 55.0   & 34.6   & \hphantom{0}7.5    & 14.2    & \textbf{23.7}    & 23.1     & 43.8     & 31.1 & 65.1 & \textbf{28.6}& 50.9 \\
Old-Net &                                                                           & 17.3   & \hphantom{0}2.3    & \hphantom{0}0.1    & \hphantom{0}0.1     & \hphantom{0}0.3     & \hphantom{0}1.1      & \hphantom{0}4.8      & \hphantom{0}3.9  & 27.7 & \hphantom{0}0.6  & \hphantom{0}7.3  \\
\bottomrule
\end{tabular}
    }
\caption{Ablation study of robustness with corrupted ROIs.}
\label{tab:noise_rois}

\end{table*}

\section{Qualitative Results on Corrupted Images}
\begin{figure*}
    \centering
    \renewcommand{\arraystretch}{1.2}  
    \setlength{\tabcolsep}{3pt}  
    \resizebox{\textwidth}{!}{%
     \begin{tabular}{*{6}{c}} 
        \rotatebox[origin=l]{90}{\small Gaussian Noise} &
        \includegraphics[width=0.18\textwidth]{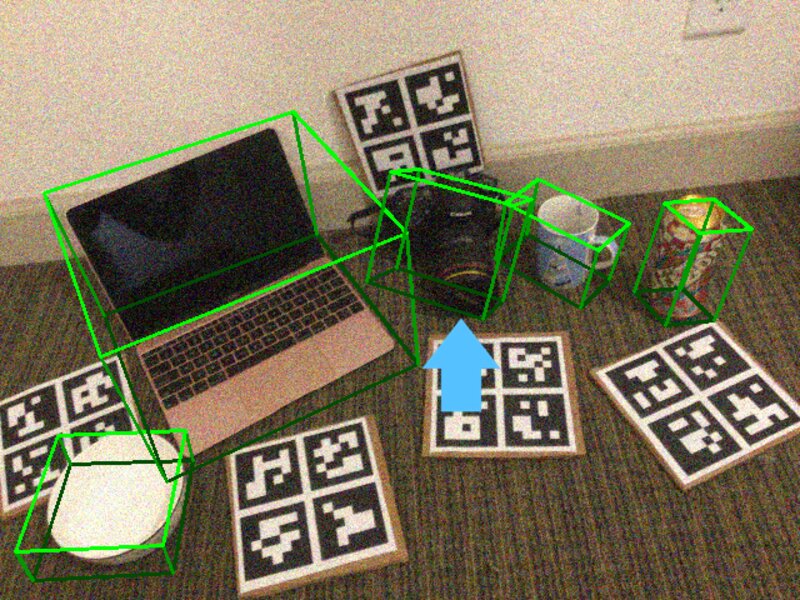} &
        \includegraphics[width=0.18\textwidth]{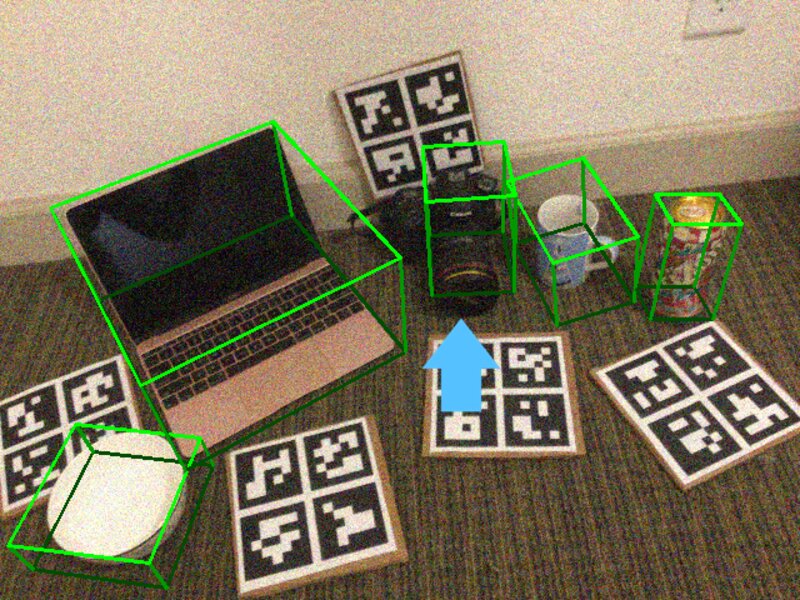} &
        \includegraphics[width=0.18\textwidth]{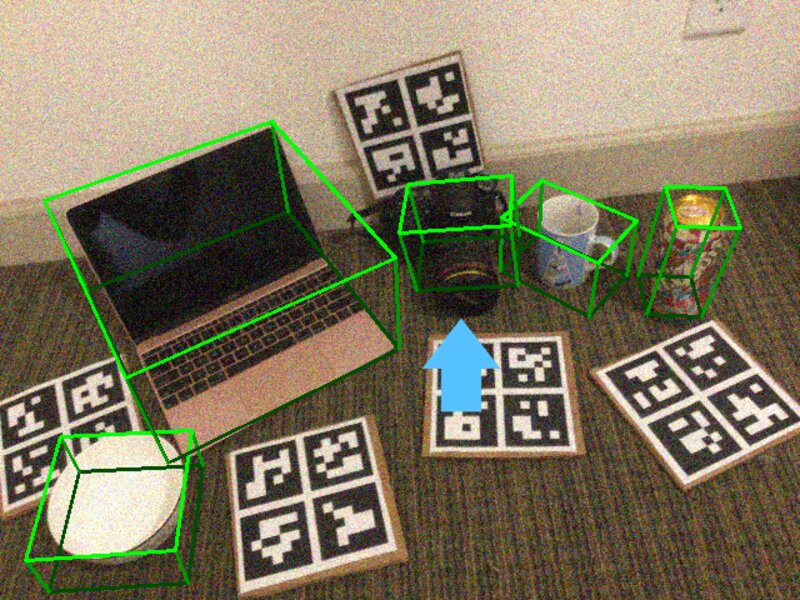} &
        \includegraphics[width=0.18\textwidth]{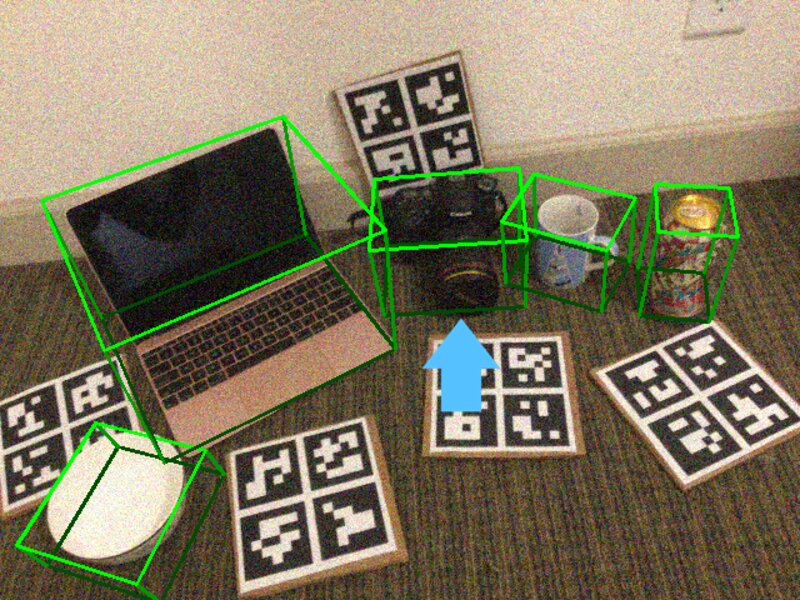}&
        \includegraphics[width=0.18\textwidth]{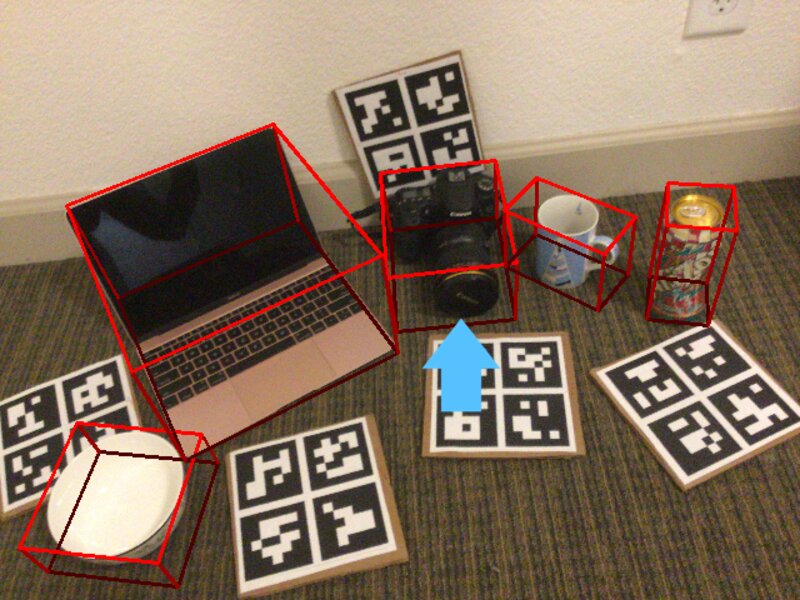}
        \\
        
        \rotatebox[origin=l]{90}{\small Speckle Noise} &
        \includegraphics[width=0.18\textwidth] {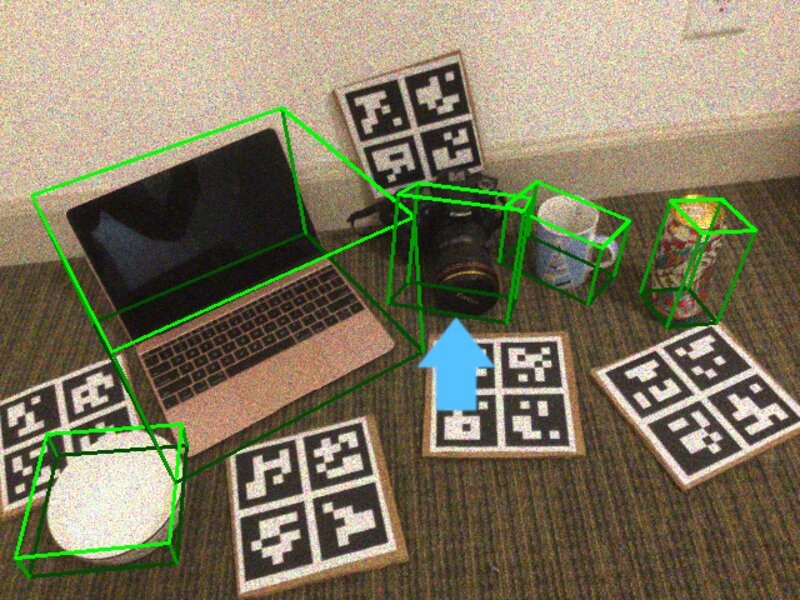} &
        \includegraphics[width=0.18\textwidth]{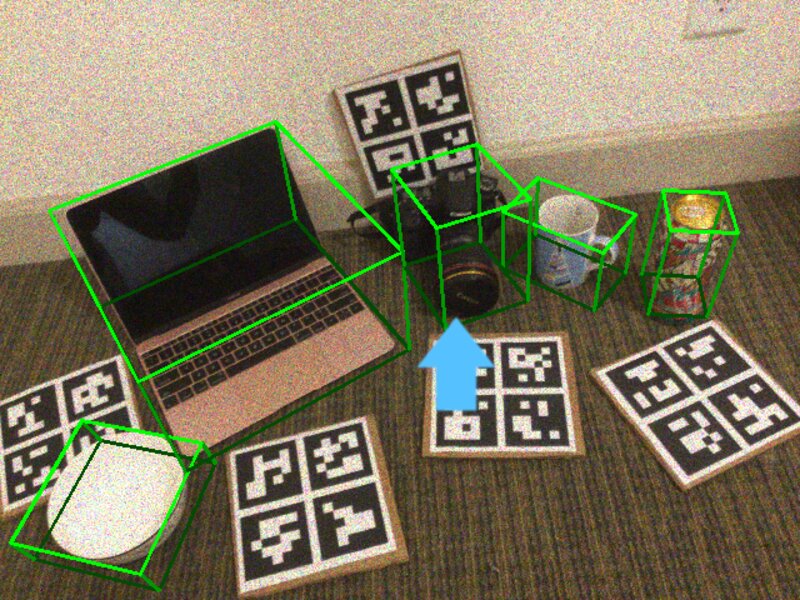} &
        \includegraphics[width=0.18\textwidth]{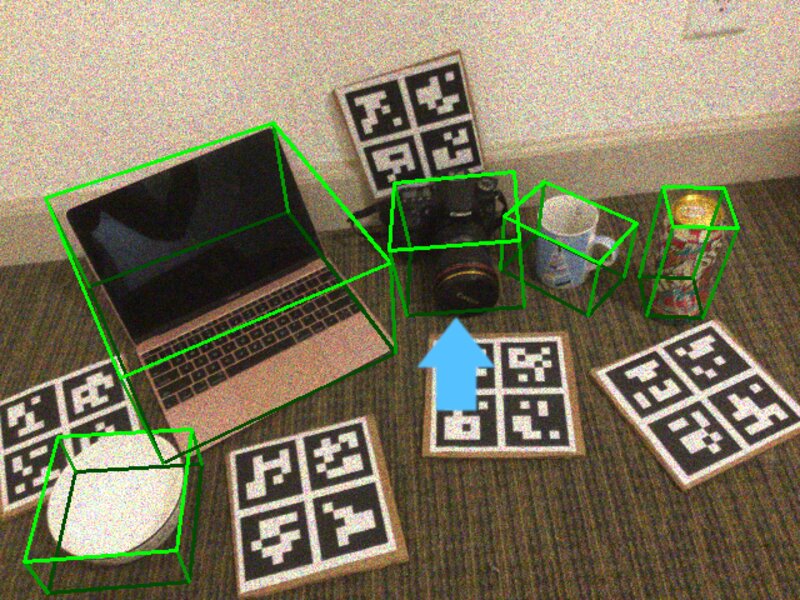} &
        \includegraphics[width=0.18\textwidth]{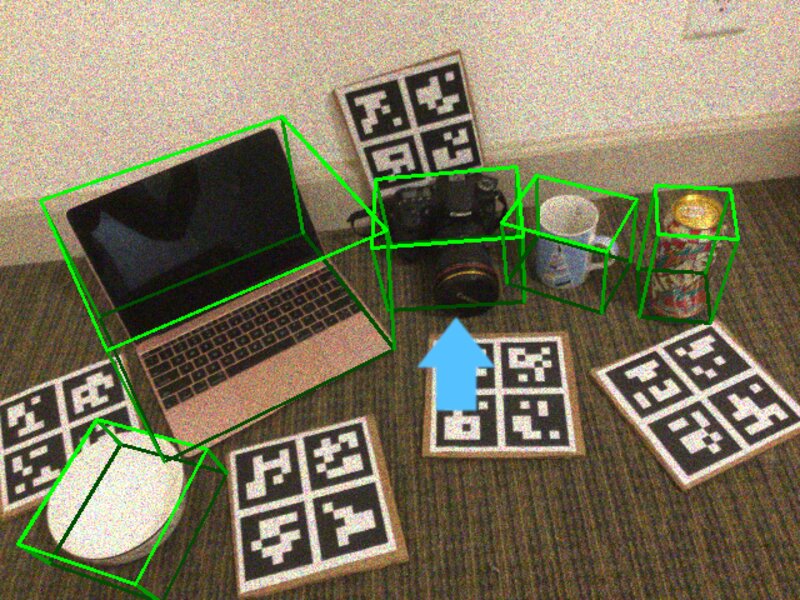}&
        \includegraphics[width=0.18\textwidth]{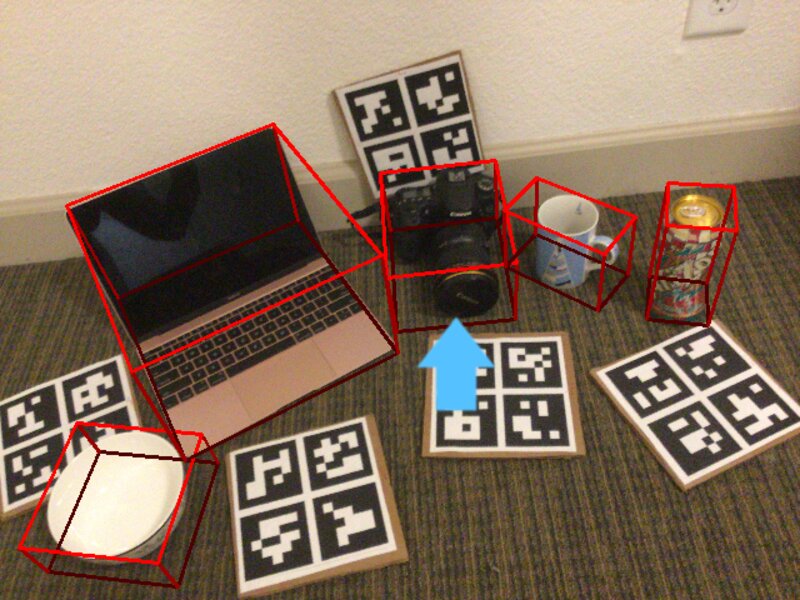} \\
        
        \rotatebox[origin=l]{90}{\small Gaussian Blur} &
        \includegraphics[width=0.18\textwidth]{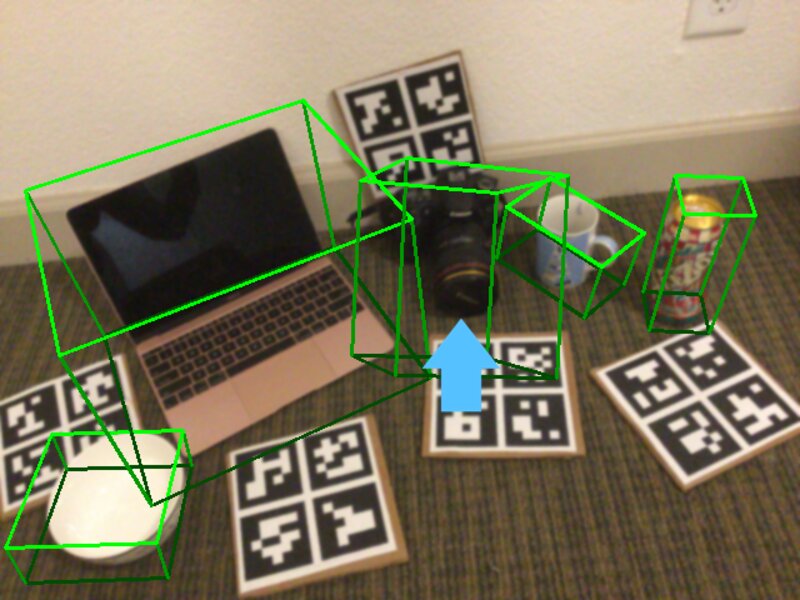} &
        \includegraphics[width=0.18\textwidth]{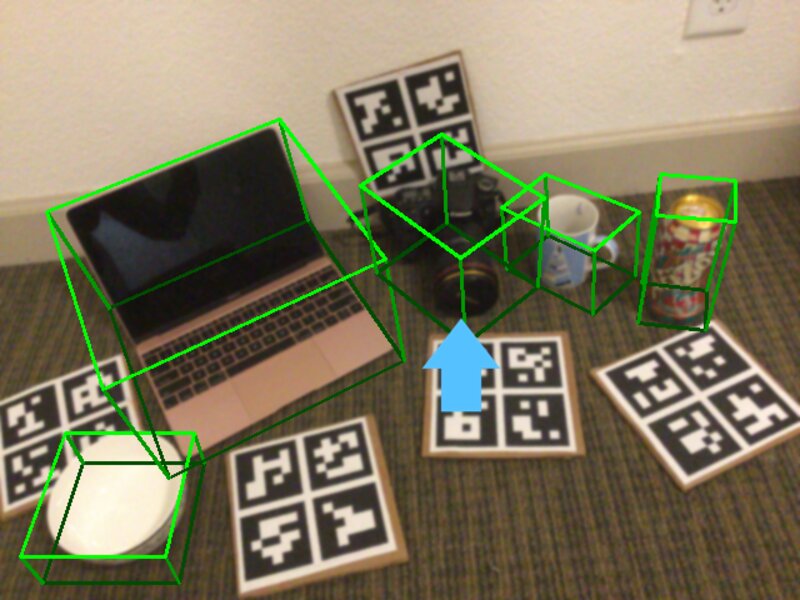} &
        \includegraphics[width=0.18\textwidth]{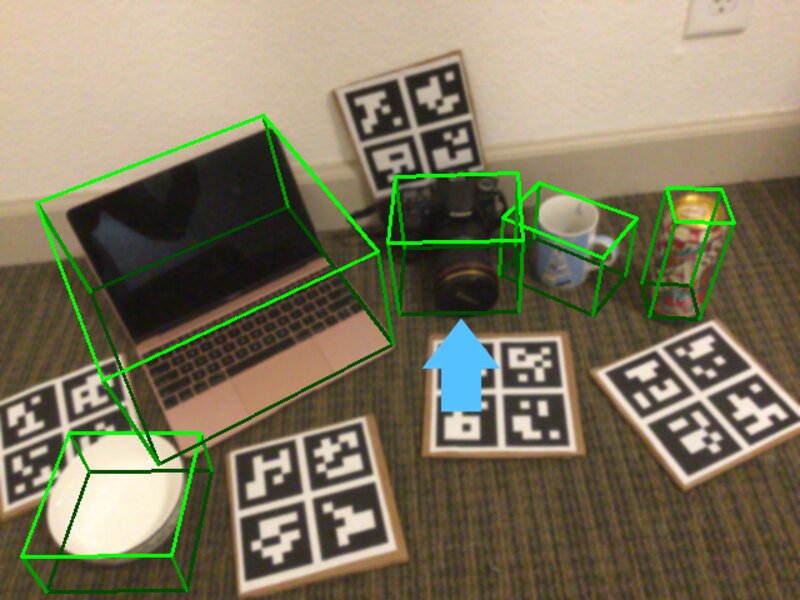} &
        \includegraphics[width=0.18\textwidth]{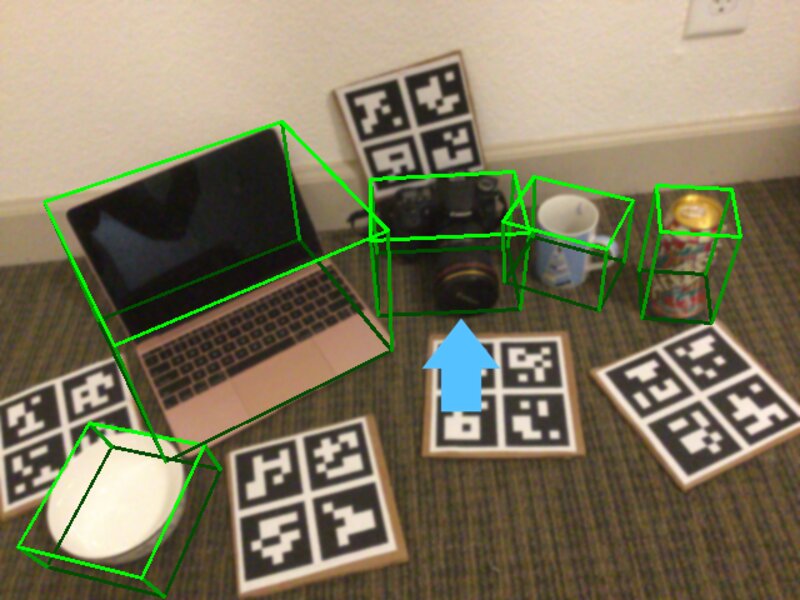}&
        \includegraphics[width=0.18\textwidth]{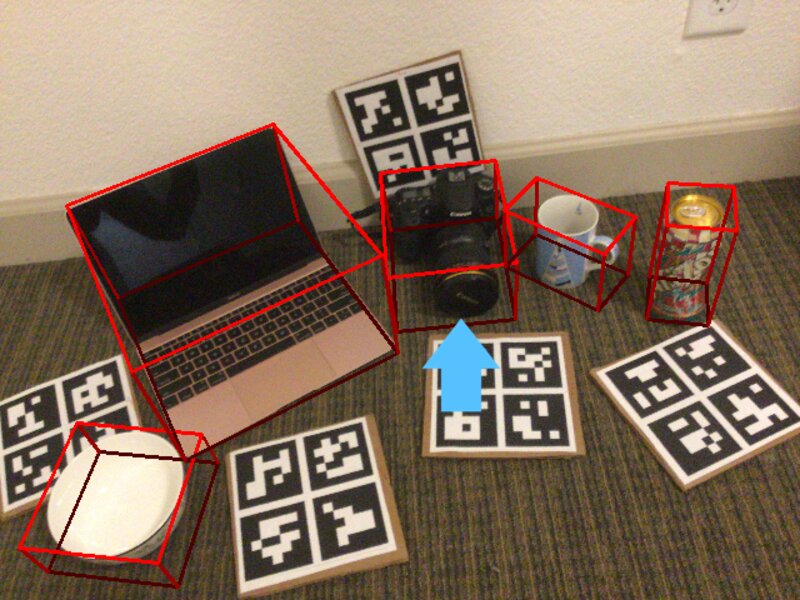} \\
        
        \rotatebox[origin=l]{90}{\small Defocus Blur} &
        \includegraphics[width=0.18\textwidth]{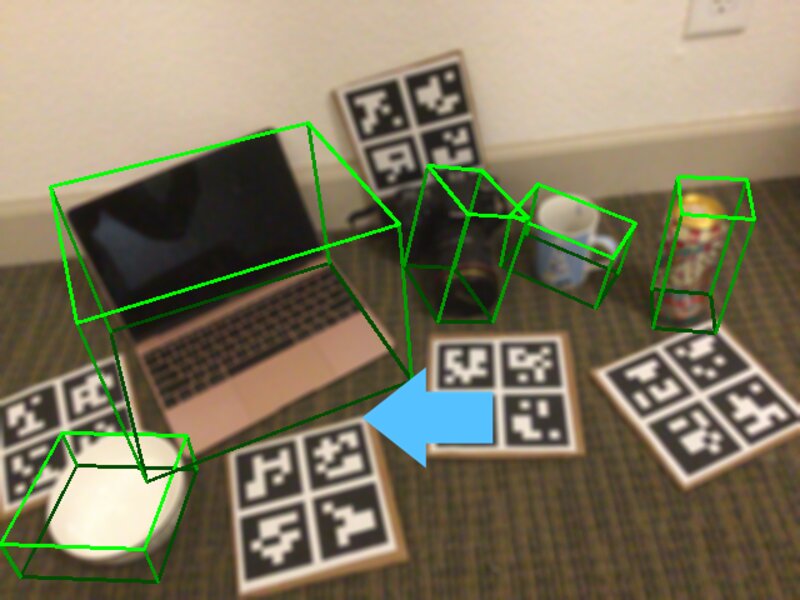} &
        \includegraphics[width=0.18\textwidth]{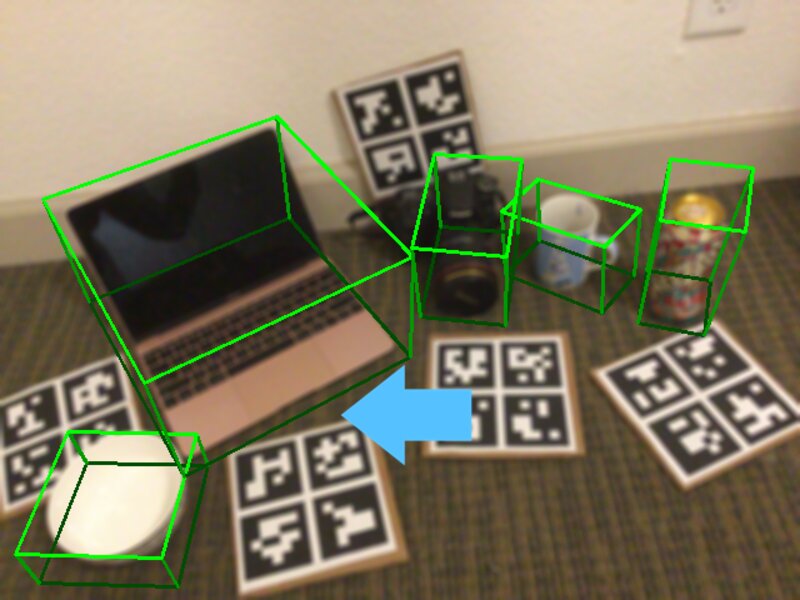} &
        \includegraphics[width=0.18\textwidth]{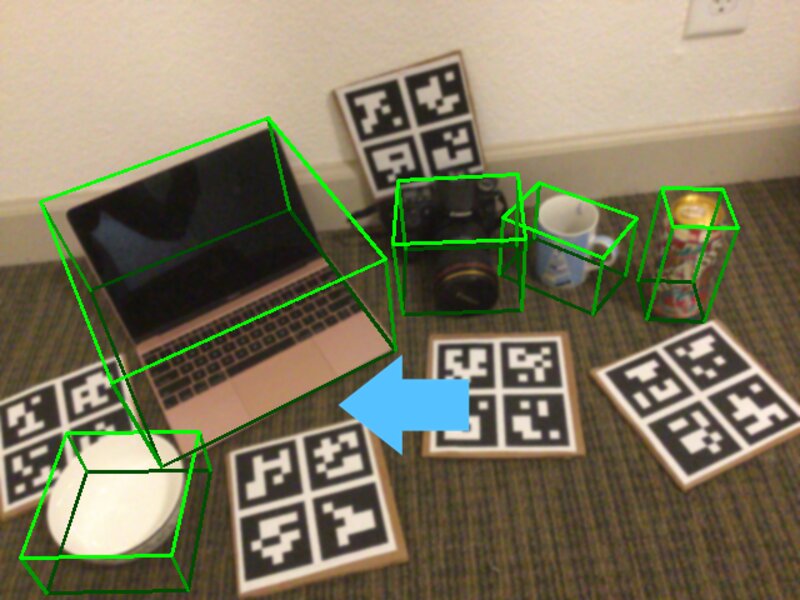} &
        \includegraphics[width=0.18\textwidth]{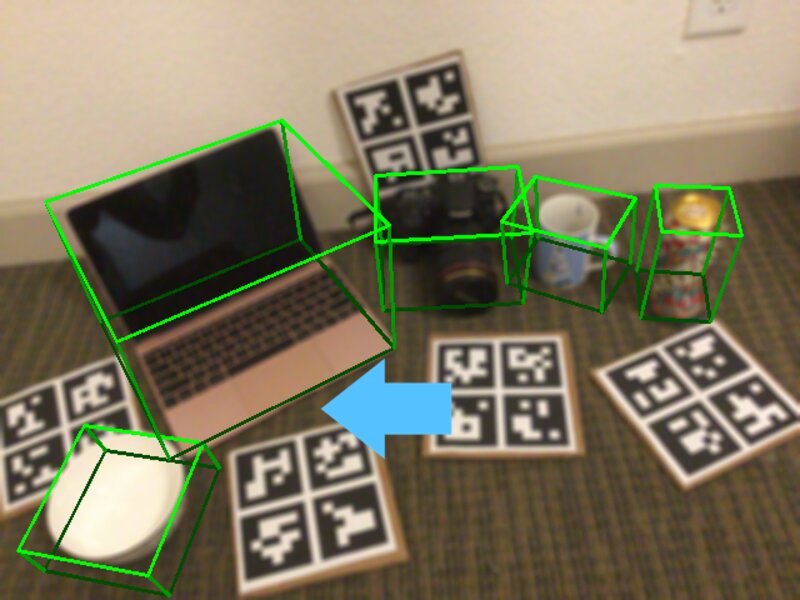}&
        \includegraphics[width=0.18\textwidth]{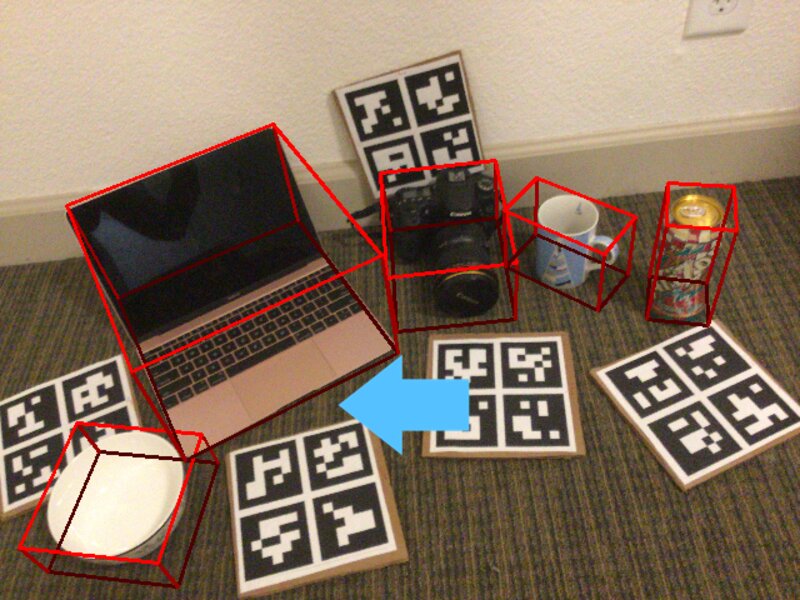}\\
        
        \rotatebox[origin=l]{90}{\small JPEG Compression} &
        \includegraphics[width=0.18\textwidth]{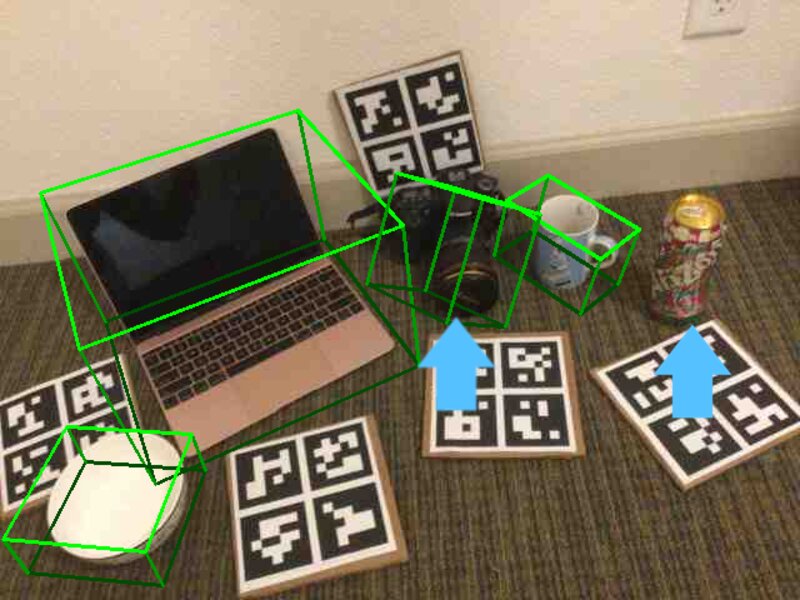} &
        \includegraphics[width=0.18\textwidth]{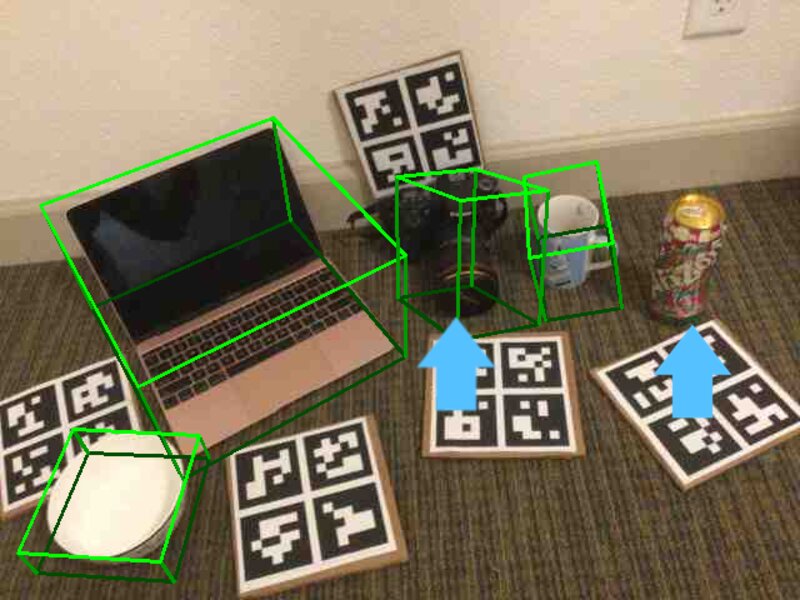} &
        \includegraphics[width=0.18\textwidth]{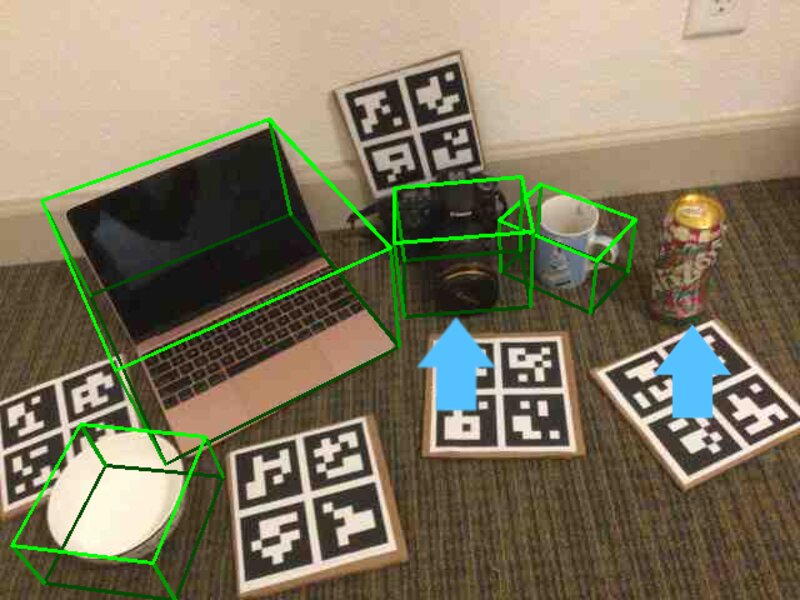} &
        \includegraphics[width=0.18\textwidth]{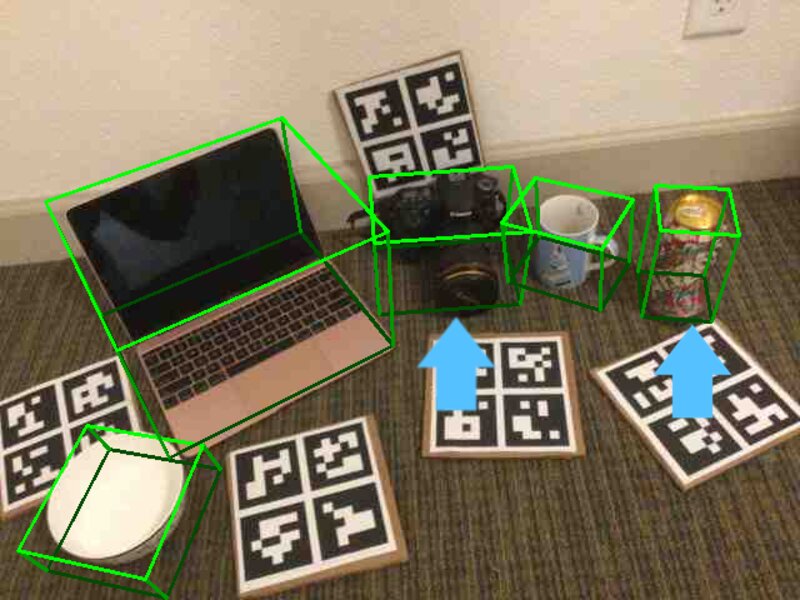}&
        \includegraphics[width=0.18\textwidth]{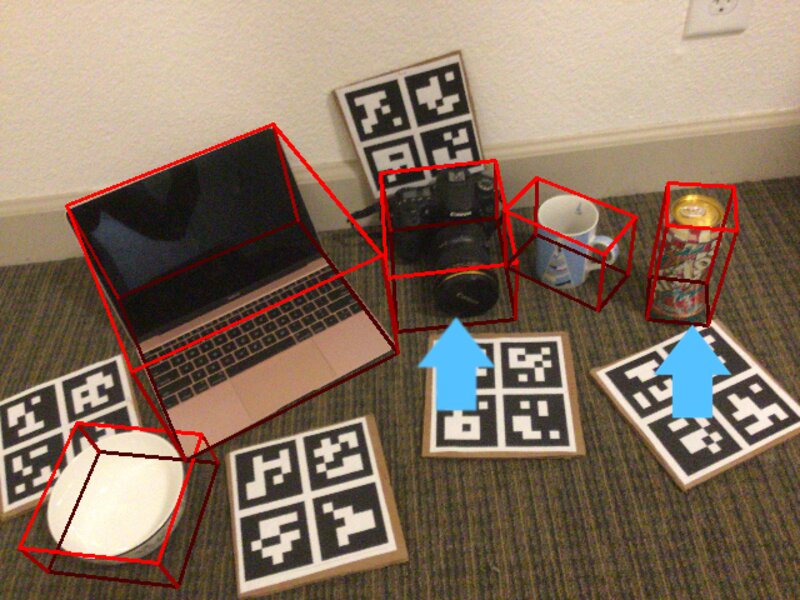}\\

        \rotatebox[origin=l]{90}{\small Elastic Transform} &
        \includegraphics[width=0.18\textwidth]{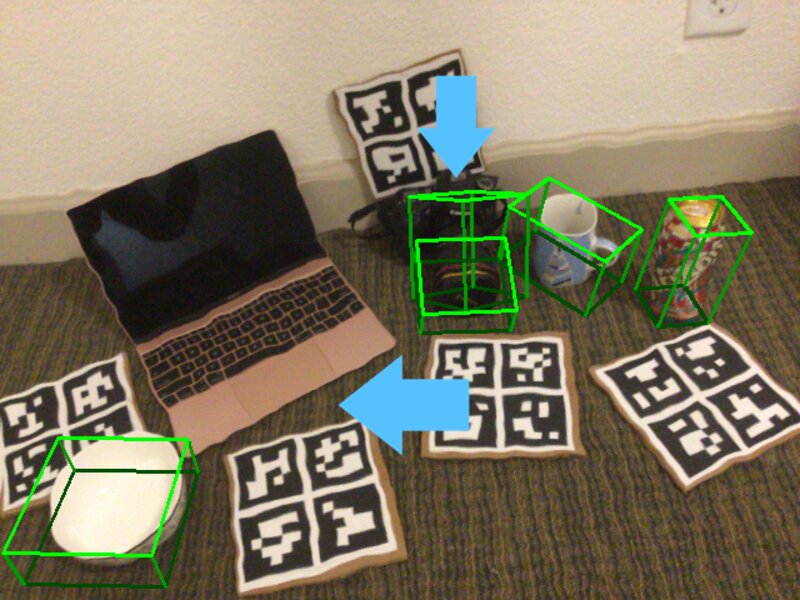} &
        \includegraphics[width=0.18\textwidth]{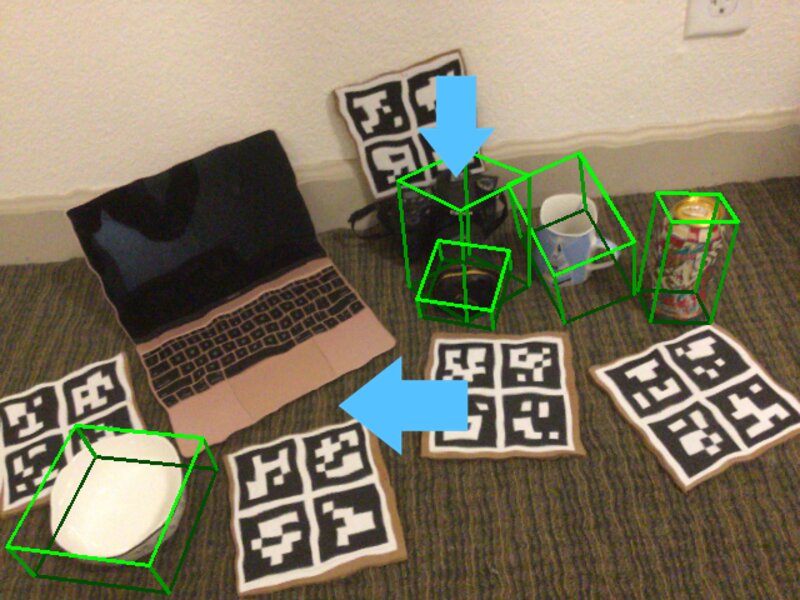} &
        \includegraphics[width=0.18\textwidth]{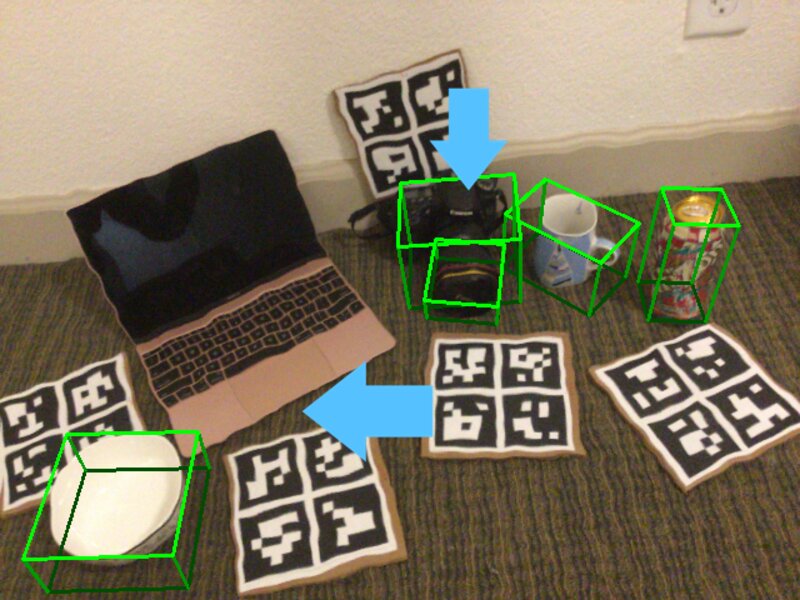} &
        \includegraphics[width=0.18\textwidth]{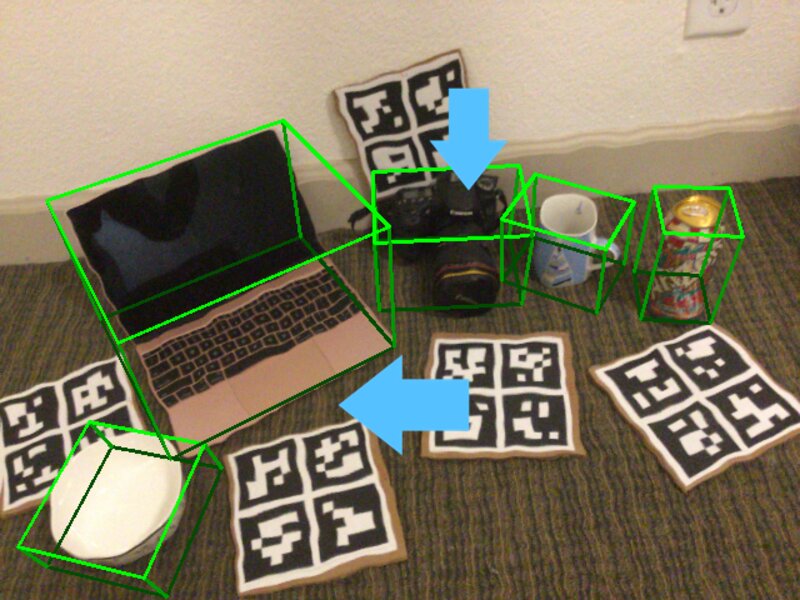}&
        \includegraphics[width=0.18\textwidth]{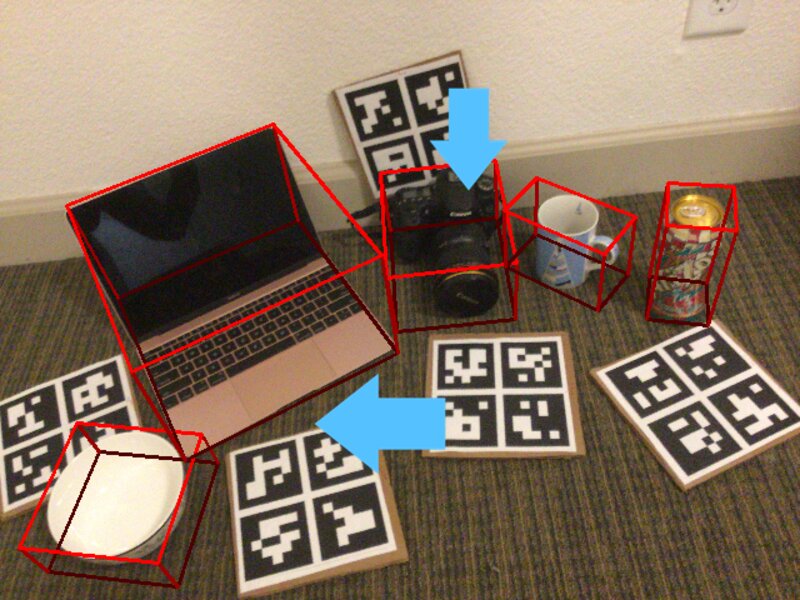}\\

        \raisebox{1.4\height}{\rotatebox{90}{\small Frost}} &
        \includegraphics[width=0.18\textwidth]{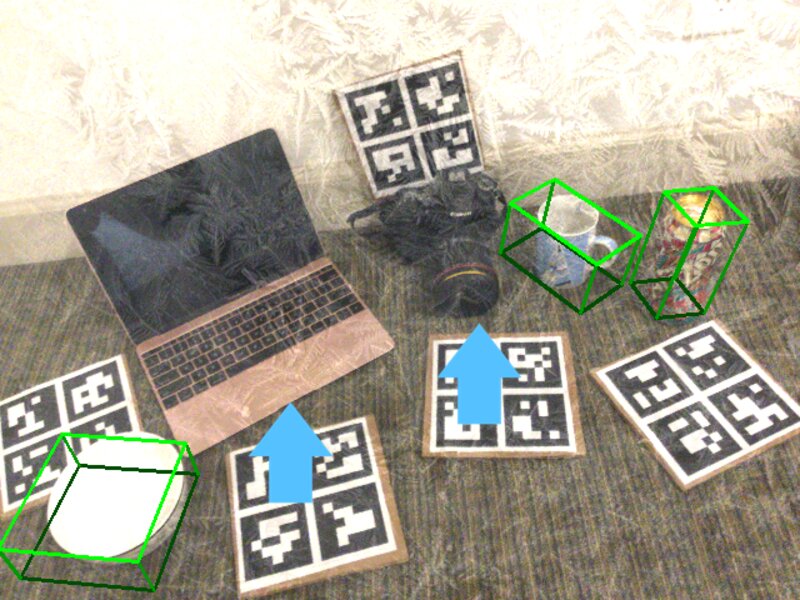} &
        \includegraphics[width=0.18\textwidth]{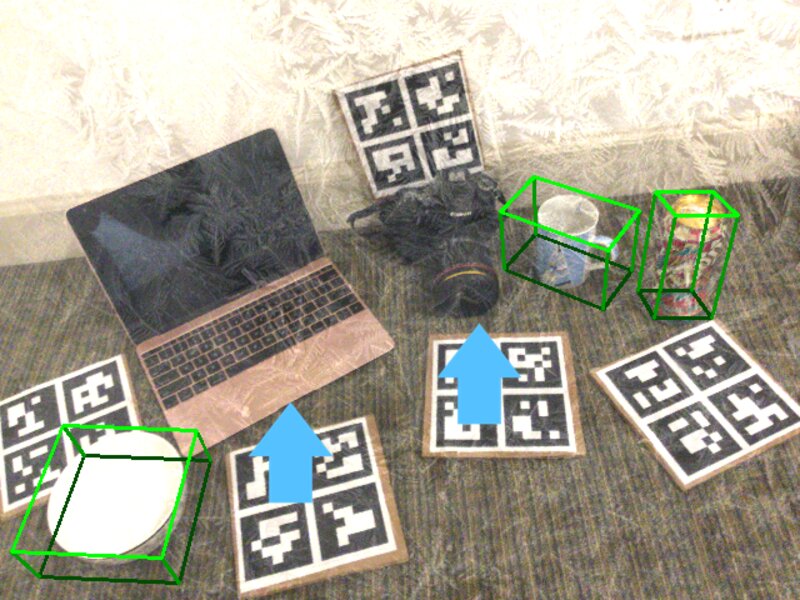} &
        \includegraphics[width=0.18\textwidth]{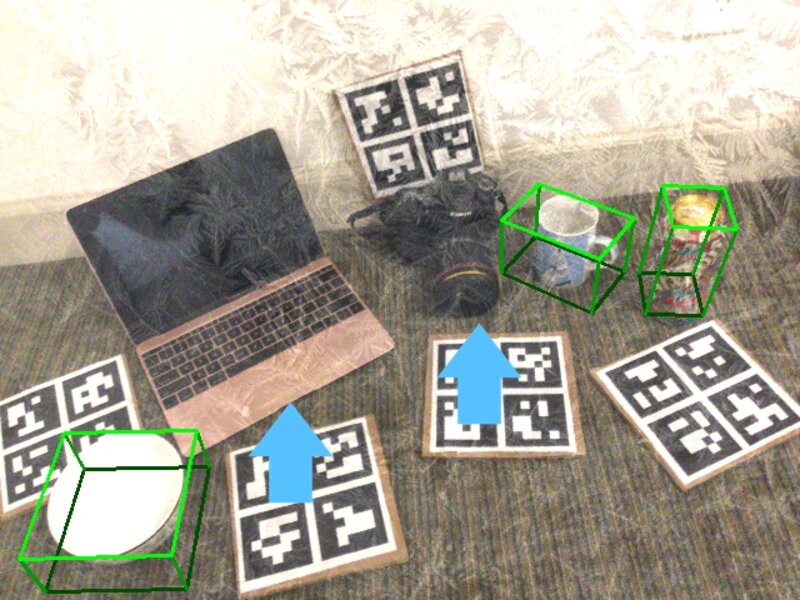} &
        \includegraphics[width=0.18\textwidth]{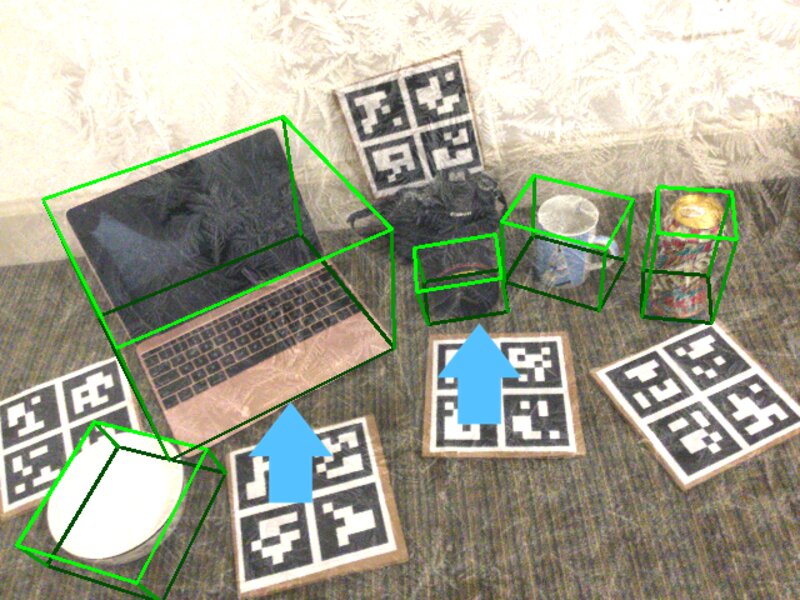}&
        \includegraphics[width=0.18\textwidth]{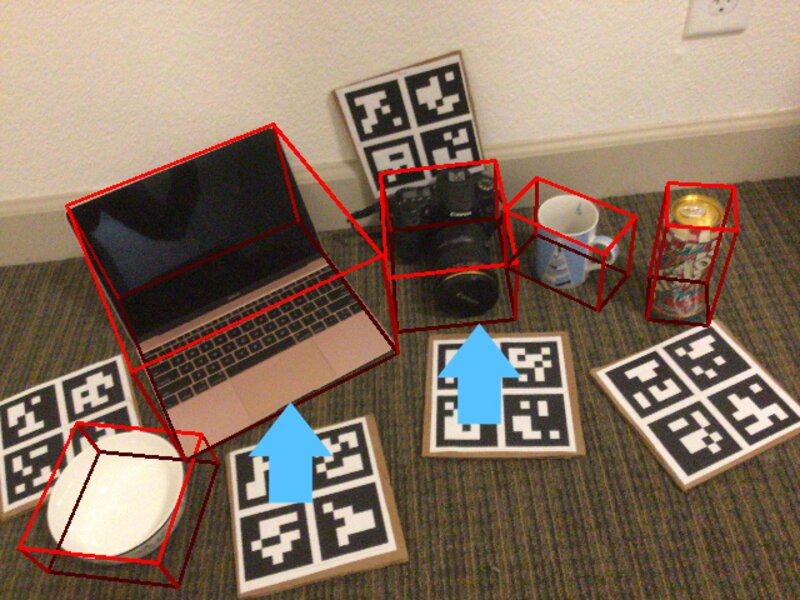}\\

         \raisebox{1.9\height}{\rotatebox{90}{\small Fog}} &
       \includegraphics[width=0.18\textwidth]{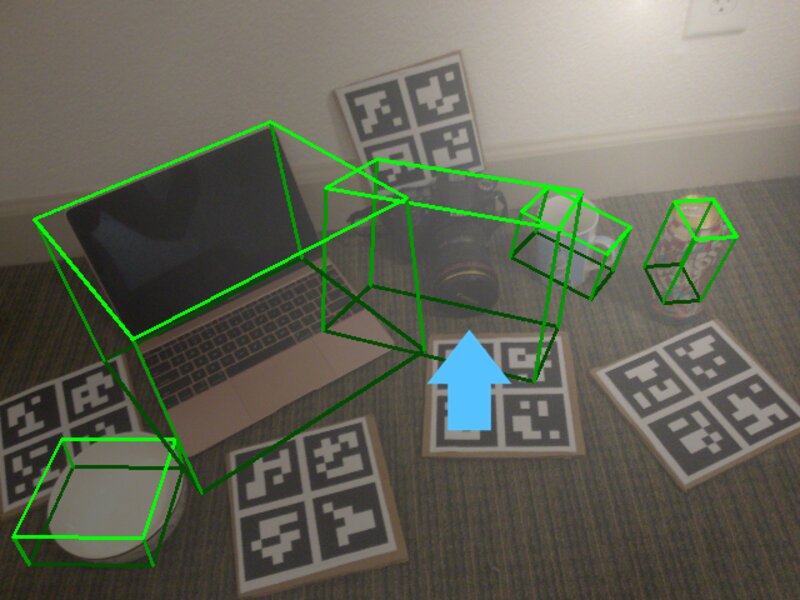} &
        \includegraphics[width=0.18\textwidth]{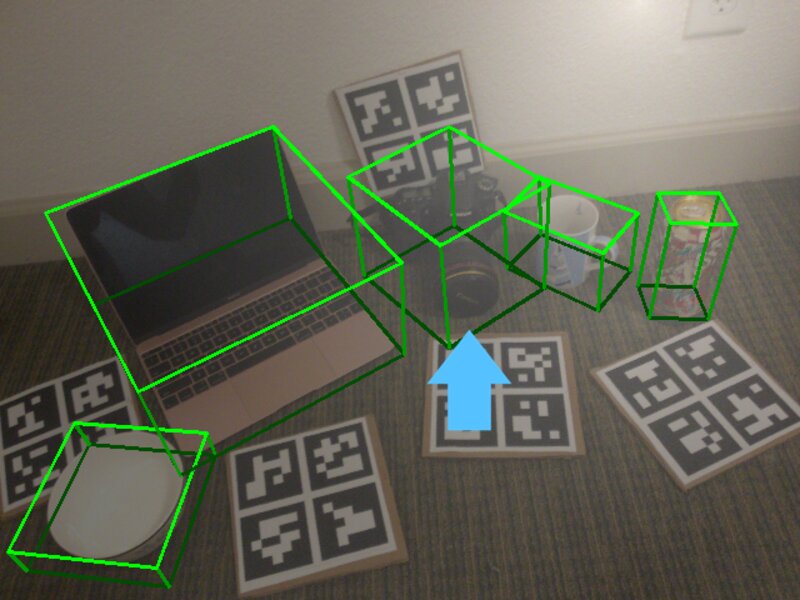} &
        \includegraphics[width=0.18\textwidth]{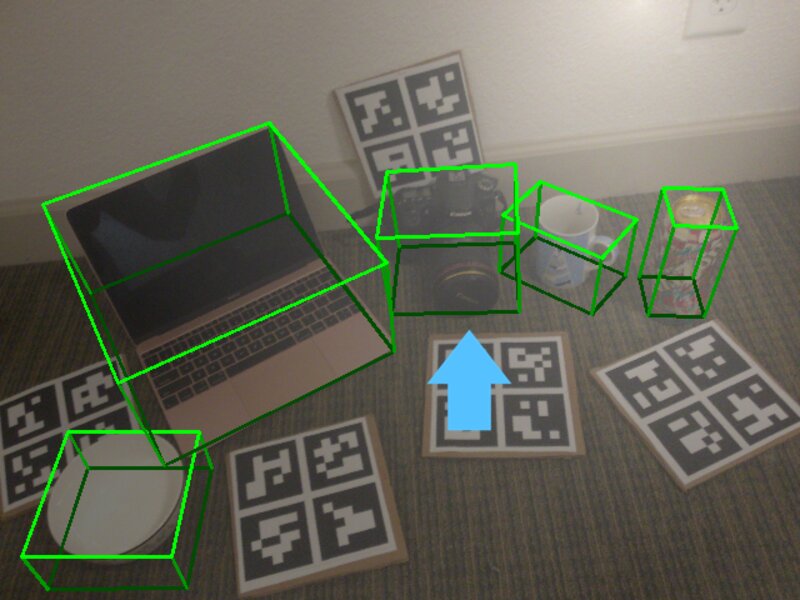} &
        \includegraphics[width=0.18\textwidth]{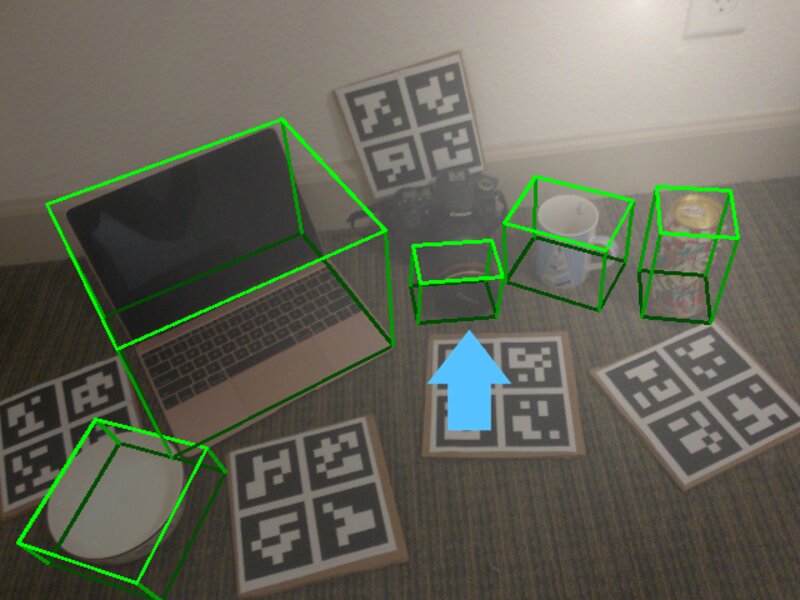}&
        \includegraphics[width=0.18\textwidth]{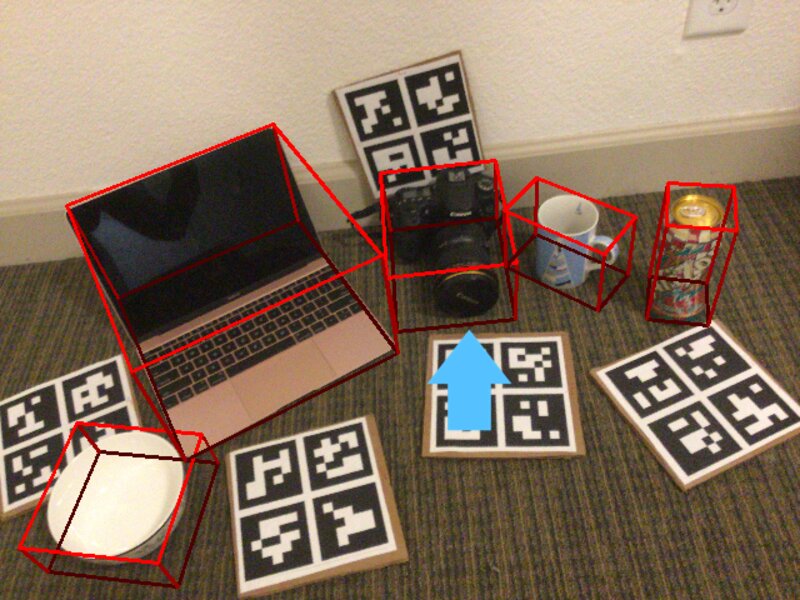}\\

        \space &
        \multicolumn{1}{c}{{OLD-Net}} &
        \multicolumn{1}{c}{{DMSR}} &
        \multicolumn{1}{c}{{LaPose}} &
        \multicolumn{1}{c}{{Ours}} &
        \multicolumn{1}{c}{{Ground Truth}} \\
    \end{tabular}
    }
    \caption{Qualitative comparison on Corrupted NOCS-REAL275\cite{nocs}. We compare our model with all baselines (first to third columns) and with ground truth (last column) across 8 types of corruption.}
    \label{fig:qualitative_noise}
\end{figure*}

\cref{fig:qualitative_noise} presents qualitative examples of object detection and pose estimation on the corrupted REAL275 dataset. 
We can observe that image degradations negatively affect the detection and in turn, the final pose estimation accuracy.
For instance, when applying elastic transformations the reduced detector accuracy causes the laptop to be missed entirely and introduces redundant detections of the camera.
This shows that for degraded images the detection model is a performance bottleneck in current two-stage approaches.
In the single-stage approach, we benefit from significantly more robust detection and pose estimation quality.

\section{Qualitative Results of Dense Matching}
\begin{figure*}
    \centering
    \renewcommand{\arraystretch}{1}  
    \setlength{\tabcolsep}{5pt}  
     \resizebox{\textwidth}{!}{%
    \begin{tabular}{p{3cm}p{3cm}p{3cm}p{3cm}p{3cm}} 
        \includegraphics[width=0.18\textwidth] {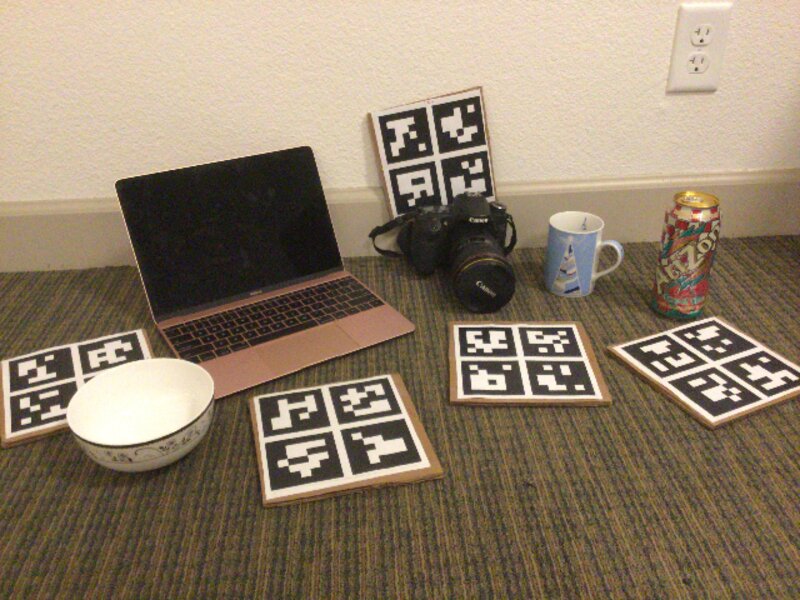}&
        \includegraphics[width=0.18\textwidth]{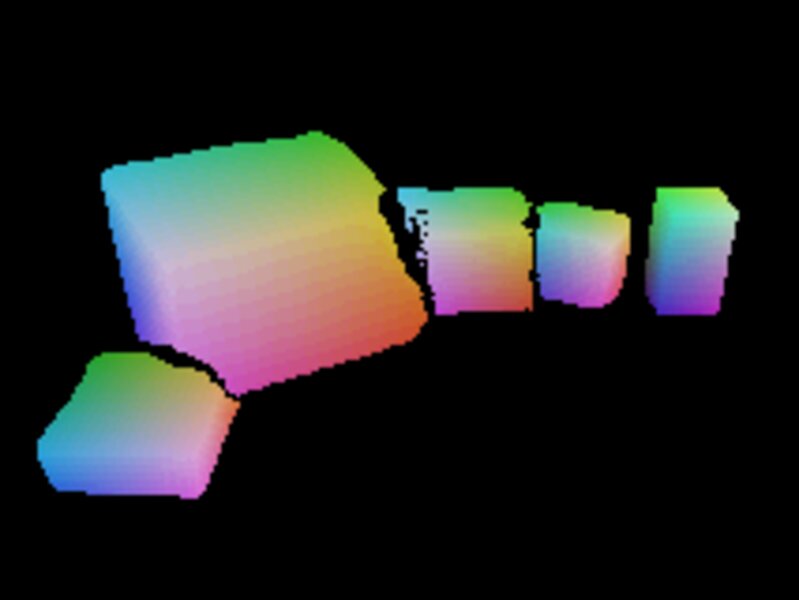} &
        \includegraphics[width=0.18\textwidth]{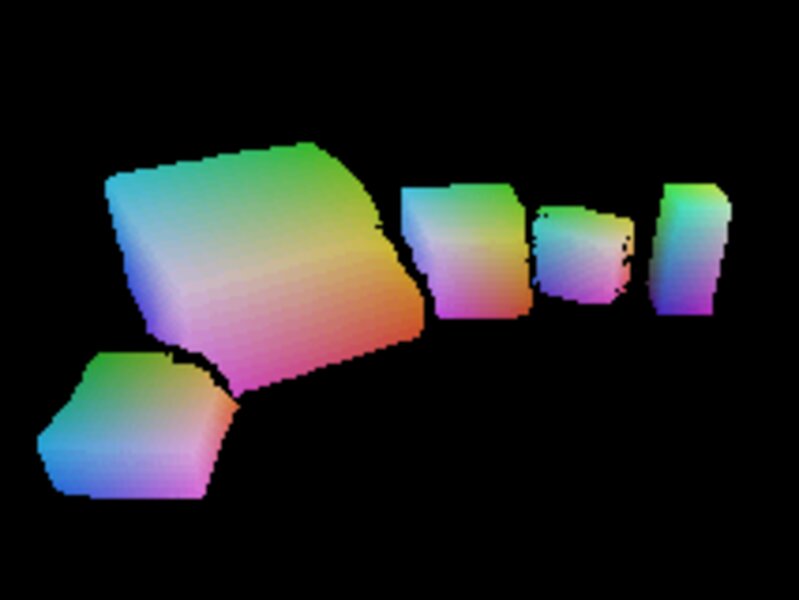} &
        \includegraphics[width=0.18\textwidth]{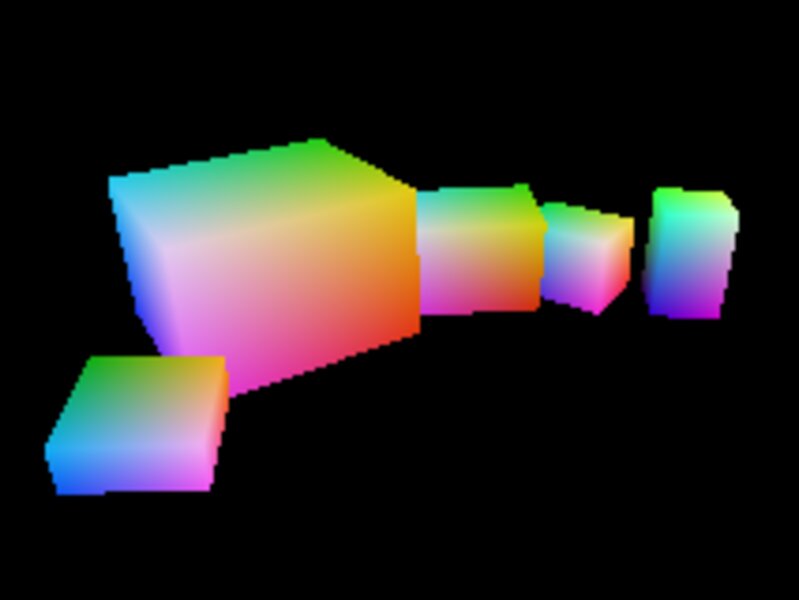}&
        \includegraphics[width=0.18\textwidth]{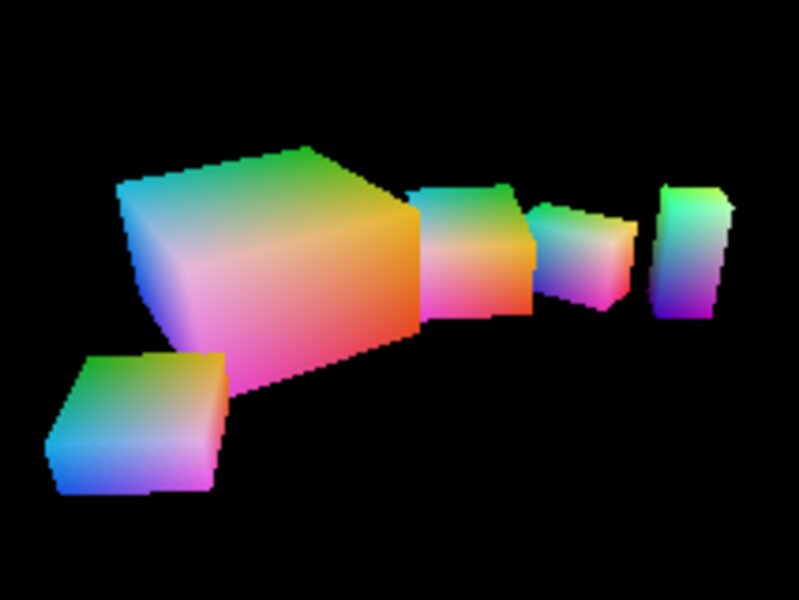}
        \\

        \includegraphics[width=0.18\textwidth] {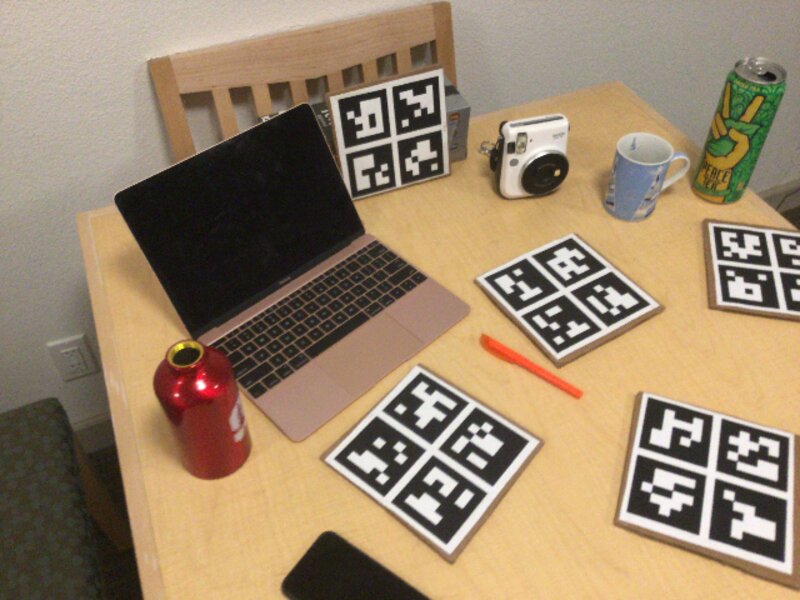}&
        \includegraphics[width=0.18\textwidth]{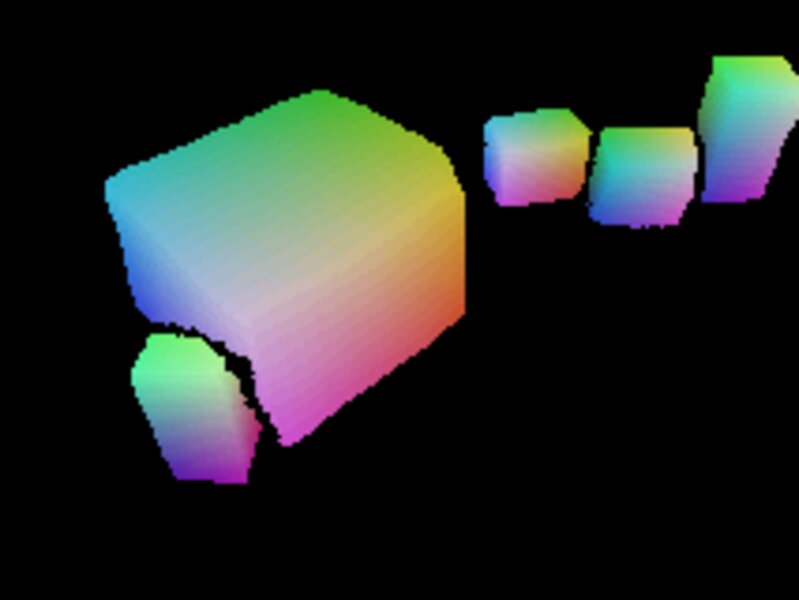} &
        \includegraphics[width=0.18\textwidth]{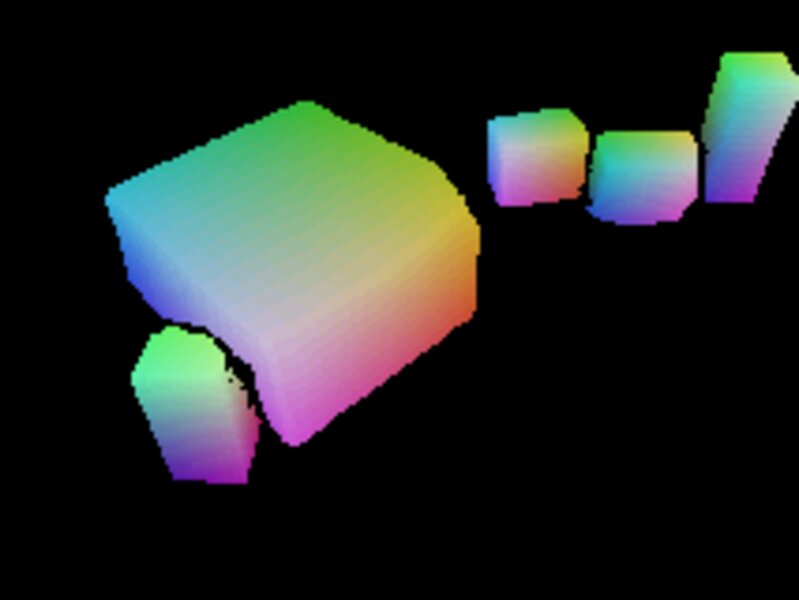} &
        \includegraphics[width=0.18\textwidth]{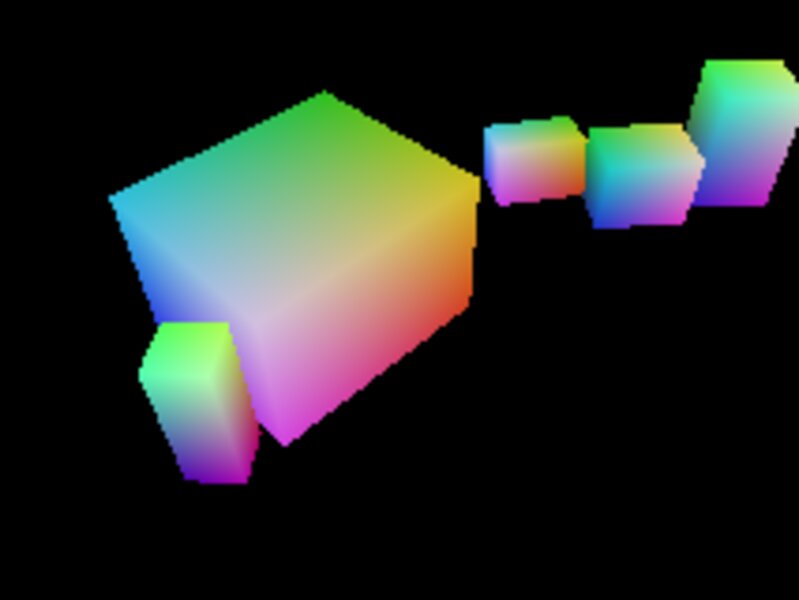}&
        \includegraphics[width=0.18\textwidth]{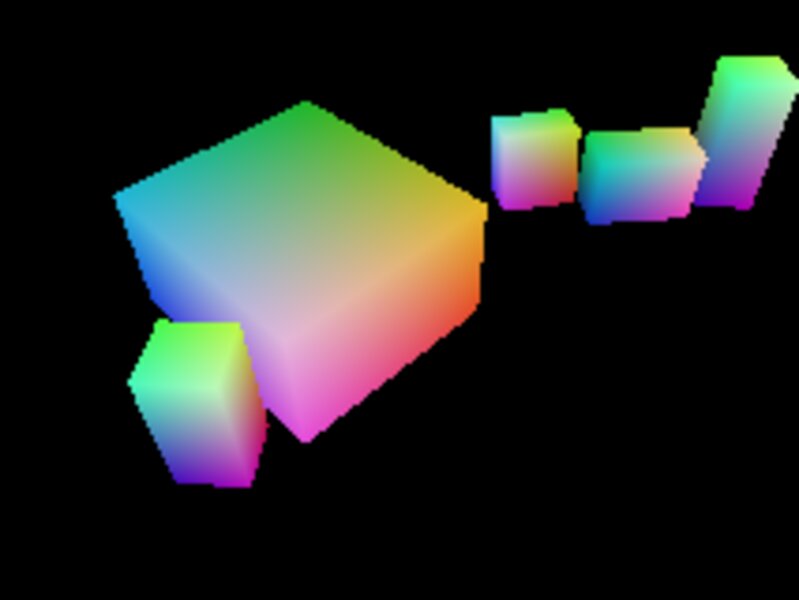}
        \\

        \includegraphics[width=0.18\textwidth] {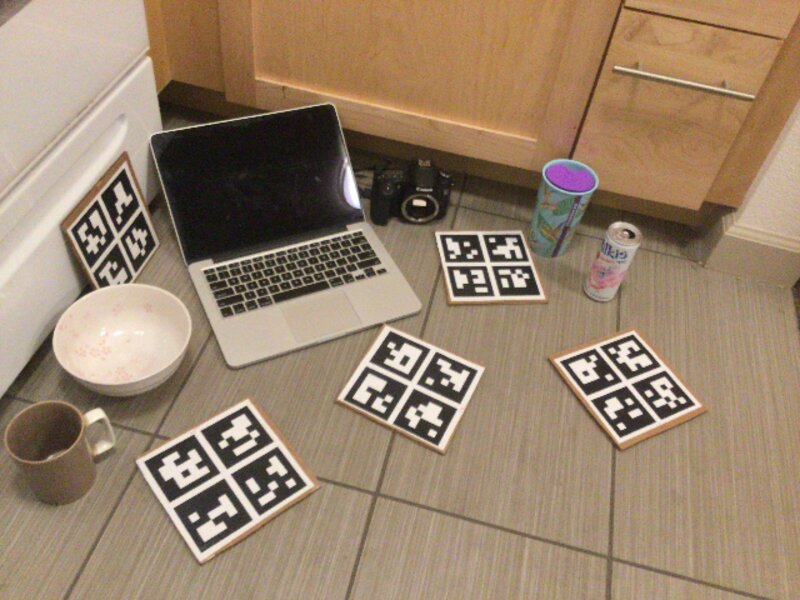}&
        \includegraphics[width=0.18\textwidth]{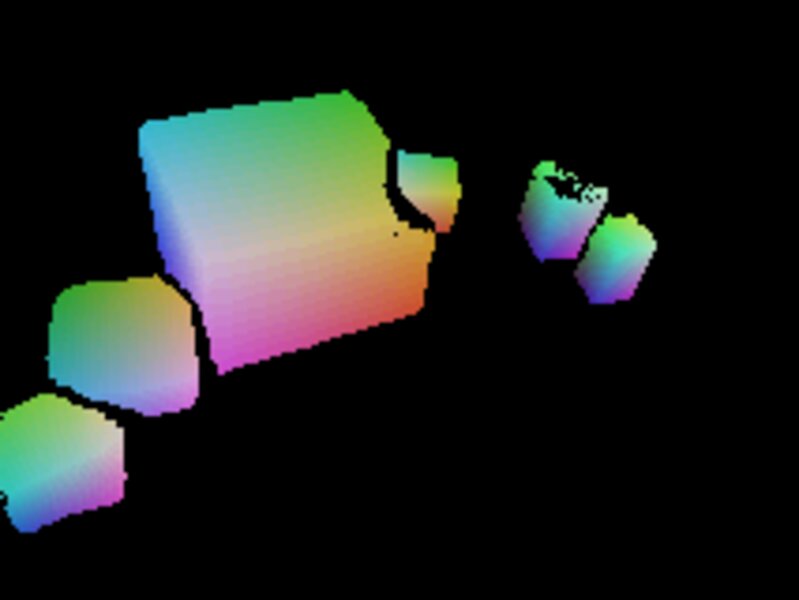} &
        \includegraphics[width=0.18\textwidth]{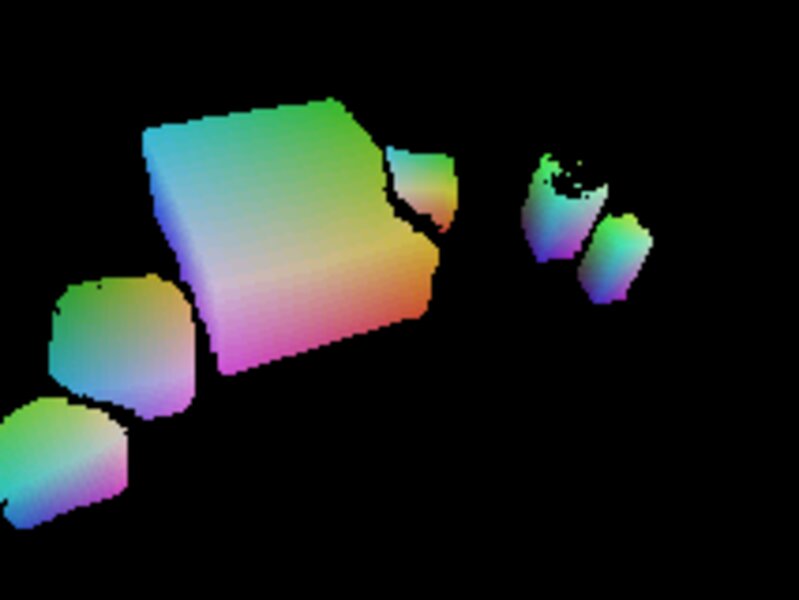} &
        \includegraphics[width=0.18\textwidth]{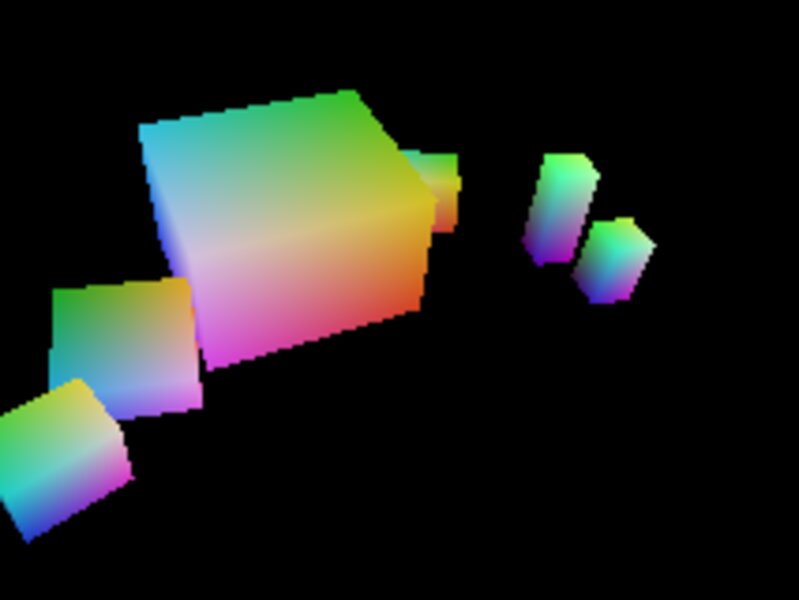}&
        \includegraphics[width=0.18\textwidth]{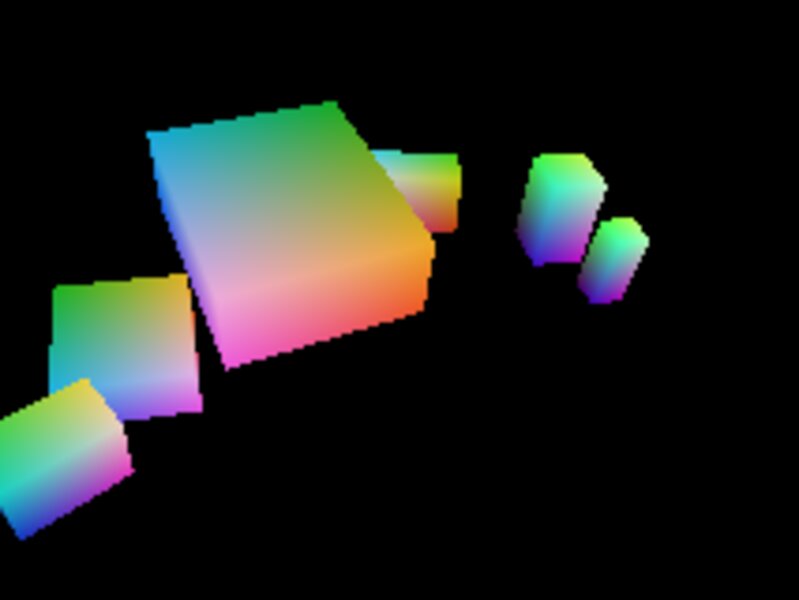}
        \\

        \includegraphics[width=0.18\textwidth] {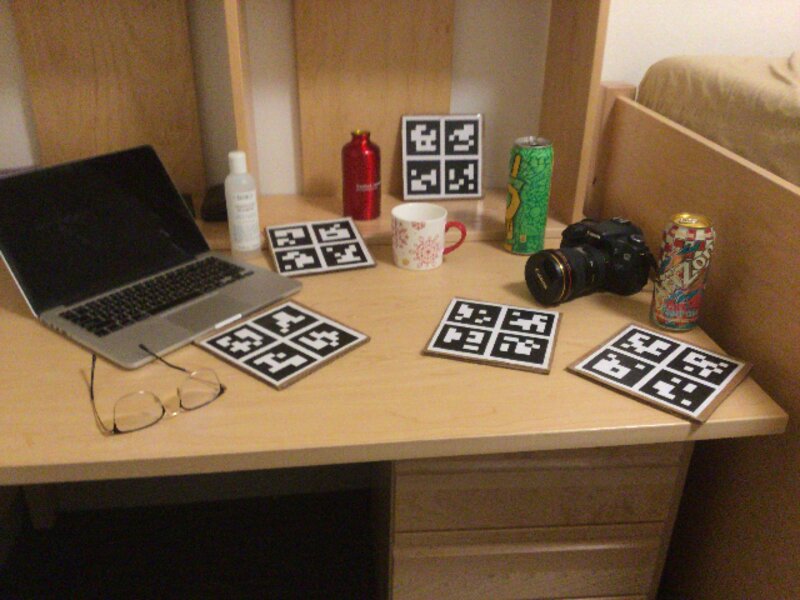}&
        \includegraphics[width=0.18\textwidth]{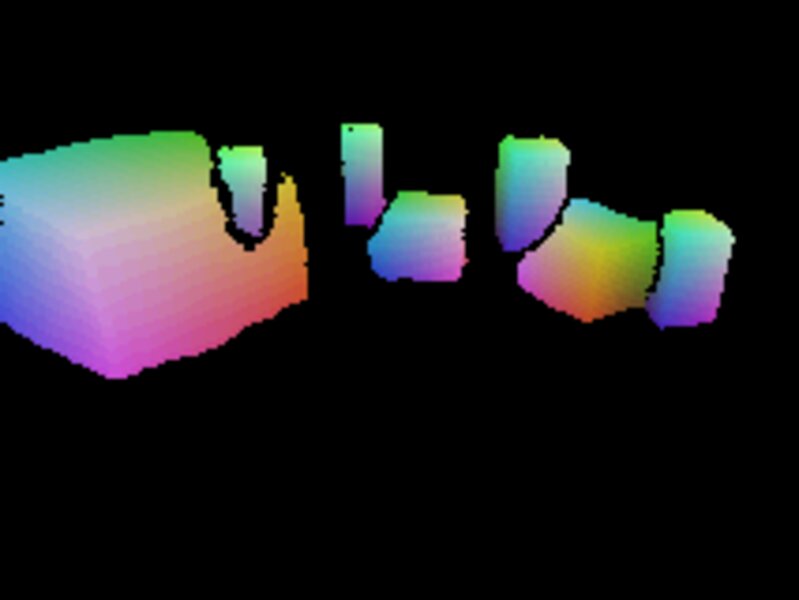} &
        \includegraphics[width=0.18\textwidth]{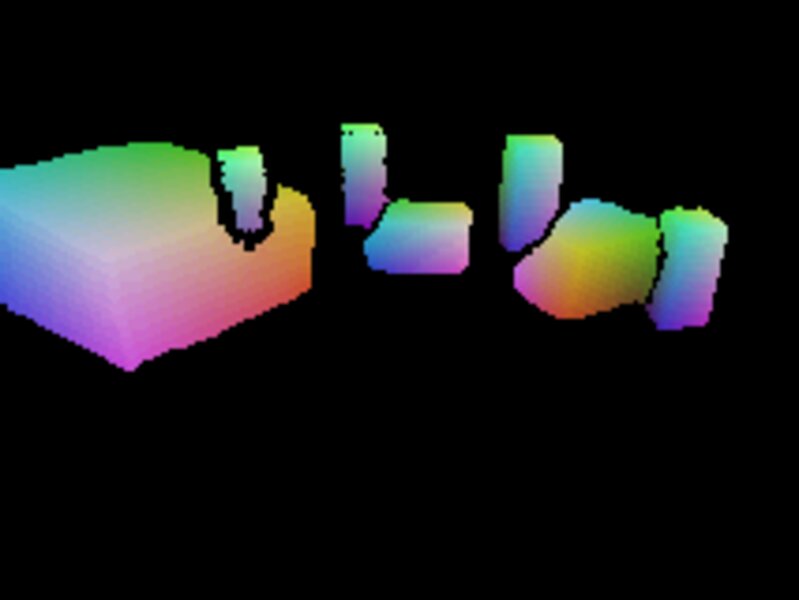} &
        \includegraphics[width=0.18\textwidth]{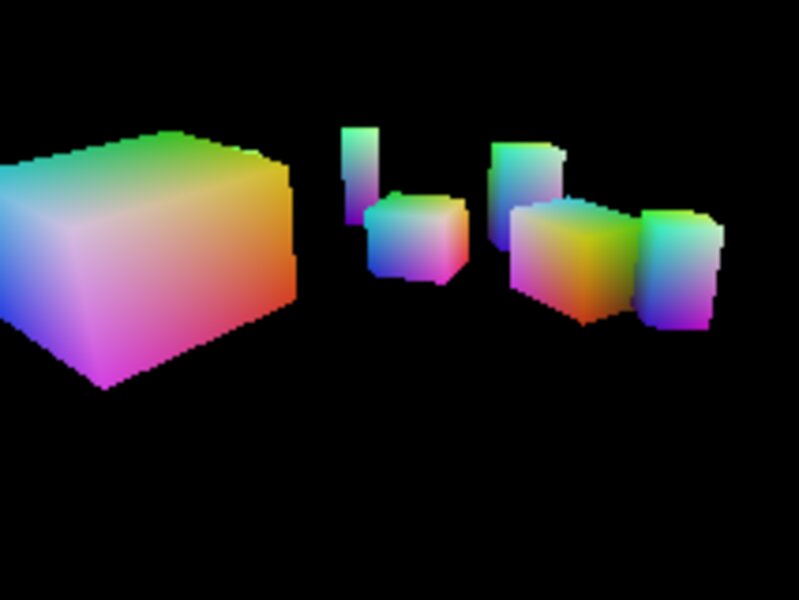}&
        \includegraphics[width=0.18\textwidth]{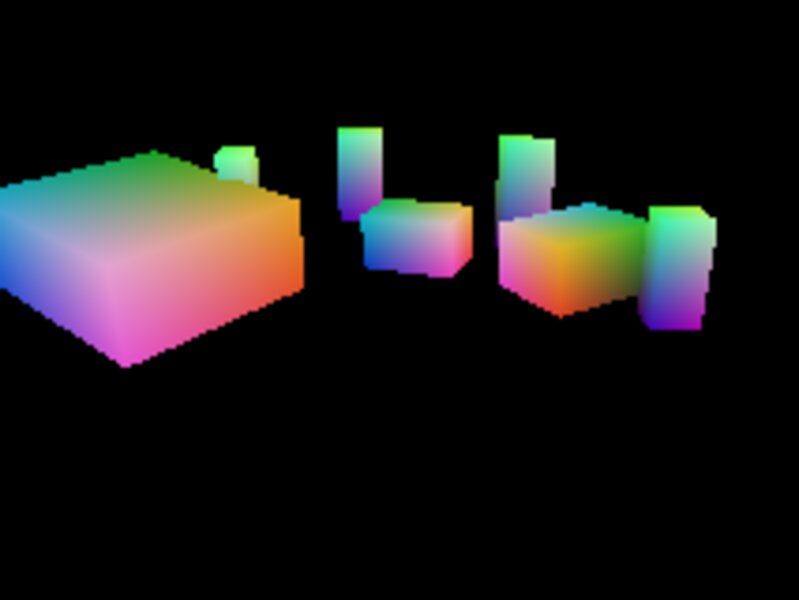}
        \\

        \includegraphics[width=0.18\textwidth] {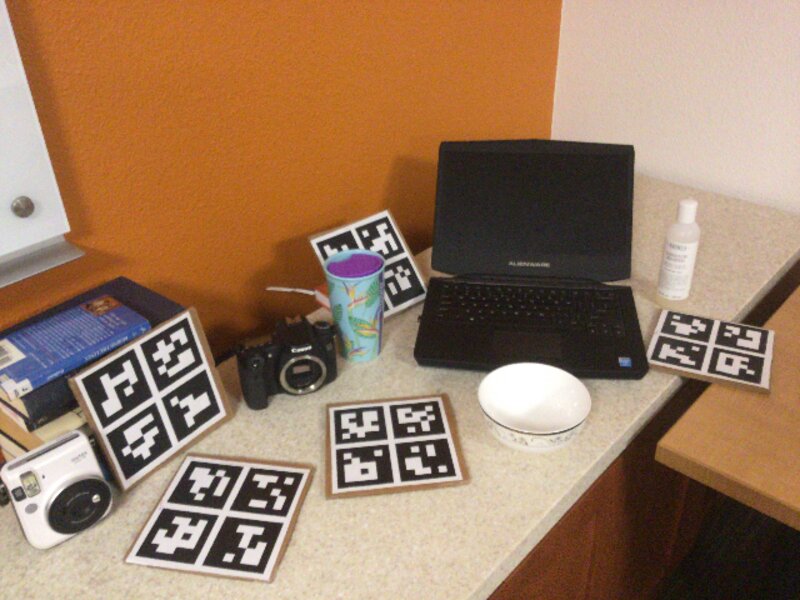}&
        \includegraphics[width=0.18\textwidth]{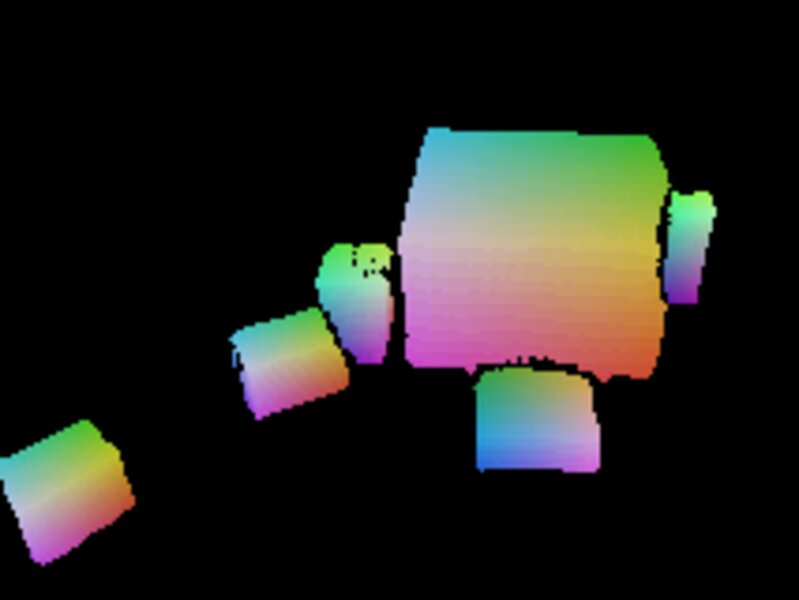} &
        \includegraphics[width=0.18\textwidth]{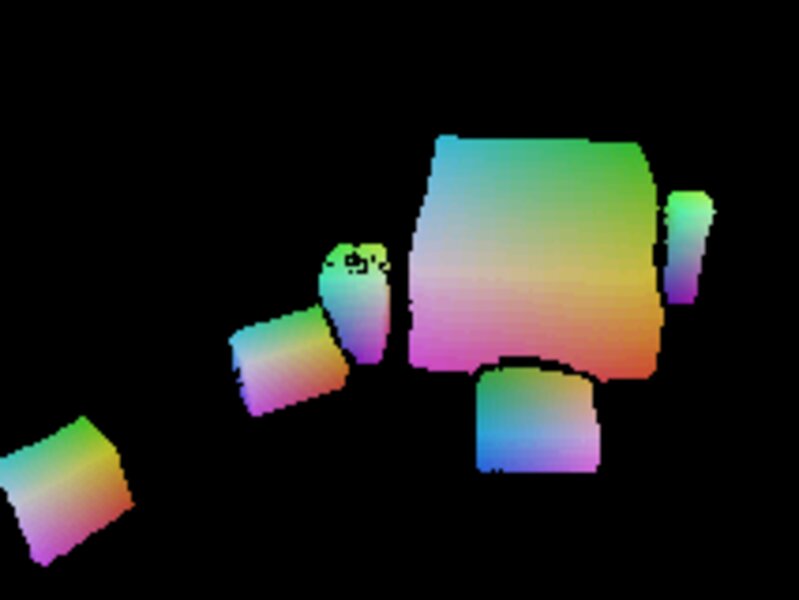} &
        \includegraphics[width=0.18\textwidth]{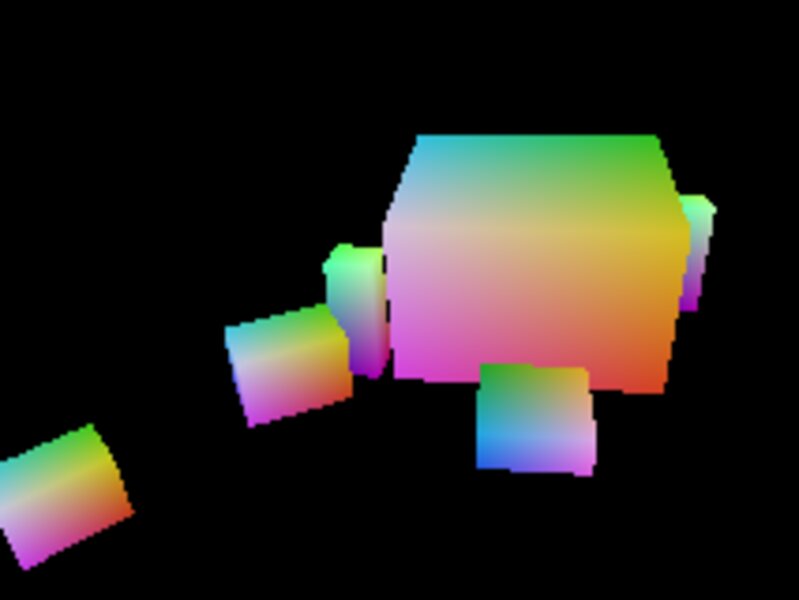}&
        \includegraphics[width=0.18\textwidth]{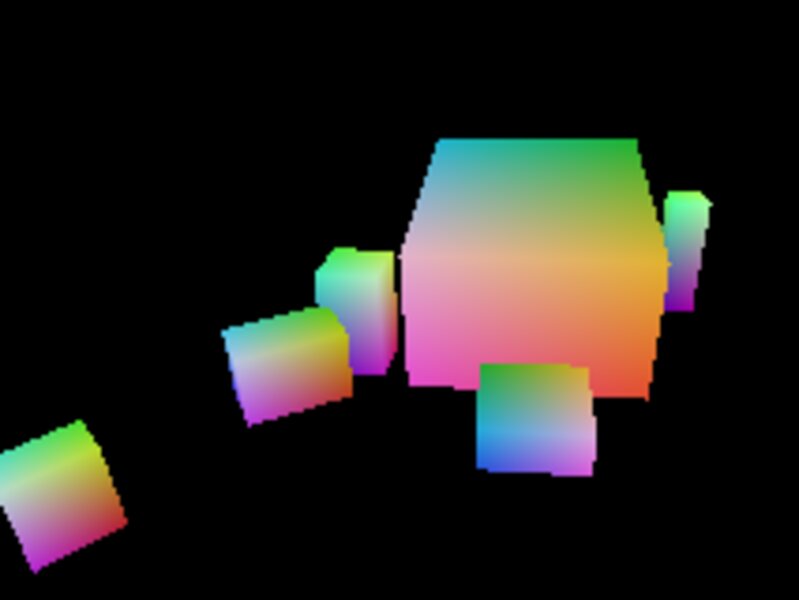}
        \\

        \includegraphics[width=0.18\textwidth] {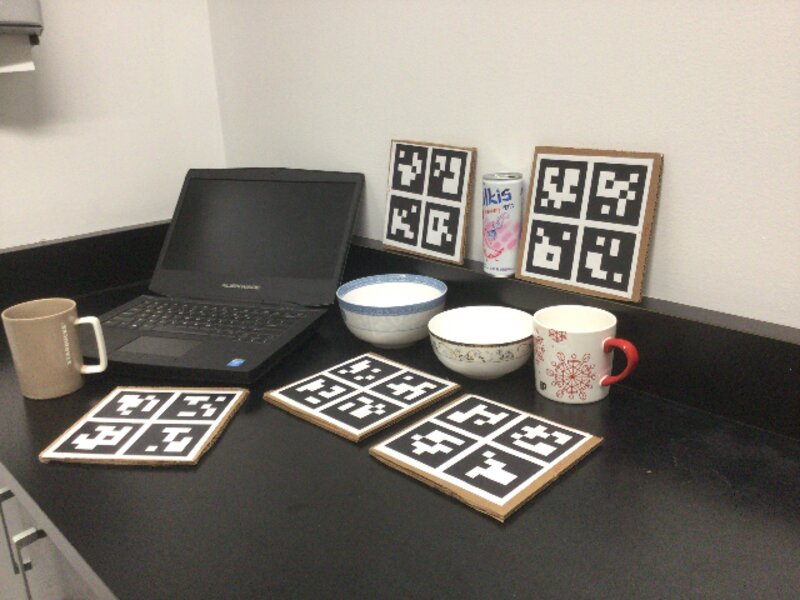}&
        \includegraphics[width=0.18\textwidth]{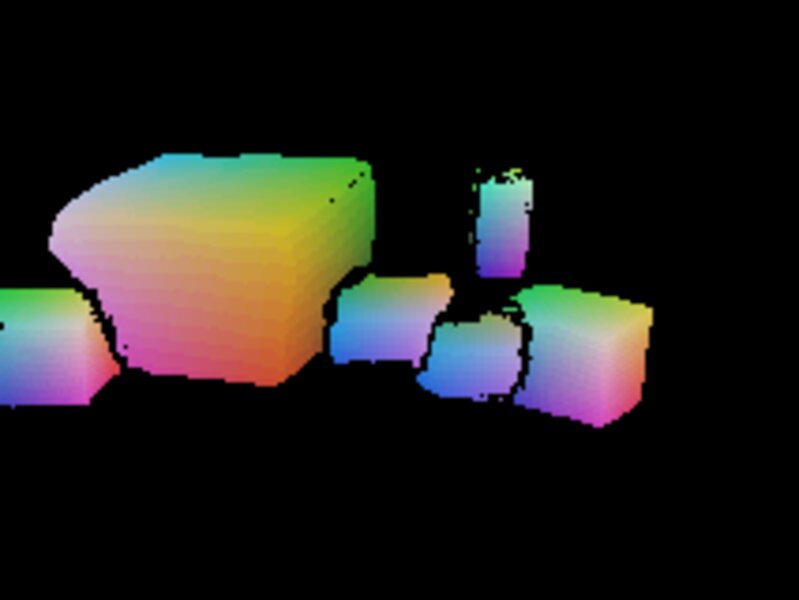} &
        \includegraphics[width=0.18\textwidth]{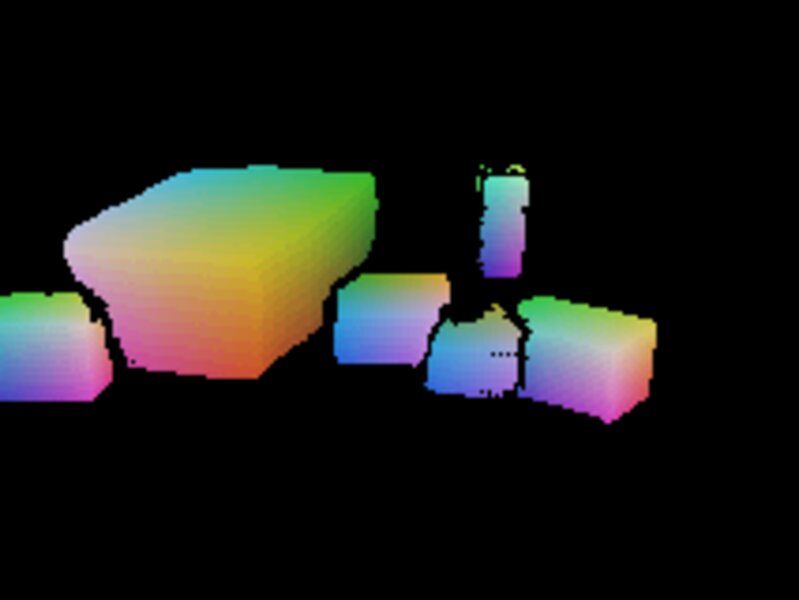} &
        \includegraphics[width=0.18\textwidth]{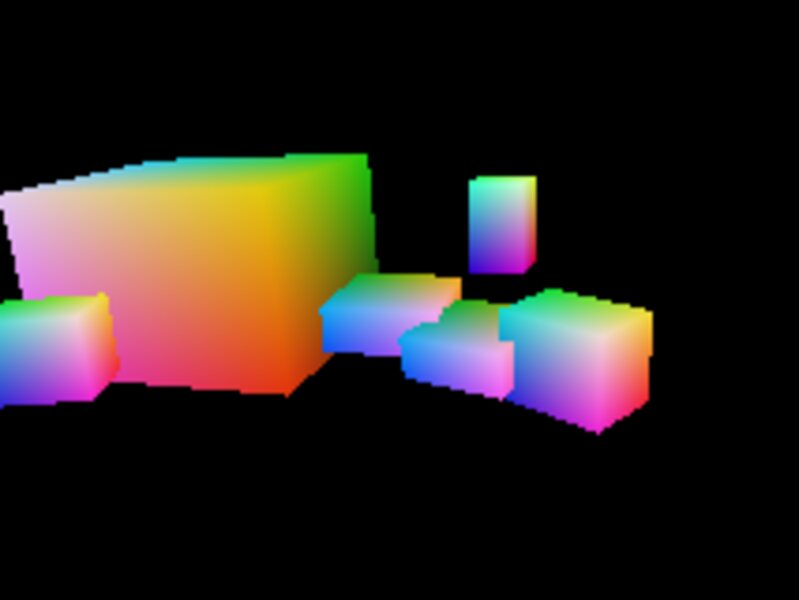}&
        \includegraphics[width=0.18\textwidth]{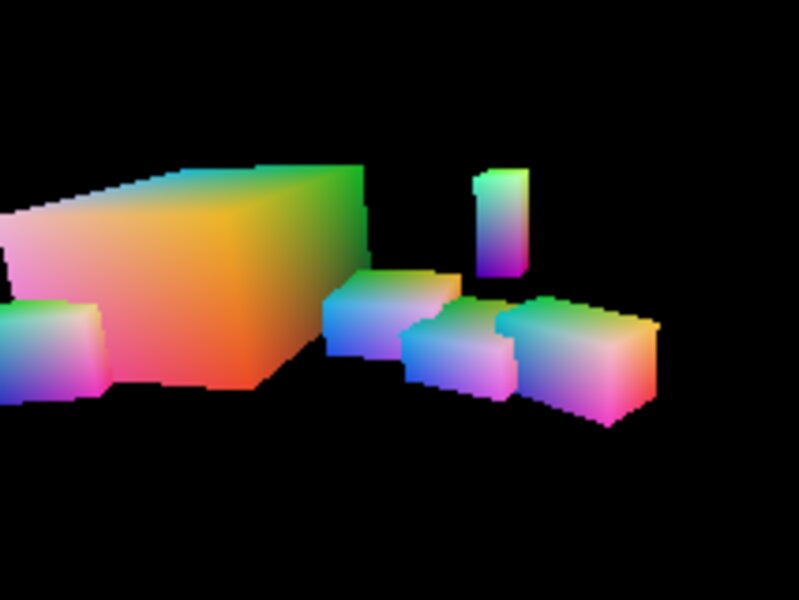}
        \\

        \centering{Input} & 
        \centering{Mean-Scale \\NOCS Estimation} & 
        \centering{Instance-Scale  \\NOCS Estimation}  & 
        \centering{Mean-Scale \\NOCS GT}  & 
        \centering{Instance-Scale \\ NOCS GT}  \\
       
    \end{tabular}
    }
    
    \caption{Visualization of the our dense matching results. We estimate 2D-3D correspondences between image features and our object prototypes with mean scales and instance-level scales. We remove low noisy correspondences using our foreground modeling strategy and confidence scores.
    Our method produces reliable and mostly noise-free correspondences in object regions.
    }
    \label{fig:dense_matching}
\end{figure*}

In \cref{fig:dense_matching}, we present qualitative results of dense 2D-3D correspondence matching. 
We render all NOCS-maps with the geometry of our object prototypes using PyTorch3D~\cite{pytorch3d}. 
To be consistent with our main inference, we remove correspondences with a confidence score below $t_2=0.7$. 
To address object overlap, we prioritize rendering the object closer to the camera.
We observe that our method generates less confident correspondences near object edges, likely due to the ambiguity of these regions during optimization.
Overall, our method produces high-quality correspondences using a simple nearest-neighbor matching approach, indicating that the contrastive training approach produces descriptive features.

\section{Per Category Results}
\begin{figure*}
    \centering
    \renewcommand{\arraystretch}{0.1}  
    \setlength{\tabcolsep}{0pt}  
     \begin{tabular}{p{1cm} c} 
         \rotatebox[origin=l]{90}{\parbox{5cm}{\centering OLD-Net~\cite{oldnet}} } &
         \includegraphics[width=0.8\linewidth ,trim={8cm 0cm 0 15},clip]{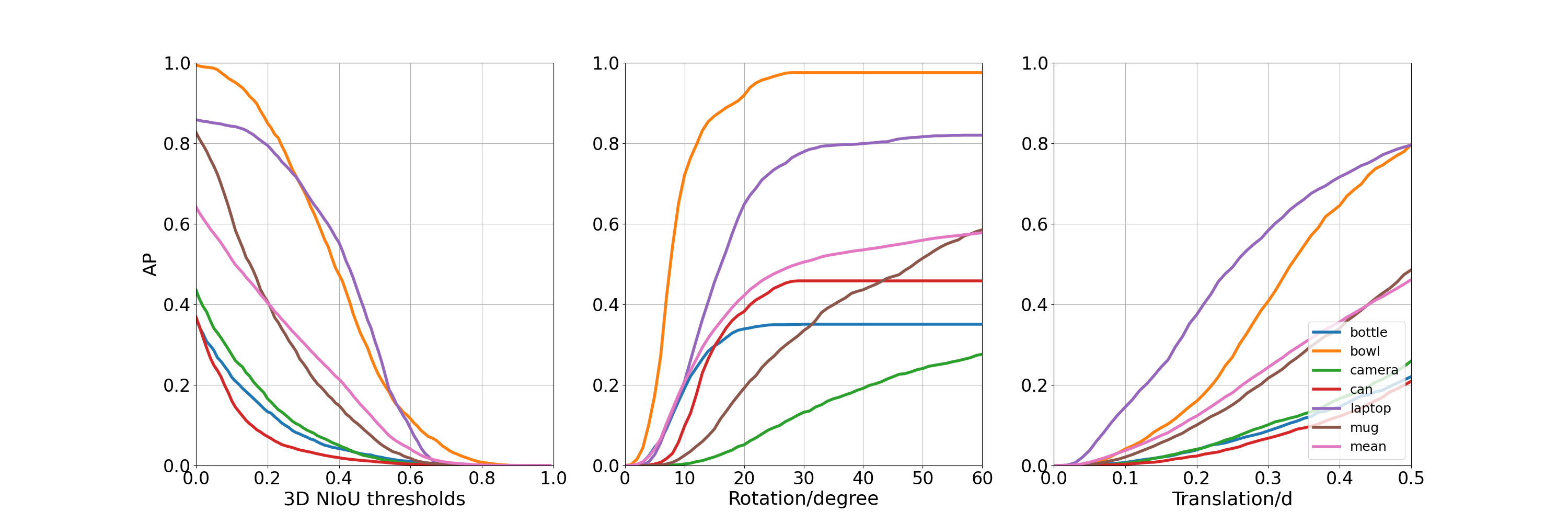}\\

         \rotatebox[origin=l]{90}{\parbox{5cm}{\centering DMSR~\cite{dmsr}}} & \includegraphics[width=0.8\linewidth ,trim={8cm 0cm 0 15},clip]{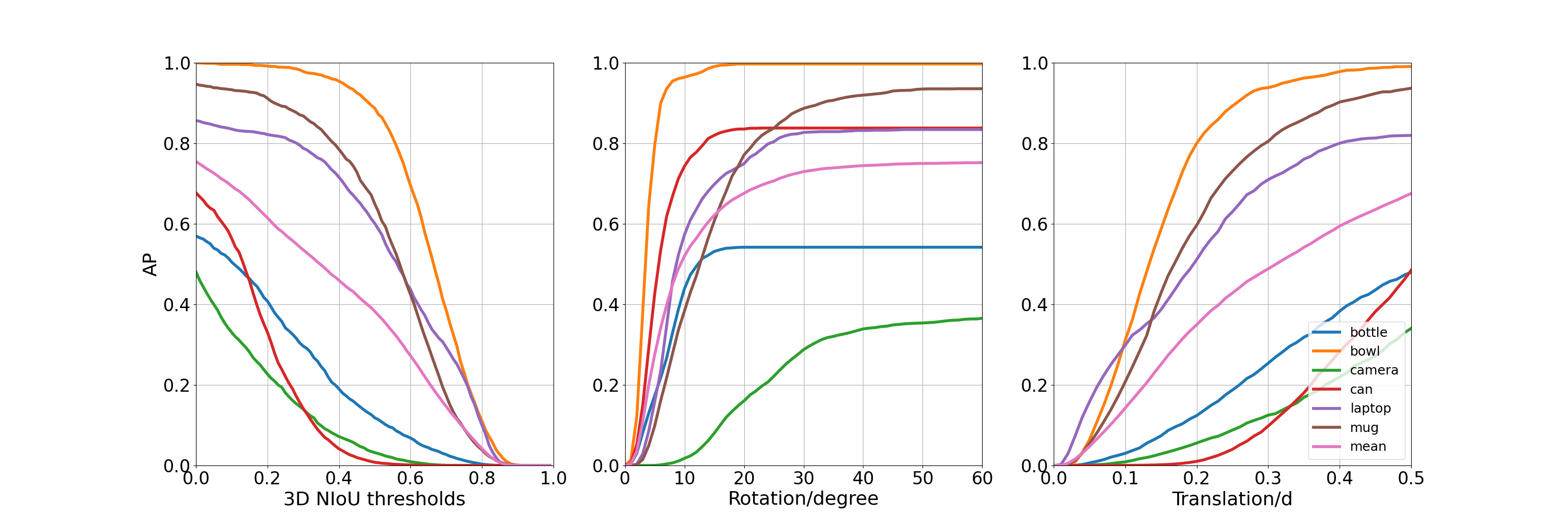} \\

        \rotatebox[origin=l]{90}{\parbox{5cm}{\centering LaPose~\cite{lapose}}} & \includegraphics[width=0.8\linewidth ,trim={8cm 0cm 0 15},clip]{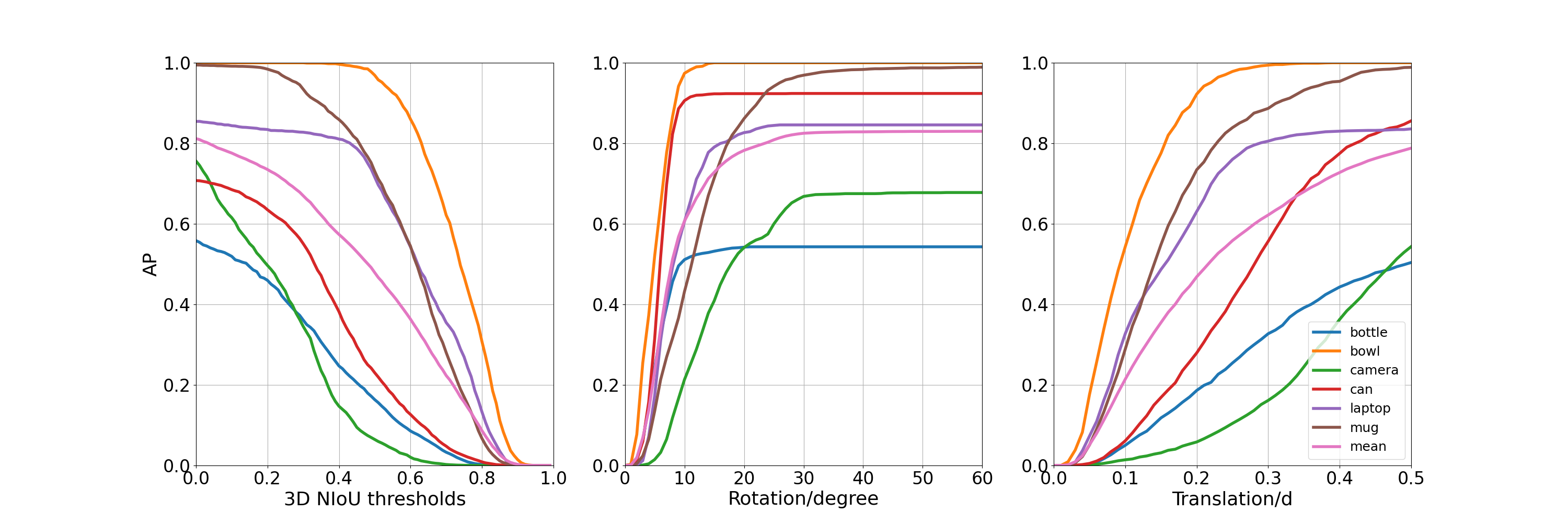} \\

        \rotatebox[origin=l]{90}{\parbox{5cm}{\centering Ours}} & 
         \includegraphics[width=0.8\linewidth ,trim={8cm 0cm 0 15},clip]{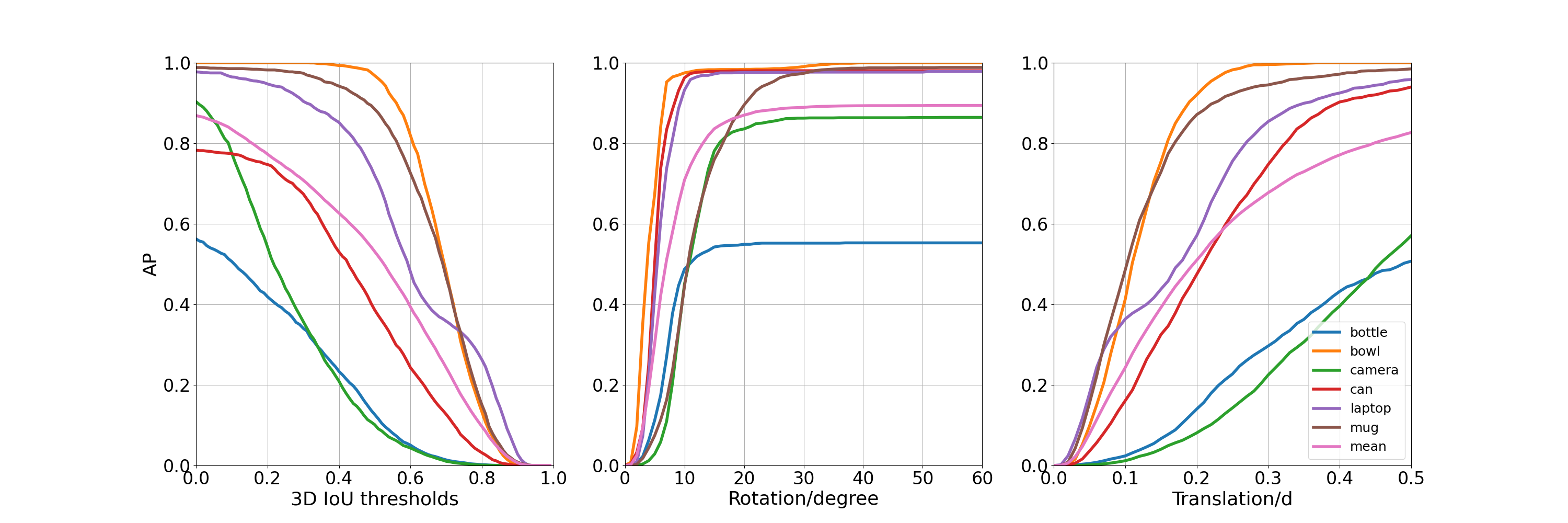} \\
    \end{tabular}
    \caption{We show mean Average Precision (mAP) on REAL275\cite{nocs} using scale-agnostic metrics. We compare our model with all baselines that have public code. Noticeably, our method has significantly increased rotation accuracy on the challenging non-symmetric categories camera and laptop.}
    \label{fig:mAP}
\end{figure*}

\cref{fig:mAP} shows category-level pose estimation results using our method and each baseline which has public code~\cite{lapose,dmsr,oldnet}. 
Notably, our approach consistently improves the mean performance. 
Furthermore, our method outperforms others in the challenging non-symmetric camera and laptop categories. 
This result highlights the effectiveness of our object representation and the single-stage modeling strategy.


\end{document}